\documentclass[journal]{IEEEtran}
\ifCLASSINFOpdf
\else
\fi
\usepackage{makecell, tabularx}
\usepackage[export]{adjustbox}

\usepackage{algorithm}
\usepackage{tabularx}
\usepackage{rotating}
\usepackage[algo2e]{algorithm2e} 
\usepackage{algpseudocode}
\usepackage{url}
\usepackage[table,xcdraw]{xcolor}
\usepackage{hyperref}
\usepackage[numbers]{natbib}
\usepackage{amsmath,amssymb}
\usepackage{bbm}
\usepackage{xcolor, soul}
\usepackage[utf8]{inputenc}
\usepackage[english]{babel}
\usepackage{graphicx}
\usepackage{booktabs}
\usepackage{multirow}
\usepackage{subcaption}
\usepackage{placeins}

\usepackage{makecell}

\begin{document}
\title{MaskCD: A Remote Sensing Change Detection Network Based on Mask Classification}
\author{Weikang~Yu,~\IEEEmembership{Student Member,~IEEE,}
        Xiaokang~Zhang,~\IEEEmembership{Member,~IEEE,}
        Samiran~Das,
        Xiao Xiang~Zhu,~\IEEEmembership{Fellow,~IEEE,}
        and~Pedram~Ghamisi,~\IEEEmembership{Senior Member,~IEEE}% <-this % stops a spa
\thanks{W. Yu is with Helmholtz Institute Freiberg for Resource Technology, Helmholtz-Zentrum Dresden-Rossendorf, 09599 Freiberg, Germany, and also with the Chair of Data Science in Earth Observation, Technical University of Munich (TUM), Germany (e-mail: w.yu@hzde.de)}
\thanks{Xiaokang~Zhang is with the School of Information Science and Engineering, Wuhan University of Science and Technology, Wuhan 430081, China (e-mail: natezhangxk@gmail.com). \textit{(Corresponding Author: Xiaokang Zhang)}}
\thanks{Samiran Das is with the School of Engineering, Shiv Nadar University, Noida.}
\thanks{Xiao Xiang Zhu is with the Chair of Data Science in Earth Observation, Technical University of Munich (TUM), and with Munich Center for Machine Learning.}
\thanks{P. Ghamisi is with Helmholtz Institute Freiberg for Resource Technology, Helmholtz-Zentrum Dresden-Rossendorf, 09599 Freiberg, Germany, and also with Lancaster University, LA1 4YR Lancaster, U.K. (e-mail: p.ghamisi@hzdr.de).}% <-this % stops a space
}

\markboth{Journal of \LaTeX\ 2024}%
{Shell \MakeLowercase{\textit{et al.}}: Maskformer for Remote Sensing Image Change Detection}

\maketitle

\begin{abstract}
Change detection (CD) from remote sensing (RS) images using deep learning has been widely investigated in the literature. It is typically regarded as a pixel-wise labeling task that aims to classify each pixel as changed or unchanged. Although per-pixel classification networks in encoder-decoder structures have shown dominance, they still suffer from imprecise boundaries and incomplete object delineation at various scenes. For high-resolution RS images, partly or totally changed objects are more worthy of attention rather than a single pixel. Therefore, we revisit the CD task from the mask prediction and classification perspective and propose MaskCD to detect changed areas by adaptively generating categorized masks from input image pairs. Specifically, it utilizes a cross-level change representation perceiver (CLCRP) to learn multiscale change-aware representations and capture spatiotemporal relations from encoded features by exploiting deformable multihead self-attention (DeformMHSA). Subsequently, a masked-attention-based detection transformers (MA-DETR) decoder is developed to accurately locate and identify changed objects based on masked attention and self-attention mechanisms. It reconstructs the desired changed objects by decoding the pixel-wise representations into learnable mask proposals and making final predictions from these candidates. Experimental results on five benchmark datasets demonstrate the proposed approach outperforms other state-of-the-art models. Codes and pretrained models are available online (\url{https://github.com/EricYu97/MaskCD}).
\end{abstract}
\begin{IEEEkeywords}
Mask classification, Masked attention, Deformable attention, Change detection, Deep learning, Remote Sensing
\end{IEEEkeywords}

\IEEEpeerreviewmaketitle

\section{Introduction}
\IEEEPARstart{T}{he} constant evolution of Earth's surface necessitates advanced change detection (CD) methodologies, which aim to distinguish the surface changes from the co-registered images captured in the same scene at different times. With an exceptional wealth of data provided by satellite and airborne sensors, many CD datasets and approaches have been proposed for various real-world applications, including urban management \cite{papadomanolaki2021deep}, disaster assessment \cite{zhang2023cross}, and agriculture \cite{mardian2021evaluating}. However, analyzing these vast datasets for subtle and significant changes presents a substantial challenge. Generally, CD methods based on bi-temporal RS images fall into three categories: image algebra, image transformation, and machine learning. The image algebra-based approaches utilize image differencing \cite{bruzzone2000automatic}, change vector analysis (CVA) \cite{johnson1998change}, or image regression \cite{sun2021sparse} to generate difference features containing the change magnitudes. Conversely, image transformation methods, employing techniques like principal component analysis and histogram trend similarity \cite{lv2019novel}, enhance the change information within the bi-temporal images by transforming them into alternative spaces. Both image algebra and transformation methods rely on a proper threshold or a segmentation algorithm to decide the change regions \cite{li2016semi}. Lastly, machine learning-based approaches leverage spatio-temporal features learned from training samples to classify pixels into changed and unchanged categories, employing a decision boundary for discrimination \cite{nemmour2006multiple}.
\begin{figure}[htp]
    \centering
\subfloat[Per-pixel Classification]{%
  \includegraphics[clip,width=0.75\columnwidth]{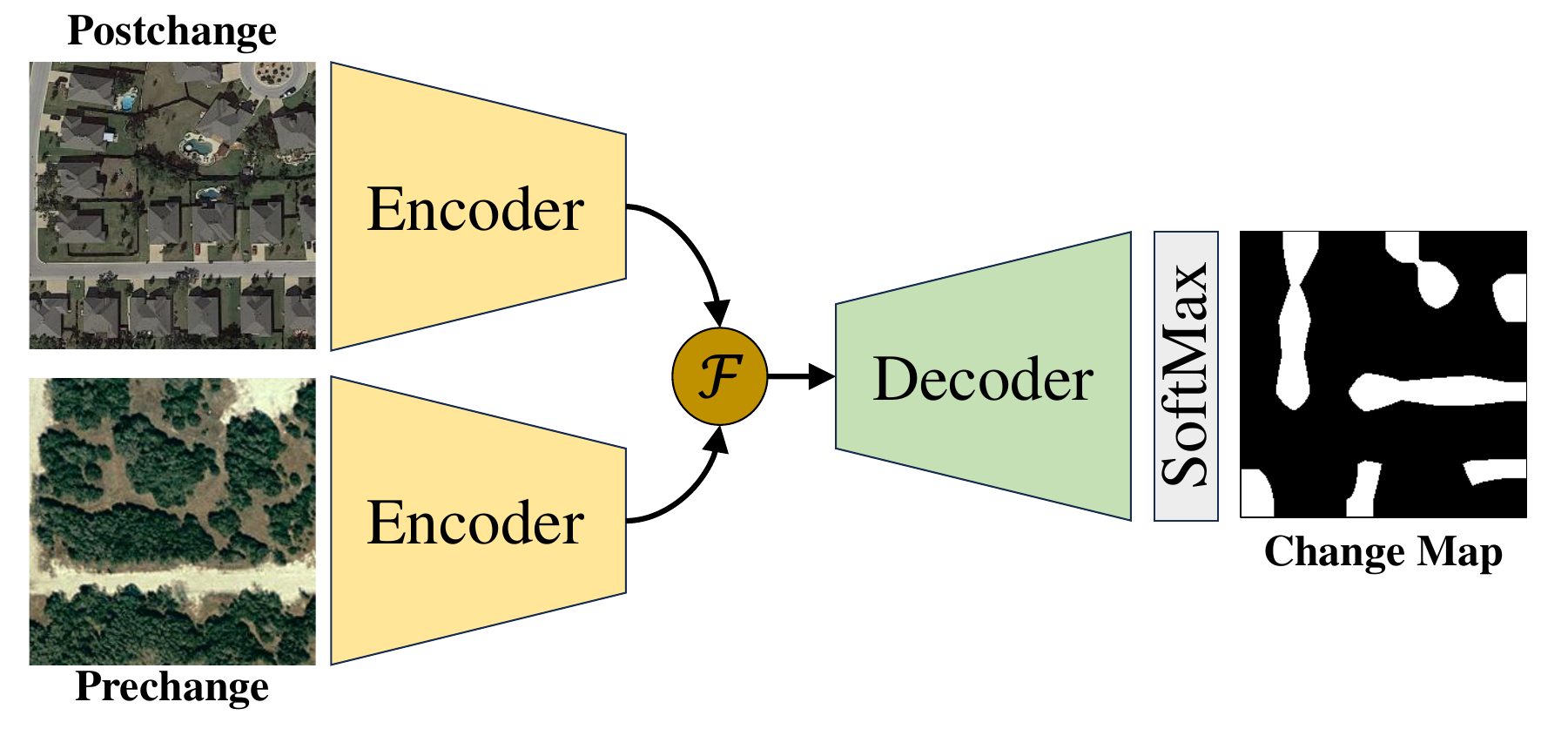}%
}

\subfloat[Mask Classification]{%
  \includegraphics[clip,width=\columnwidth]{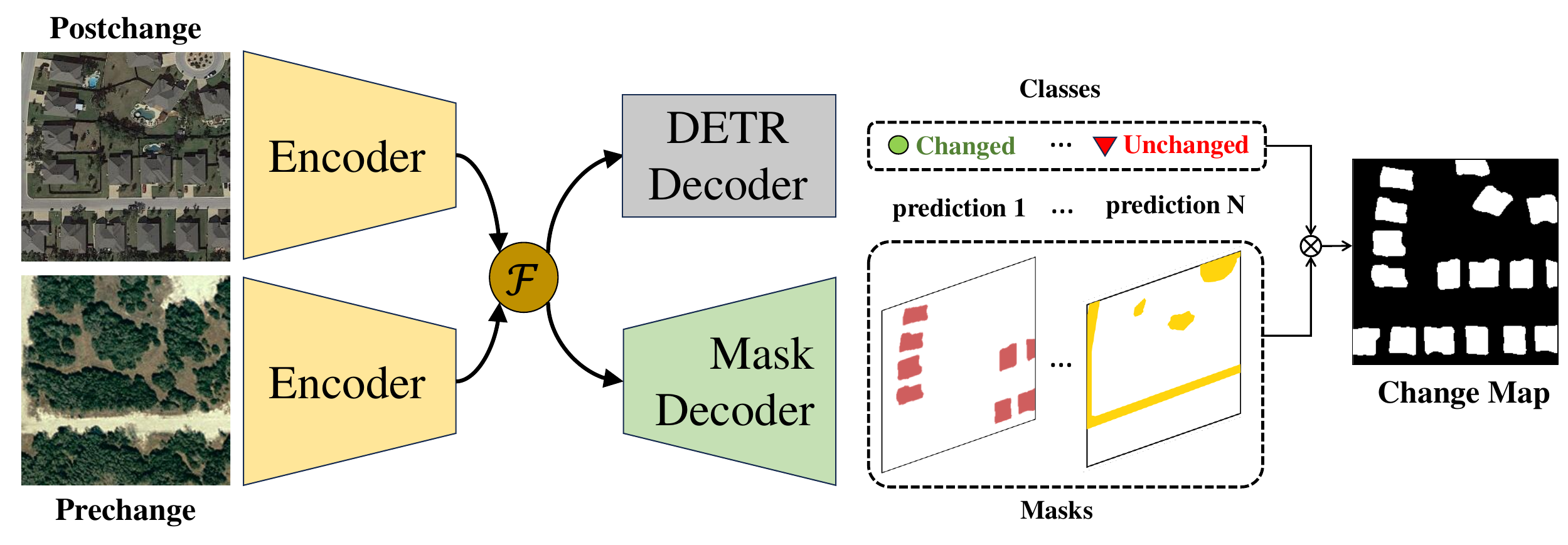}%
}
\caption{Comparisons between (a) per-pixel classification-based and (b) mask classification-based CD methods.}
\label{fig:comparison}
\end{figure}

With the development of deep learning techniques over the decades, CNN-based methods have become the dominant solution for RS-CD tasks \cite{zheng2021clnet}. In contrast to using hand-crafted features in traditional machine learning-based CD methods, deep features generated by CNNs can effectively characterize change information by exploiting conceptual high-level and abstract representations of bi-temporal image pairs \cite{zhang2020deeply}.
Inspired by the remarkable performance of the Transformer models, many recent CD methods attempted to incorporate the self-attention mechanism to model long-range dependencies within deep features for better learning of change-aware representation to generate finer change maps \cite{guo2021deep,chen2021remote, Hong2023SpectralGPTSR} or change captions \cite{10305516}. 
In summary, current research based on CNNs and transformers focuses on improving multiscale feature learning \cite{liu2023attention} and contextual information modeling \cite{liu2022cnn} to capture more discriminative representations and robust spatio-temporal relationships from bi-temporal image data \cite{10419228}. 

However, most approaches regard RS-CD as a pixel-wise labeling task, which assigns a changed or unchanged status for each pixel through the pixel-level encoder with a Sofmax or Sigmoid layer \cite{10419228, fang2021snunet}. This paradigm suffers from several issues in the change maps, including discontinuous detected objects, speckle noises, and omissions within the homogeneous objects. Although object or superpixel-based deep learning approaches have drawn research interest in CD,  they utilize off-the-shelf segmentation algorithms to obtain certain outlines of objects as constraints for the generation of change maps \cite{zheng2021building,liu2021change}. The masking process is not learnable, which could hinder the networks' representation capability and the quality of final CD results. Furthermore, the vision foundation model, like the segment anything model (SAM), has been applied in CD to generate object segments \cite{ding2024adapting,wang2023cs}. Unfortunately, model fine-tuning on RS images is still needed because SAM is pretrained on natural images and its performance degrades on RS images \cite{ding2024adapting}.

Inspired by the successful exploitation of mask classification-based segmentation, we propose a mask classification-based CD (MaskCD) approach that divides the prediction of change maps into two separate processes of generating a series of object masks and their corresponding changed or unchanged labels, respectively. In particular, 
a cross-level change representation perceiver (CLCRP) based on
a deformable multi-head self-attention (DeformMHSA) is employed to reform a multi-level pixel-level embedding for change-aware representation learning from the spatio-temporal features extracted by a transformer-based Siamese backbone. After that, the change-aware representations are forwarded into masked attention-based detection transformer decoder (MA-DETR) modules that progressively predict several segmented masks with the changed or unchanged class at multiple scales, which can be combined into change maps. Moreover, the MA-DETR module incorporates a masked-attention mechanism to alleviate the background noises in the change-aware representations. As shown in Fig. \ref{fig:comparison}, object masks and their categories are adaptively learned through a mask and DETR decoder, respectively, in our approach, which could alleviate background noises and losses of details of land cover objects compared with the approach using a per-pixel classifier in the decoder. 

%It exploits a multi-head self-attention mechanism to capture the global dependencies within the deep features.

The main contributions of this study are summarized as follows:
\begin{itemize}
    \item We propose a novel mask classification-based CD model, MaskCD, which introduces a paradigm shift from per-pixel classification to identify changes in the images as individual objects in an end-to-end manner. 
    \item  A cross-level change representation perceiver (CLCRP) based on DeformMHSA is introduced to generate multi-level change-aware representation while enhancing the modeling of spatiotemporal relations from bi-temporal deep features extracted by a Swin transformer-based backbone.
    \item We propose a masked attention-based transformer decoder, MA-DETR, to locate various changed objects. It can significantly restrain the background noises while retrieving the desired objects from all the feature elements.
    \item Experimental results on five benchmark datasets indicate that our MaskCD achieves more competitive performance in various scenes while retaining accurate object outlines compared with eight state-of-the-art models.
\end{itemize}

The rest of this article is organized as follows: In Section \ref{s2}, a brief review is given regarding the CNN-based CD methods, attention-augmented learning, and mask classification. Section \ref{s3} presents the details of the proposed MaskCD method. Section \ref{s4} provides the experimental results along with the ablation studies. Finally, Section \ref{s5} concludes this article.

\section{Related Work}\label{s2}
\subsection{CNN-Based RS-CD Methods}
In the literature, most CNN-based CD approaches employ encoder-decoder-based networks, consisting of a Siamese convolutional encoder to extract deep representations from bi-temporal images and a decoder to obtain a change map from change-aware representation with a pixel-wise classifier. 
%Specifically, the encoder typically adopts a Siamese convolutional neural network (CNN), accepting bi-temporal images from two branches. This design enables the encoder to adeptly capture both spatial and temporal dependencies, encoding a change-aware representation through strategies such as feature difference and feature stacking. Following this, the decoder processes the change-aware representation by employing upsampling and transformed convolutional layers to reconstruct the spatial information, ultimately generating an output map. The final layer of the decoder typically functions as a change classifier, utilizing a Sigmoid or Softmax activation, thereby producing a probability for each pixel to signify whether a change has occurred. 
For example, \citet{zhang2020feature} proposed a deep feature difference CNN composing a two-channel network with shared weight to generate a multi-scale and multi-depth feature difference map for CD. Leveraging memory networks in encoding temporal dependencies, \citet{mou2018learning} proposed a network combining recurrent neural network and CNN to learn a joint spectral-spatial-temporal feature representation in a unified framework. 
On this basis, many works focus on multiscale feature learning strategies such as deep supervision \cite{zhang2020deeply,shi2021deeply} and dense connection \cite{fang2021snunet} to capture rich contextual information. Since the aim of CD is to identify differences from multitemporal images, cross-scale feature interaction and fusion methods have been explored to obtain change-aware representations \cite{feng2023change,feng2022icif,10129139}.
Furthermore, attentions were developed to recalibrate and integrate the multiscale features for enhancement \cite{10093022,lei2023ultralightweight, Li_2023_A2Net}
% DMINet\cite{feng2023change}
% A2Net\cite{Li_2023_A2Net}
% DSAMNet\cite{shi2021deeply}
% USSFC-Net\cite{lei2023ultralightweight}
% ICIF-Net\cite{feng2022icif}
% RDP-Net\cite{chen2022rdp}
% HANet-CD\cite{10093022}
% DASNet\cite{chen2020dasnet}
% changer\cite{10129139}
%an adversarial training has been integrated with encoder-decoder-based networks to construct a generative adversarial network (GAN) to address the domain shift problem for heterogeneous RS images in change detection missions. For a more comprehensive overview of related works, interested readers can consult a recent review.

Previous approaches treat CD as a pixel-wise labeling task involving the assignment of changed or unchanged labels to individual pixels on a change map. Although they attempt to capture various changed objects through multilevel feature learning, the per-pixel classification approach often generates results with fragmented object boundaries and isolated noises because the convolution operation and the decision process are conducted on individual pixels.

\subsection{Transformer-based RS-CD Methods}
To mitigate the limitations of receptive fields in CNNs, recent transformer models based on self-attention (SA) are adopted for CD, which can capture the long-range contextual relationships across the deep feature elements. For example, \citet{bandara2022transformer} proposed a hierarchical transformer encoder for feature learning. \citet{9736956} developed a pure transformer network with a Siamese U-shaped network using the Swin Transformer as the basic unit. To benefit from both advantages of the CNN and transformer models, \citet{liu2023attention} developed an attention-based multiscale transformer network (AMTNet) that combines the strengths of CNNs, transformers and attention mechanisms.  In \cite{chen2021remote} and \cite{bandara2022transformer}, the transformer has been used to tokenize patch-wise features obtained from CNNs to extract rich global context information.
%In our previous work \cite{zhang2022multilevel}, cross-level deformable self-attention was developed to learn multi-scale change-aware representations. 

Despite the significant success of context modeling in these approaches, they still may struggle to retain local and fine-grained details of objects when modeling their long-range dependencies through transformers.
Moreover, high computational costs of self-attention hinder cross-scale representation learning in transformers.

\subsection{Object-based CD Methods}
Object-oriented CD methods are advanced techniques used in RS and image analysis to identify and characterize changes in the Earth's surface over time. Unlike pixel-based approaches, object-based methods consider the spatial arrangement of pixels and group them into meaningful objects or segments \cite{hussain2013change}, which are further used as the basic unit for developing a CD strategy.   The core idea involves extracting relevant features from the segmented objects, such as shape, size, texture, and spectral information, and then applying machine learning algorithms to identify changes. By focusing on objects rather than individual pixels, these methods can delineate landscape features at different levels and reduce small spurious changes \cite{lv2022object,xiao2016change}.
Recently, object-level deep learning-based CD methods have been developed. For example, \citet{zheng2021building} combined object-based CD with a patch-based CNN, in which a deep object localization network was developed for instance segmentation, and an object-based voting method was applied for post-processing.
In \cite{liu2021change}, superpixel segmentation algorithms were applied to generate objects before bi-temporal image patches containing the segmented objects were fed into the CNN. 
Recently, SAM, as a popular vision foundation model, has been introduced to generate bitemporal segments for CD, while a fine-tuning strategy is normally required to address the failed segmentation of small and irregular objects in complex RS images \cite{ding2024adapting}.

 % \textcolor{red}{Superpixel sampling network was embedded into change detection networks for superpixel segmentation of bitemporal images \cite{zhang2021escnet}. }
 
Unfortunately, these deep-learning methods rely on pre-generated outlines of objects and can not obtain desirable object-level CD results in a one-step and end-to-end fashion.
Furthermore, the masking process using certain off-the-shelf segmentation algorithms could not fully exploit the representation capability of networks.
%enhance the ability to distinguish between real changes and random variations, providing more accurate and contextually meaningful results. In particular, these methods usually adopt a two-step change detection process, which first generates an object-based segment mask for the image and then distinguishes the change status of each segment. 

\subsection{Mask-Based Vision Tasks}
% (looked once, twice)
Recent works have explored the effectiveness of mask classification strategy in semantic segmentation \citep{cheng2021per,cheng2022masked,li2022maskformer}, instance segmentation \cite{zhong2023multi}, and cloud detection \cite{zhang2022cloudformer}. The core idea is to predict a set of binary masks associated with object instances. However, its potential is still under-explored in CD. Compared to other vision tasks, generating and predicting masks in CD is more challenging. This is because the features of unchanged objects can be temporally variant, resulting in pseudo-changes. Furthermore, in some cases, only specific parts of land cover objects change, with no evident outlines.

\begin{figure}[htp]
\subfloat[Self-attention\label{fig:sa}]{%
  \includegraphics[clip,width=\columnwidth]{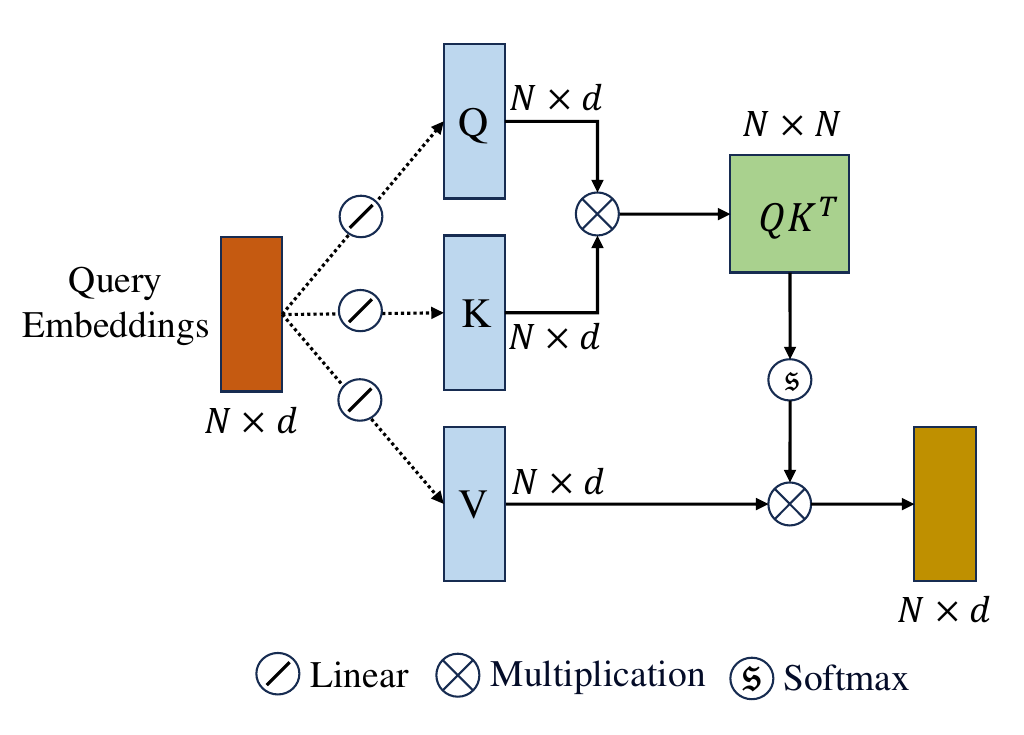}%
}

\subfloat[Masked Attention\label{fig:ma}]{%
  \includegraphics[clip,width=\columnwidth]{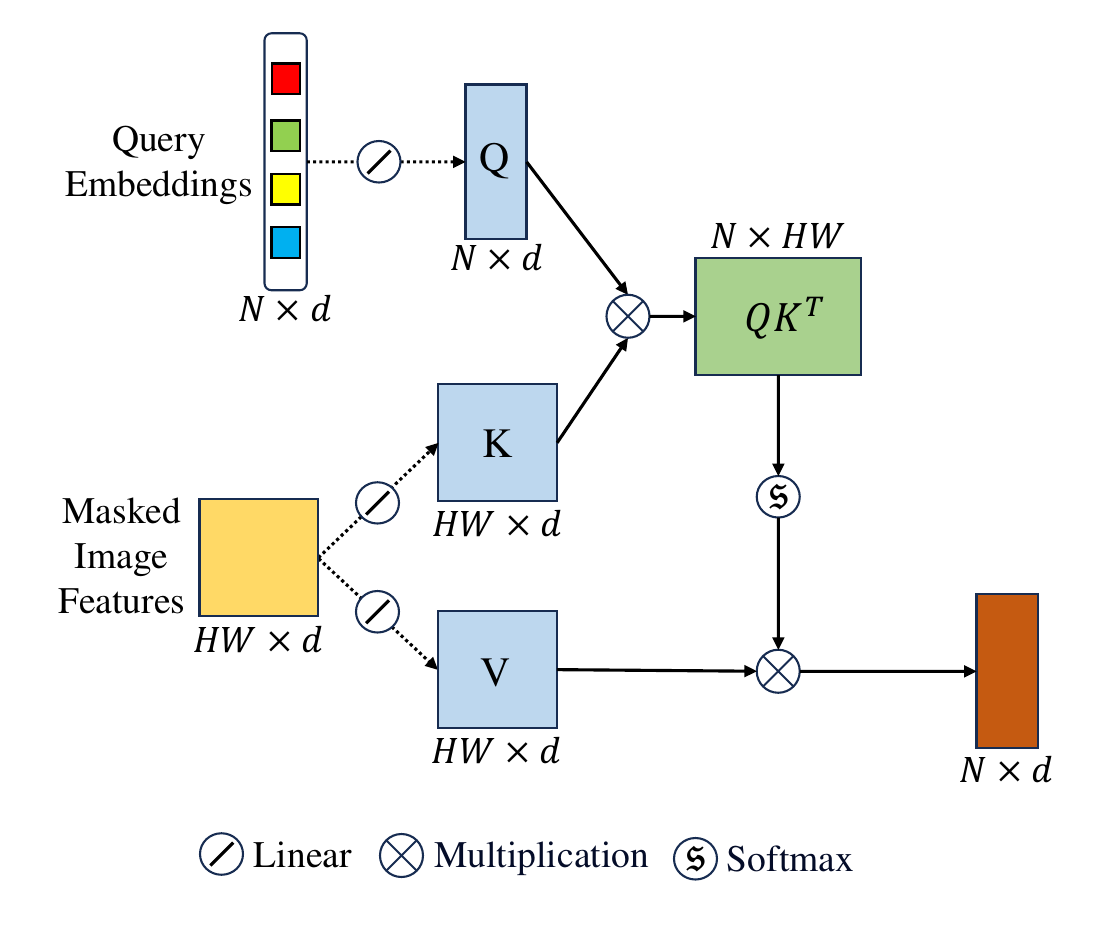}%
}
\caption{Illustration of (a) self-attention and (b) masked attention mechanism.}
\label{fig:attention}
\end{figure}
%Although CNN-based change detection (CD) methods yield promising results due to their robust change-aware representation capabilities, they face limitations stemming from the inductive bias associated with local receptive fields in convolutional layers. Consequently, these methods encounter challenges in effectively capturing dependencies across channels and spatial dimensions within deep features.

% the change-aware representations may still contain a large amount of background information, which can 
% introduce distractions for detecting the changed objects.

This motivates us to develop robust mask attention mechanisms and mask classification-based CD methods, which can detect and output changed objects adaptively. 
As shown in Fig. \ref{fig:ma}, the proposed enhanced masked attention approach selectively focuses on the foreground of the representations, compared to the original self-attention mechanism that calculated full-image dependencies between any two elements in a feature map shown in Fig.~\ref{fig:sa}. In particular, the localization of attention around the segments aggregates the most informative features that lead to a better formulation of query embeddings by emphasizing the most critical objects in the full image, leading to a more effective convergence.

\section{Methodology}\label{s3}
\begin{figure*}[htp]
    \captionsetup{singlelinecheck=false}
    \centering
    \includegraphics[scale=0.35]{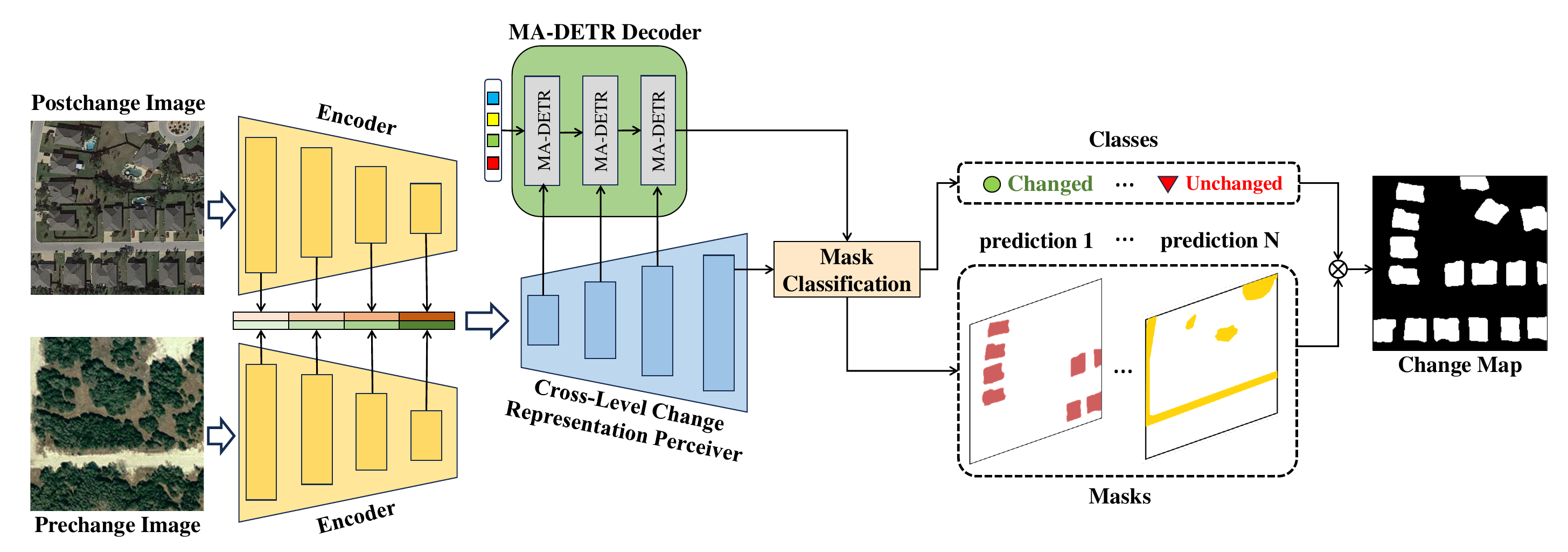}
    \caption{Proposed MaskCD framework for bi-temporal CD.}
    \label{fig:maskcd}
\end{figure*}

This section provides a detailed description of the proposed CD framework, MaskCD, as shown in Fig. \ref{fig:maskcd}. First, multi-level deep features are extracted by modeling long-range dependencies through the hierarchical Transformer-based Siamese encoder. Then, the multi-level features are fed into cross-level change representation perceiver for learning change-aware representations. After that, a masked-attention-based decoder is constructed to obtain per-segment embeddings as foundations for generating mask embeddings and the class labels for the masks. Consequently, the mask classification module is built to generate a series of masks associated with their class labels from per-segment and per-pixel embeddings. Finally, a dual loss function is designed to optimize the masks and their class labels in the training process, respectively, while they can be directly multiplied into the change maps in the inference stage.
\subsection{Hierarchical Transformer-based Siamese Encoder}
In this study, we employ a hierarchical transformer-based Siamese encoder to obtain multi-level deep representations from bitemporal images. It adopts a parameter-shared Siamese architecture, with each branch containing multi-stages of transformer modules to progressively extract comparable deep features from the bitemporal images simultaneously in the same feature space. Let $\{f_{l}^{T_{1}}\}^{4}_{l=1}$ and $\{f_{l}^{T_{2}}\}^{4}_{l=1}$ be the feature maps at $4$ levels extracted from the pre- and post-change images, respectively. %The feature maps $f_{l}^{T_{1}}$ and $f_{l}^{T_{2}}$ extracted by the $l$-th level transformer module have the size of $H_{l}\times W_{l}$. 
In contrast to directly using the feature differences between bitemporal deep feature maps for change information, we utilize a combination of the bitemporal feature maps $\delta^{l}$ by channel-wise concatenation to preserve the spatio-temporal association between the bitemporal images, i.e., $\{\delta_{l}\}^{4}_{l=1}=\{[f_{l}^{T_{1}};f_{l}^{T_{2}}]\}_{l=1}^{4}$.

Motivated by the great success of Swin-Transformer models applied in many recent RS approaches, we incorporate Swin-Transformer blocks to improve the understanding of complex spectral characteristics of multitemporal RS images. The Swin-Transformer blocks are built based on a window-shifted self-attention mechanism that can capture intricate patterns and spatial relationships in the extracted deep features. To alleviate the high demand for computational resources in common element-wise self-attention algorithms, the Swin-Transformer adopts a window-attention strategy that divides the images into several patches by non-overlapping windows and computes self-attention locally within the windows \cite{liu2021swin}. This strategy can be defined as follows:
\begin{equation}
	\text{Attention}(Q,K,V) = \text{SoftMax}\left(\frac{QK^{T}}{\sqrt{d}} + B\right)V,
\end{equation}
where $Q$, $K$, and $V$ denote the query, key, and value matrices, $d$ is the dimension of the query and key and $B \in \mathbb{R}^{M^{2} \times M^{2}}$ is the relative position bias, where $M^{2}$ indicates the number of patches in a window. To enable interaction between different divided windows, a shifted-window self-attention mechanism is adopted to introduce cross-window connections, alternating the partitioning configuration of the windows between successive Swin-Transformer blocks at the same stage. Consequently, the bitemporal representations at multiple stages of resolutions and abstractions can be obtained through various stages of stacked Swin-Transformer blocks. The shifted window attention integrated into successive Swin-Transformer blocks can be computed as:
\begin{align}
	&\hat{z}_{l}=\mathrm{W\_MSA}(\mathrm{LN}({z}_{l-1}))+{z}_{l-1}, \\
	&z_{l}=\mathrm{MLP}(\mathrm{LN}(\hat{z}_{l}))+\hat{z}_{l}, \\
	&\hat{z}_{l+1}=\mathrm{SW\_MSA}(\mathrm{LN}({z}_{l}))+{z}_{l},\\
	&z_{l+1}=\mathrm{MLP}(\mathrm{LN}(\hat{z}_{l+1}))+\hat{z}_{l+1},
\end{align}
where $\mathrm{W\_MSA}$ and $\mathrm{SW\_MSA}$ denote the window-based multi-head self-attention using regular and shifted window partitioning configurations, respectively. Moreover, $\hat{z}_{l}$ and $z_{l}$ denote the output of (S)W\_MSA module and the MLP layer of the $l$-th Swin-Transformer block, respectively. LN represents the layer normalization module that can accelerate the convergence of the model with a more stable training process.

\subsection{Cross-Level Change Representation Perceiver} \label{pixelwise}
To obtain change-aware representations from the concatenated bi-temporal deep features, we introduce a novel Cross-Level Change Representation Perceiver (CLCRP) built upon a Deformable Multi-Head Self-Attention (DeformMHSA) encoder. Drawing inspiration from recent advancements in deformable attention-aggregated models within RS-CD methodologies \cite{zhang2022multilevel,zuo2021deformable}, we leverage the DeformMHSA encoder to efficiently capture the most salient information from the multi-temporal deep features. Subsequently, the attention-aggregated features undergo decoding via an additional convolutional layer, enriching spatial features with residual multi-temporal deep features extracted by the DeformMHSA encoder.
\begin{figure*}[h]
    \captionsetup{singlelinecheck=false}
    \centering
    \includegraphics[width=0.85\linewidth]{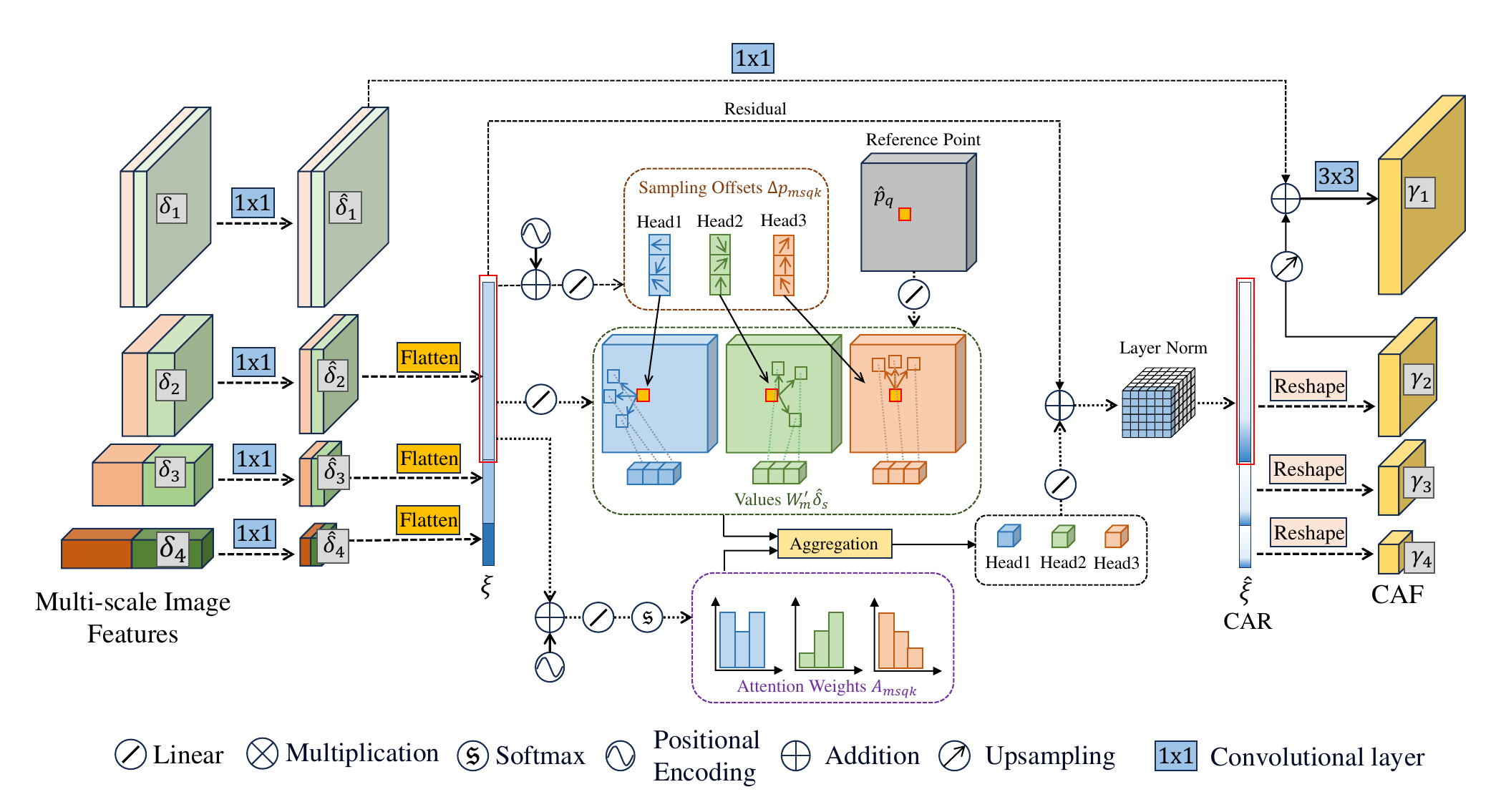}
    \caption{Proposed Cross-Level Change Representation Perceiver.}
    \label{fig:deformmhsa}
\end{figure*}

As shown in Fig. \ref{fig:deformmhsa}, the pixel-level decoder takes the input of deep feature maps $\{\delta_{s}\}_{s=1}^{4}$ to reconstruct multi-level change-aware pixel-wise embeddings. 
In the DeformMHSA encoder, we adopt a similar strategy as \cite{zhang2022multilevel} to build a sequence-to-sequence encoding that combines the multi-level deep features into a feature sequence for the processing of DeformMHSA. In particular, the multi-level deep features $\{\delta_{l}\}_{l=1}^{4}$ are transformed into same hidden dimension $d=256$ with $1\times 1$ convolutional layers. Notably, the first level of deep features $\delta_{1}$ is preserved through a residual connection to preserve high-resolution spatial information, while the other three levels of deep features $\{\delta_{l}\}_{l=2}^{4}$ are flattened into multi-level sequences $\{\hat{\delta_{l}}\}_{l=2}^{4}$, where $\hat{\delta_{l}} \in \mathbb{R}^{C \times(H_{l} W_{l})}$.

The flattened multi-level sequences are then concatenated into a new feature sequence $\xi$ providing multi-scale information, i.e., $\xi=[\hat{\delta_{2}};\hat{\delta_{3}};\hat{\delta_{4}}]$, where $\xi \in \mathbb{R}^{\sum_{l=2}^{4} d\times(H_{l}W_{l})}$. As the 2-D spatial information is no longer available in the flattened sequence $\xi$, we supplement each query element with a unique positional encoding based on its level and coordinate in the original features before passing the sequence to the DeformMHSA encoder. Specifically, sine and cosine functions are used to generate the positional encodings. Let $(x_{l},y_{l})$ indexes a 2-D spatial coordinate of the original feature $\hat{\delta_{l}}$ at the $l$-th level, the coordinate $(x_{l},y_{l})$ is normalized to the range (0,2$\pi$). Consequently, sine and cosine functions are independently applied to generate positional encodings $\mathrm{PE}_{l}$ at the $l$-th level along the $x$- and $y$-axes as follows:
\begin{equation}\label{eqn:pos_embeds}
	\mathrm{PE}_{l}^{\phi}=\left\{
	\begin{aligned}
		&PE_{l}(\phi_{l},2i)=cos(\phi/10000^{2i/(C/2)}), \\
		&PE_{l}(\phi_{l},2i+1)=sin(\phi/10000^{2i/(C/2)}),
	\end{aligned}
	\right.
\end{equation}
where as $\phi_{l}=(\phi_{l}^{x},\phi_{l}^{y})$ indicates the coordinate of a query feature at the $l$-th level, and $i \in [0,...,C/4]$ indexes the channel dimension of the features. For a feature sequence with $C$ dimensions, sine and cosine functions with different frequencies correspond to encodings with $C/2$ dimensions. As a result, the sine and cosine curves can be delivered within one circle, ensuring no duplicated values for each spatial location of different dimensions $i$.

Instead of constructing a common global self-attention computation for feature element at each spatial location, the DeformMHSA only attends to a small set of key sampling points around a reference point by generating several sampling offsets, as illustrated in Fig. \ref{fig:deformmhsa}. Let $q$ indexes a query element with representation feature $z_{q}$, and $\hat{p}_{q}\in[0,1]^{2}$ be the normalized coordinates of the reference point for each query element $q$, respectively. The DeformMHSA encoder calculates the multi-scale attentive features as follows:
\begin{align}
	&\mathrm{DeformMHSA}(z_{q}, \hat{p}_{q}, \{\hat{\delta_{l}}\}_{l=2}^{4})= \notag \\
	&\sum_{m=1}^{M}W_{m}[\sum_{l=2}^{4}\sum_{k=1}^{K}A_{mlqk}\cdot W_{m}^{'}\hat{\delta_{l}}(\varphi_{l}(\hat{p}_{q})+\bigtriangleup p_{mlqk})],
\end{align}
where $m$ and $k$ index the attention head and sampled keys, respectively. $K$ represents the total amount of the sampled keys and $K\ll HW$. $W$ and $W^{'}$ are linear projections with learnable weights. $\bigtriangleup p_{mlqk}$ and $A_{mlqk}$ denote the sampling offset and attention weight of the $k$-th sampling point in the $m$-th attention head of the DeformMHSA encoder, respectively. The DeformMHSA encoder delivers a sequence of multi-level change-aware representations $\hat{\zeta} \in \mathbb{R}^{\sum_{l=2}^{4} d\times(H_{l}W_{l})}$ with the same shape of the input $\zeta$. $\hat{\zeta}$  is then split and reshaped into multi-level 2-D change-aware feature maps${\{\gamma_{l}\}}_{l=2}^{4}$ according to the shapes of the input multi-level features ${\{\hat{\delta_{l}}\}}_{l=2}^{4}$. 

To recover the feature map resolution for pixel-wise semantic information, the encoded features with the highest resolution $\gamma_{2}$ are upsampled and added with the residual feature map $\hat{\delta}_{1}$ before they are fused with a $3\times3$ convolutional layer into $\gamma_{1} \in \mathbb{R}^{d\times(H_{1}W_{1})}$. To this end, the multi-scale per-pixel embeddings $\{\gamma_{l}\}_{l=1}^{4}$ are gained.

\subsection{Masked Attention-Based Detection Transformer Decoder}
\begin{figure*}[h]
    \captionsetup{singlelinecheck=false}
    \centering
    \includegraphics[scale=0.3]{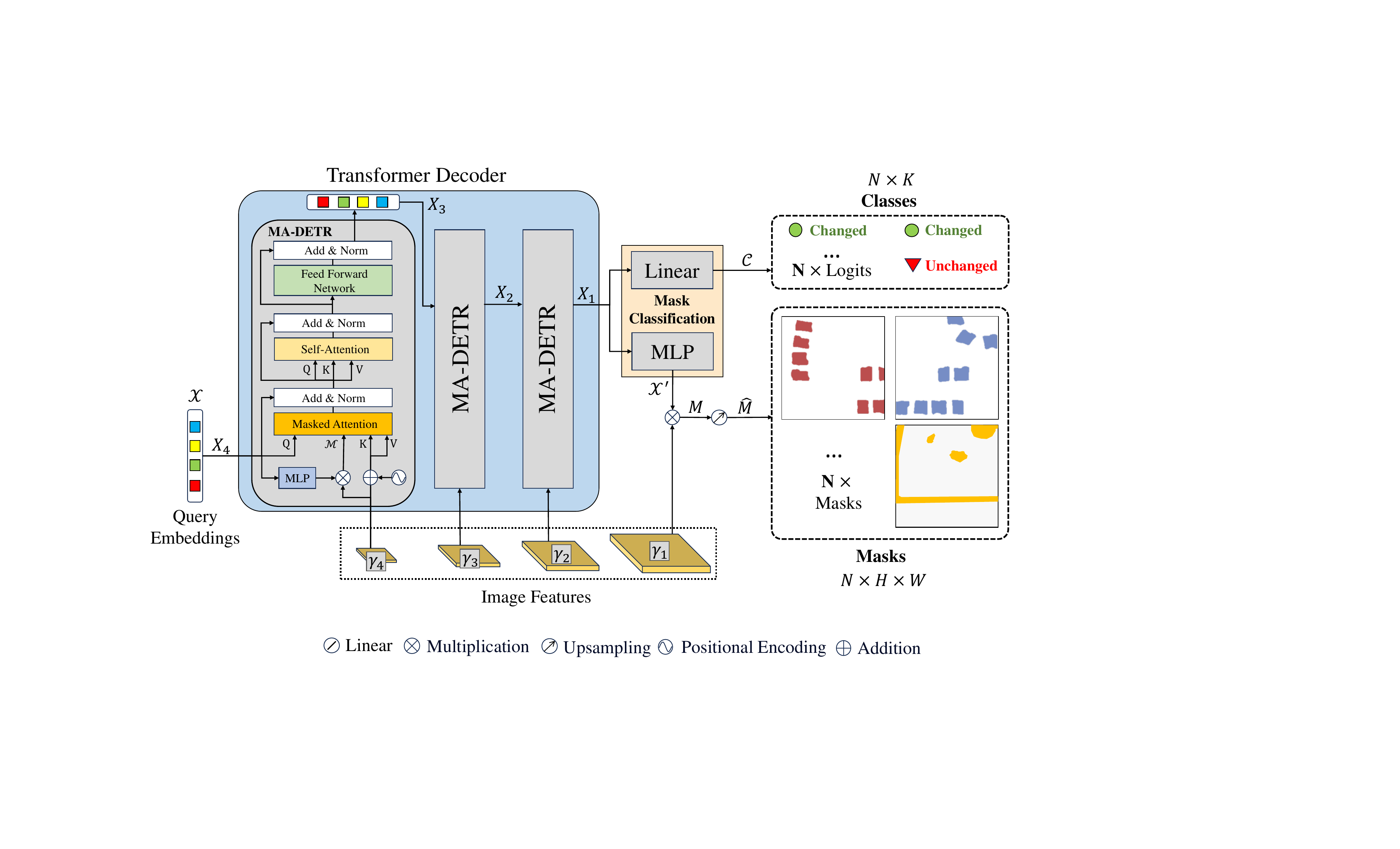}
    \caption{Proposed transformer decoder and mask classification module for bi-temporal CD.}
    \label{fig:madetr}
\end{figure*}

In the literature, most of the current CD approaches utilized a pixel-wise decoder to directly predict the change map from the deep feature representations. However, CD based on pixel-by-pixel classification usually suffers from a noncontinuous detection of the changed objects due to the perturbations of the background noises. Therefore, we propose a masked attention-based detection transformer (MA-DETR) encoder that considers the objects in the images as objects rather than predicting them from discrete pixels. As shown in Fig. \ref{fig:madetr}, the decoder consists of $3$ stages of MA-DETR modules that accept multi-scale per-pixel embeddings $\{{\gamma_l\}}_{l=2}^{4}$ as input, where $\gamma_{1}$ with highest resolution is retained for decoding in mask classification module. Simultaneously, object queries $\mathcal{X} \in \mathbb{R}^{N\times d}$ are initialized as learnable embeddings and decoded by the MA-DETR into the object maks and their corresponding change categories, where $N$ denotes the numbers of detected objects and $d$ is the hidden dimension. The key components of each MA-DETR module include an MLP module as the attention mask predictor, a masked attention block, and a self-attention block.
\subsubsection{Masked Attention Block}
The masked attention block exploits object queries to perform a masked cross-attention and extract relations from per-pixel image embeddings. 
In contrast to using the cross-attention mechanism in the standard detection transformer decoder that involves the full global context, we adopt the masked attention blocks that restrict the attention to localized features centered around predicted segments in the attention masks, leading to faster convergence and improved performance. Specifically, a multiple-layer perceptron (MLP) is employed to enable the interaction of query features and per-pixel embeddings from each level to generate attention masks $\{M_{l}\}_{l=2}^{4}$ as the localized guidance of each query embedding for the consequent masked attention block, where $M_{l} \in \{0,1\}^{N\times H_{l}W_{l}}$. For the computational needs, the binary masks at each level are transformed to masks $\{\mathcal{M}_{l}\}_{l=2}^{4}$ as follows:
\begin{equation}
	\mathcal{M}_{l}(x,y)=
	\begin{cases}
		0 &\mathrm{if} \ M_{l}(x,y)=1\\
		-\infty &\mathrm{otherwise}
	\end{cases},
\end{equation}
where $\mathcal{M}_{l} \in \mathbb{R}^{N\times H_{l}W_{l}}$. After that, the masked attention $ X_{l}^{Mask}$ can be computed as follows:
\begin{equation}
	X_{l}^{Mask}=\mathrm{Norm}(\mathrm{softmax}(\mathcal{M}+Q_{l}K_{l}^{T})V_{l}+X_{l}),
\end{equation}
where $Q_{l}\in\mathbb{R}^{N\times C}$ denotes the query feature that is the query embeddings $X_{l}$ under linear transformation.
$K_{l}$, $V_{l} \in \mathbb{R}^{C\times H_{l}W_{l}}$ are the key and query elements that are linearly transformed per-pixel embeddings $\gamma_{l}$ with positional embeddings generated by Eqn. \ref{eqn:pos_embeds}. 
The masked attention will degenerate to the cross attention when $M_{l}=1$. As a result, the masked attention restricts the computation only to the masked locations with enhanced features to improve the detection of CD objects by ignoring the background noises and concentrating only on foreground objects.
\subsubsection{Self-Attention Block}
The self-attention block takes the input of the attentive query features $X_{l}^{'}$ and utilizes the self-attention mechanism to gather the context information by interacting with the queries. Compared with the masked attention mechanism, the self-attention block only calculates the global dependencies within the object queries. Therefore, the complexity of the self-attention module is limited to a moderate number of object queries, which is more acceptable than the computation of all the elements of the per-pixel embeddings. The standard self-attention block can be computed as follows:
\begin{equation}
	X_{l-1}=\mathrm{Norm}(\mathrm{softmax}(\frac{QK^{T}}{\sqrt{\varepsilon}})V+X_{l}^{Mask}),
\end{equation}
where $Q$, $K$, $V$ $\in \mathbb{R}^{N\times \varepsilon}$ denote the query, key, and value elements of the self-attention mechanism. $X_{l-1}$ is the output object query and will be passed to the next stage masked attention block. At the final stage of the MA-DETR module, $X_{1}$ will be obtained as the per-segment embeddings for the following mask classification module. 
\subsection{Mask Classification module}
The mask classification module takes the input of the per-segment embeddings $X_{1}$ and the per-pixel embeddings $\gamma_{1}$ to generate a series of binary masks associated with their classes. In particular, the mask classification module consists of a linear classification network and an MLP module that classify $X_{1}$ into a series of classes $\mathcal{C} \in \mathbb{R}^{N\times 2}$ and mask embeddings $\mathcal{X}^{Mask}\in \mathbb{R}^{N\times d}$, respectively, for binary CD. Consequently, the mask embeddings $\mathcal{X}^{Mask}$ are multiplied with the preserved per-pixel embeddings $\gamma_{1} \in \mathbb{R}^{d\times H_{1}\times W_{1}}$ to generate a series of masks $M \in \mathbb{R}^{N\times H_{1}\times W_{1}}$. To retrieve the complete resolution, the masks $M$ are upsampled into $\hat{M}$ with a scale of $4$ according to the height and width of the input RS images.
\subsection{Optimization}
In traditional CD approaches, pixel-wise loss functions are usually adopted for training objectives that directly measure the pixel-wise differences between predictions and the ground truth. However, pixel-wise optimization strategy cannot be utilized to train the MaskCD model, as the predicted masks and their categories cannot be directly matched to the ground truths. Therefore, we propose an optimization function that produces an optimal bipartite matching between predictions and the ground truths and then optimizes the losses of masks and their categories, respectively.
\subsubsection{Bipartite Matching}
For the pixel-wise CD task, a trivially fixed matching is possible if the number of predictions $N$ exactly matches the number of categories $K$. In this case, the $i$-th prediction is matched to a ground truth region with class label $i$ and to $\varnothing$ if a region with class label $i$ is not present in the ground truth. However, it still remains limited to match both the masks and their classes of the predicted objects with the ground truths at the same time due to the complexity of the change maps. Therefore, optimal bipartite matching is introduced to search for permutation of $N$ elements with the lowest cost, while optimal matching is calculated using a Hungarian algorithm that takes both masks and their classes into consideration. Let $\{m_{i}\}_{i=1}^{N}$ denote the mask predictions $M$ of the last stage, and $p_{i}\in \triangle^{K}$ represents the probability distribution of the $m_{i}$, then the outputs as a set of probability-mask pairs can be defined as $z=\{(p_{i},m_{i})\}_{i=1}^{N}$. Moreover, let $z^{gt}=\{(c_{i}^{gt},m_{i}^{gt})|c_{i}^{gt} \in \{1,...,K\},m_{i}^{gt}\in {\{0,1\}}^{H\times W}\}$ indicates the ground truth, the matching cost between a pair of prediction and ground truth can be defined as follows:
\begin{equation}
	\mathcal{L}_{\mathrm{match}}(z,z^{gt})=-p_{i}(c_{j}^{gt})+\mathcal{L}_{mask}(m_{i},m_{j}^{gt}),
\end{equation}
where $\mathcal{L}_{mask}$ is a binary mask loss. After that, a permutation $\hat{\sigma}$ for $N$ elements  with the lowest cost can be obtained as:
\begin{equation}
	\hat{\sigma}=\underset{\sigma \in \mathfrak{S}}{\arg\min} \sum_{i}^{N}\mathcal{L}_{match}(z_{i},z^{gt}_{\sigma(i)}).
\end{equation}
where the optimal assignment is computed efficiently with the Hungarian algorithm.
\subsubsection{Loss Functions}
A hybrid multi-scale loss function $L_{MaskCD}$ is designed for the training of MaskCD, including a main loss function $\mathcal{L}_{main}$ and an auxiliary loss function $\mathcal{L}_{aux}$, which intend to minimize the error of final prediction and multi-scale intermediate features, respectively.
In particular, the main loss function $\mathcal{L}_{main}$ is composed of a cross-entropy classification loss $\mathcal{L}_{cls}$ and a binary mask loss $\mathcal{L}_{mask}$ for each segment $j$, which can be defined as follows:
\begin{align}
	&\mathcal{L}_{main}(z,z^{gt})=  \lambda_{cls}\mathcal{L}_{cls}+\lambda_{m}\mathcal{L}_{mask}, \\
	&\mathcal{L}_{cls}= \sum_{j=1}^{N}[-\log p_{\hat{\sigma}(j)}(c_{j}^{gt}),\\
	&\mathcal{L}_{mask}=\mathbbm{1}_{c_{j}^{gt}\neq\varnothing} \mathcal{L}_{mask}(m_{\hat{\sigma}(j)},m_{j}^{gt})],
\end{align}
where $\lambda_{cls}$ and $\lambda_{m}$ are adjustable weights.
As the mask predictions are of binary values, the mask loss $\mathcal{L}_{mask}$ is a linear combination of a binary cross-entropy (BCE) loss and a dice loss multiplied by hyper-parameters $\lambda_{b}$ and $\lambda_{d}$ respectively, i.e., $\mathcal{L}_{mask}=\lambda_{b}\mathcal{L}_{bce}+\lambda_{d}\mathcal{L}_{dice}$.

To further introduce the abstract information of the predicted masks through each MA-DETR decoder layer, a multi-level auxiliary loss $\mathcal{L}_{aux}$ is developed that has the same formation as the main loss but accepts the multi-level intermediate features of the decoder layers. Finally, the complete loss function can be calculated as $\mathcal{L}_{MaskCD}=\mathcal{L}_{main}+\mathcal{L}_{aux}$.

\section{Experimental Results}\label{s4}
\subsection{Dataset Description} In this work, five publicly available datasets are employed to evaluate the proposed MaskCD network concerning various types of changed targets, such as cropland, buildings, landslides, and urban objects at several spatial scales. The samples of the datasets are cropped into image pairs of size $256 \times 256$ without overlapping for the convenience of a mini-batch learning strategy and randomly split into subsets of training, validation, and testing for our experiments. The detailed information and implementation of these datasets can be found via \href{https://huggingface.co/ericyu}{HuggingfaceHub}.
\subsubsection{\textbf{CLCD}} The CropLand change detection (CLCD) dataset \cite{liu2022cnn} includes 600 bi-temporal agricultural image pairs of size $512\times 512$ pixels with spatial resolutions ranging from 0.5 to 2 m, which were collected by Gaofen-2 satellite from the Guangdong Province, China, in 2017 and 2019, respectively. The main types of change annotated in CLCD include buildings, roads, lakes, and bare soil lands.
\subsubsection{\textbf{EGY-BCD}} The Egypt building change detection (EGY-BCD) dataset \cite{holail2023afde} consists of 6091 pairs of Google Earth images of size $256\times256$ with a resolution of 0.25 m captured in Egypt over two different periods between 2015 and 2022. 
\subsubsection{\textbf{GVLM-CD}} The Global Very High-Resolution Landslide Mapping (GVLM) Dataset \cite{zhang20231cross} contains a total of $17$ image pairs  wth a spatial resolution of $0.59$ m acquired via Google Earth service. They were cropped into $7496$ pairs of patches with a size of $256 \times 256$. The landslide maps representing a natural change in the land cover exhibit noticeable variations in size and shape of the change maps. 
\subsubsection{\textbf{LEVIR-CD}} This building CD dataset \cite{chen2020spatial} contains 637 high-resolution RGB image pairs of the size of $1024 \times 1024$ with spatial resolution $0.5$ m captured from 2002 to 2018 in several urban areas of Texas, USA. The LEVIR-CD dataset covers various types of buildings, such as villa residences, tall apartments, small garages, and large warehouses.
\subsubsection{\textbf{SYSU-CD}} This dataset \cite{shi2021deeply} consists of $20000$ bi-temporal aerial images of size $256 \times 256$ with the spatial resolution of 0.5m taken between the years 2007 and 2014 in Hong Kong. The dataset mainly focuses on several types of changes related to urban development, such as newly built urban buildings and suburban dilation.

\subsection{Baseline Approaches}
The proposed MaskCD is compared against several state-of-the-art deep learning-based RS-CD methods: a lightweight network with progressive aggregation and
supervised attention (A2Net) \cite{Li_2023_A2Net}, a bitemporal image transformer (BIT) network \cite{chen2021remote}, a transformer-based Siamese network for change detection (ChangeFormer) \cite{bandara2022transformer}, a dual-branch multi-level inter-temporal network (DMINet) \cite{feng2023change}, an intra-scale cross-interaction and inter-scale feature fusion network (ICIFNet) \cite{feng2022icif}, a region detail preserving network (RDPNet) \cite{chen2022rdp}, a fully convolutional Siamese difference (Siam-Diff) network \cite{daudt2018fully}, an integrated Siamese network and nested U-Net (SNUNet) \cite{fang2021snunet}. 

\subsection{Implementation Details}
The proposed MaskCD model is implemented based on the open-source \textit{Transformers} deep learning framework and initialized with a pre-trained Swin transformer backbone. For the model training, the Adam optimizer was adopted with an initial learning rate of $5e-5$, a momentum of $0.9$, and parameters $\beta_{1}, \beta_{2}$ as $0.9$ and $0.99$, respectively. The batch size was set to $16$ for each GPU.  In addition, we employ a cosine annealing scheduler that gradually reduces the learning rate to $1e-7$ for better model convergence. For the loss function, $\lambda_{cls}$, $\lambda_{m}$, $\lambda_{b}$, $\lambda_{d}$ were set to $2.0$, $1.0$, $5.0$, and $2.0$, respectively. And the amount of query embeddings $\mathcal{X}$ is set to $75$.

In our experiments, the hyper-parameters for the methods considered for comparison were set to the default values according to their original publications. 
To ensure the reproduction of the experiments, the random seed is set to $8888$ while the trained weights are publicly accessible via \href{https://huggingface.co/ericyu}{HuggingfaceHub}.
All experiments were conducted under the Slurm High-performance computing (HPC) system with a 128-core CPU and 8 NVIDIA Tesla A100 GPUs (40GB of RAM). In addition, the Accelerate \cite{accelerate} package is adopted for fully sharded data-parallel computing to speed up the computational of the models in our multi-GPU environment.

\subsection{Evaluation Metrics}
For quantitative evaluation, five commonly used metrics are employed: overall accuracy (OA), precision (Pre), recall (Rec), F1-score, and mean Intersection over Union (mIoU). The OA represents the percentage of pixels that are correctly detected among all samples. The precision is the percentage of correctly detected changed pixels among all the pixels that are identified as changed in the CD map. The recall is the number of correctly detected changed pixels divided by the number of all the pixels that should be detected as changed. The F1-score is the harmonic mean of the precision and recall that also considers the class imbalance problem in CD, and the mIoU reveals the exact coverage of the changed and unchanged area of the detected change map compared with the ground truth. They are calculated according to the following equations:
\begin{align}
\mathrm{OA}&=(\mathrm{TP}+\mathrm{TN})/(\mathrm{TP}+\mathrm{TN}+\mathrm{FP}+\mathrm{FN})\\
\mathrm{Pre}&=\mathrm{TP}/(\mathrm{TP}+\mathrm{FP}),\\
\mathrm{Rec}&=\mathrm{TP}/(\mathrm{TP}+\mathrm{FN}),\\
\mathrm{F1}&=2\mathrm{TP}/\mathrm{(2TP+FP+FN)},\\
\mathrm{mIoU}&=(\mathrm{IoU}_{C}+\mathrm{IoU}_{U})/2,
\end{align}
where
\begin{align}
    \mathrm{IoU}_{C}=\mathrm{TP}/(\mathrm{TP}+\mathrm{FP}+\mathrm{FN}),\\
    \mathrm{IoU}_{U}=\mathrm{TN}/(\mathrm{TN}+\mathrm{FP}+\mathrm{FN}),
\end{align}
where the TP is the number of correctly detected changed pixels, TN is the number of correctly detected unchanged pixels, FP is the number of false alarms, and FN is the number of missed change pixels. Moreover, $\mathrm{IoU}_{C}$ and $\mathrm{IoU}_{U}$ denote the Intersection over Union for the changed and unchanged areas, respectively.
\subsection{Comparison with the state-of-the-art}
\subsubsection{Quantitative Comparisons}
Table \ref{allinonetb} shows the quantitative evaluation comparisons of different methods in terms of OA, Pre, Rec, F1, and mIoU on five RS-CD datasets. According to the quantitative results, the proposed MaskCD method consistently outperforms the existing CD methods regarding the F1 and mIoU. In particular, the MaskCD improves $2.2\%$ and $2\%$ in terms of F1 score on both GVLM and LEVIR-CD datasets, respectively. As for the more challenging  CLCD, EGY-BCD, and SYSU-CD datasets, the MaskCD gains an enhancement of more than $6\%$, $6.36\%$, and $6.18\%$ in terms of F1 score, respectively. Consequently, MaskCD achieves the best OA values among all methods on all five datasets, which means the MaskCD can generate more accurate change maps for changed and unchanged classes. 

In addition, the proposed MaskCD achieves the best performance in terms of precision among all the experimental settings, indicating that it can significantly reduce the false alarms within the changed area predicted. Although some comparison methods achieve slightly higher recall values on GVLM and SYSU-CD datasets, they suffer from poor precision that leads to over-sensitive CD results that contain a lot of false alarms. The overall results proved the effectiveness and superiority of the proposed MaskCD method.

\begin{table*}[htp]
	\centering
	\caption{Quantitative Comparisons in terms of OA, Pre, Rec, F1, and mIoU on five RS-CD datasets. The best and second-best results are highlighted in red and blue, respectively.}
	\label{allinonetb}

		\begin{tabular}{cc|m{1.3cm}<{\centering}m{1.3cm}<{\centering}m{1.3cm}<{\centering}m{1.3cm}<{\centering}m{1.3cm}<{\centering}m{1.3cm}<{\centering}m{1.3cm}<{\centering}m{1.3cm}<{\centering}m{1.3cm}<{\centering}}
			%\specialrule{1pt}{0pt}{0pt} % thicker line
			\toprule  	
			\multicolumn{2}{c|}{Datasets} & A2Net & BIT & ChangeFormer& DMINet  & ICIFNet  & RDPNet   & Siam-Diff & SNUNet  & MaskCD \\ \midrule
			\multicolumn{1}{c|}{} & OA & 0.9204 & {\color[HTML]{0096FF} 0.9488} & 0.9431 & 0.9391 &  0.9416 & 0.9288 & 0.9358 & 0.9433 & {\color[HTML]{FF0000} \textbf{0.9666}} \\
			\multicolumn{1}{c|}{} & Pre & 0.4520 & {\color[HTML]{0096FF} 0.6525} & 0.6150 & 0.5986 & 0.6354 & 0.5195 & 0.5983 & 0.6329 & {\color[HTML]{FF0000} \textbf{0.8330}} \\
			\multicolumn{1}{c|}{} & Rec & 0.3270 & {\color[HTML]{0096FF}0.6658} & 0.6280 & 0.5518 & 0.5053 & 0.5690 & 0.4168 & 0.5662 & {\color[HTML]{FF0000} \textbf{0.6889}} \\
			\multicolumn{1}{c|}{} & F1 & 0.3794 & {\color[HTML]{0096FF} 0.6591} & 0.6214 & 0.5743 & 0.5629 & 0.5431 & 0.4913 & 0.5977 & {\color[HTML]{FF0000} \textbf{0.7541}} \\
			\multicolumn{1}{c|}{\multirow{-5}{*}{CLCD}} & mIoU & 0.5763 & {\color[HTML]{0096FF} 0.7188} & 0.6955 & 0.6697 & 0.6655 & 0.6492 & 0.6297 & 0.6835 & {\color[HTML]{FF0000} \textbf{0.7850}} \\ \midrule
			\multicolumn{1}{c|}{} & OA & 0.9623 & {\color[HTML]{0096FF} 0.9735} & 0.9651 & 0.9585 & 0.9621 & 0.9612 & 0.9524 & 0.9685 & {\color[HTML]{FF0000} \textbf{0.9824}} \\
			\multicolumn{1}{c|}{} & Pre & 0.7282 & {\color[HTML]{0096FF} 0.8016} & 0.7432 & 0.6590 & 0.7241 & 0.7125 & 0.6191 & 0.7468 & {\color[HTML]{FF0000} \textbf{0.8844}} \\
			\multicolumn{1}{c|}{} & Rec & 0.6573 & {\color[HTML]{0096FF} 0.7800 }& 0.6948 & 0.7306 & 0.6595 & 0.6612 & 0.6670 & 0.7701 & {\color[HTML]{FF0000} \textbf{0.8370}} \\
			\multicolumn{1}{c|}{} & F1 & 0.6909 & {\color[HTML]{0096FF} 0.7906 }& 0.7182 & 0.6929 & 0.6903 & 0.6859 & 0.6422 & 0.7583 & {\color[HTML]{FF0000} \textbf{0.8598}} \\
			\multicolumn{1}{c|}{\multirow{-5}{*}{EGY-BCD}} & mIoU & 0.7442 & {\color[HTML]{0096FF} 0.8130 } & 0.7619 & 0.7433 & 0.7437 & 0.7407 & 0.7116 & 0.7888 & {\color[HTML]{FF0000} \textbf{0.8678}} \\ \midrule
			\multicolumn{1}{c|}{} & OA & 0.9776 & 0.9841 & 0.9831 & 0.9825 & 0.9831 & 0.9827 & 0.9801 & {\color[HTML]{0096FF} 0.9841} & {\color[HTML]{FF0000} \textbf{0.9870}} \\
			\multicolumn{1}{c|}{} & Pre & {\color[HTML]{FF0000} \textbf{0.9156}} & 0.8975 & 0.8943 & 0.8738 & 0.8735 & 0.8751 & 0.8791 & 0.8850 & {\color[HTML]{0096FF} 0.9114} \\
			\multicolumn{1}{c|}{} & Rec & 0.7285 & 0.8571 & 0.8441 & 0.8591 & 0.8710 & 0.8610 & 0.8100 & {\color[HTML]{0096FF} 0.8728} & {\color[HTML]{FF0000} \textbf{0.8905}} \\
			\multicolumn{1}{c|}{} & F1 & 0.8114 & 0.8768 & 0.8685 & 0.8664 & 0.8722 & 0.8680 & 0.8431 & {\color[HTML]{0096FF} 0.8788} & {\color[HTML]{FF0000} \textbf{0.9008}} \\
			\multicolumn{1}{c|}{\multirow{-5}{*}{GVLM}} & mIoU & 0.8296 & 0.8819 & 0.8748 & 0.8728 & 0.8778 & 0.8742 & 0.8539 & {\color[HTML]{0096FF} 0.8835} & {\color[HTML]{FF0000} \textbf{0.9029}} \\ \midrule
			\multicolumn{1}{c|}{} & OA & 0.9699 & {\color[HTML]{0096FF} 0.9888} & 0.9826 & 0.9845 & 0.9827 & 0.9808 & 0.9805 & 0.9864 & {\color[HTML]{FF0000} \textbf{0.9908}} \\
			\multicolumn{1}{c|}{} & Pre & 0.7615 & {\color[HTML]{0096FF} 0.9045} & 0.8517 & 0.8708 & 0.8871 & 0.8315 & 0.9073 & 0.8962 & {\color[HTML]{FF0000} \textbf{0.9245}} \\
			\multicolumn{1}{c|}{} & Rec & 0.5953 & {\color[HTML]{0096FF} 0.8728} & 0.7967 & 0.8171 & 0.7558 & 0.7816 & 0.6871 & 0.8281 & {\color[HTML]{FF0000} \textbf{0.8929}} \\
			\multicolumn{1}{c|}{} & F1 & 0.6683 & {\color[HTML]{0096FF} 0.8884} & 0.8232 & 0.8431 & 0.8162 & 0.8057 & 0.7822 & 0.8608 & {\color[HTML]{FF0000} \textbf{0.9084}} \\
			\multicolumn{1}{c|}{\multirow{-5}{*}{LEVIR-CD}} & mIoU & 0.7354 & {\color[HTML]{0096FF} 0.8937} & 0.8407 & 0.8563 & 0.8357 & 0.8273 & 0.8111 & 0.8707 & {\color[HTML]{FF0000} \textbf{0.9113}} \\ \midrule
			\multicolumn{1}{c|}{} & OA & 0.8813 & 0.8730 & 0.8912 & 0.8881 & 0.8641 & 0.8852 & 0.8546 & {\color[HTML]{0096FF} 0.8916} & {\color[HTML]{FF0000} \textbf{0.9204}} \\
			\multicolumn{1}{c|}{} & Pre & 0.7266 & 0.7003 & 0.7937 & 0.8014 & 0.7251 & 0.7630 & {\color[HTML]{0096FF} 0.8562} & 0.8021 & {\color[HTML]{FF0000} \textbf{0.8707}} \\
			\multicolumn{1}{c|}{} & Rec & {\color[HTML]{0096FF} 0.7963} & {\color[HTML]{FF0000} \textbf{0.8065}} & 0.7278 & 0.6984 & 0.6823 & 0.7444 & 0.4608 & 0.7172 & 0.7778 \\
			\multicolumn{1}{c|}{} & F1 & {\color[HTML]{0096FF} 0.7599} & 0.7497 & 0.7593 & 0.7464 & 0.7030 & 0.7536 & 0.5991 & 0.7573 & {\color[HTML]{FF0000} \textbf{0.8217}} \\
			\multicolumn{1}{c|}{\multirow{-5}{*}{SYSU-CD}} & mIoU & 0.7333 & 0.7214 & {\color[HTML]{0096FF} 0.7404} & 0.7307 & 0.6900 & 0.7327 & 0.6323 & 0.7394 & {\color[HTML]{FF0000} \textbf{0.7999}} \\ \bottomrule % thicker line
	\end{tabular}
\end{table*}

\subsubsection{Qualtitative Comparisons}
\begin{figure*}
		\centering
		\begin{minipage}{0.08\linewidth}
			\vspace{1pt}
			\centerline{\includegraphics[width=\textwidth]{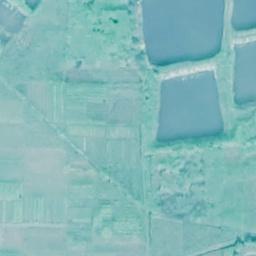}}
			\vspace{1pt}
			\centerline{\includegraphics[width=\textwidth]{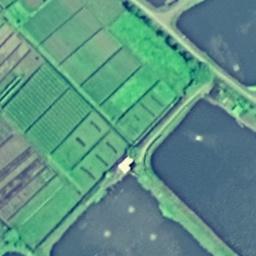}}
			\vspace{1pt}
			\centerline{\includegraphics[width=\textwidth]{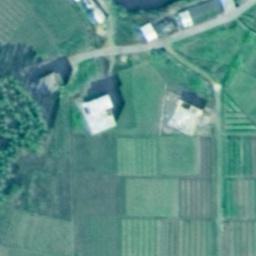}}
			\vspace{1pt}
                \centerline{\includegraphics[width=\textwidth]{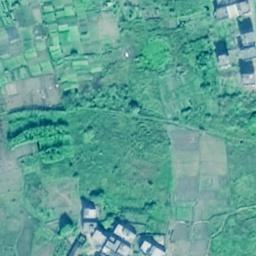}}
			\vspace{1pt}
                \centerline{\includegraphics[width=\textwidth]{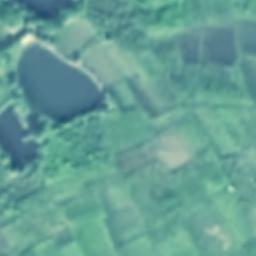}}
			\vspace{1pt}
                \centerline{\includegraphics[width=\textwidth]{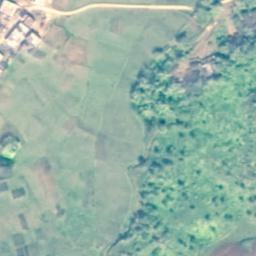}}
			\vspace{1pt}
			\centerline{(a)}
		\end{minipage}
		\hspace{-5pt}
		\begin{minipage}{0.08\linewidth}
			\vspace{1pt}
			\centerline{\includegraphics[width=\textwidth]{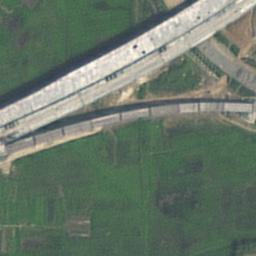}}
			\vspace{1pt}
			\centerline{\includegraphics[width=\textwidth]{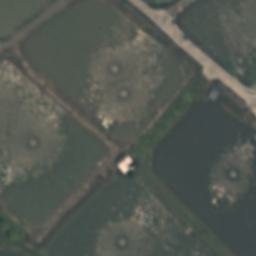}}
			\vspace{1pt}
			\centerline{\includegraphics[width=\textwidth]{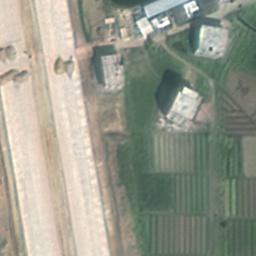}}
			\vspace{1pt}
                \centerline{\includegraphics[width=\textwidth]{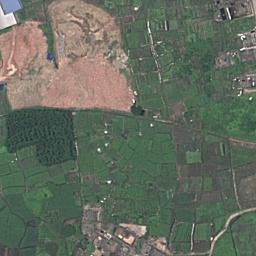}}
			\vspace{1pt}
                \centerline{\includegraphics[width=\textwidth]{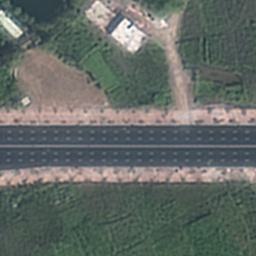}}
			\vspace{1pt}
                \centerline{\includegraphics[width=\textwidth]{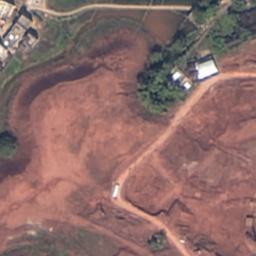}}
			\vspace{1pt}
			\centerline{(b)}
		\end{minipage}
		\hspace{-5pt}
		\begin{minipage}{0.08\linewidth}
			\vspace{1pt}
			\centerline{\includegraphics[width=\textwidth]{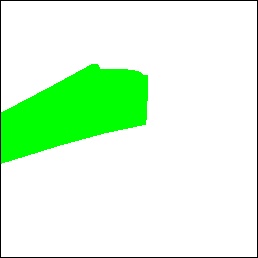}}
			\vspace{1pt}
			\centerline{\includegraphics[width=\textwidth]{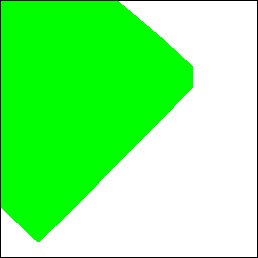}}
			\vspace{1pt}
			\centerline{\includegraphics[width=\textwidth]{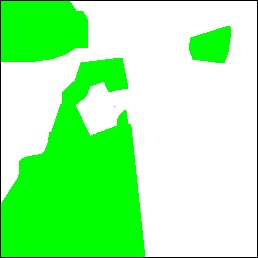}}
			\vspace{1pt}
                \centerline{\includegraphics[width=\textwidth]{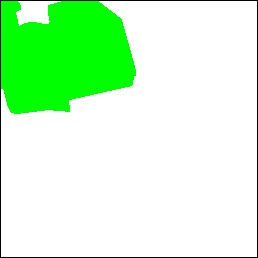}}
			\vspace{1pt}
                \centerline{\includegraphics[width=\textwidth]{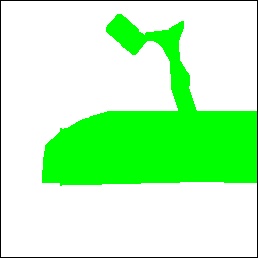}}
			\vspace{1pt}
                \centerline{\includegraphics[width=\textwidth]{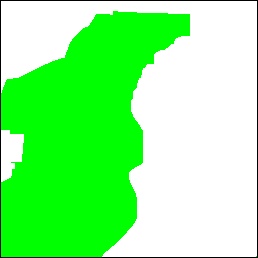}}
			\vspace{1pt}
			\centerline{(c)}
		\end{minipage}
		\hspace{-5pt}
		\begin{minipage}{0.08\linewidth}
			\vspace{1pt}
			\centerline{\includegraphics[width=\textwidth]{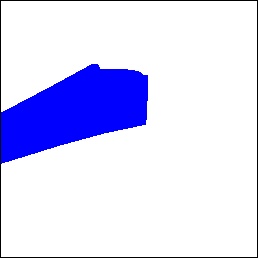}}
			\vspace{1pt}
			\centerline{\includegraphics[width=\textwidth]{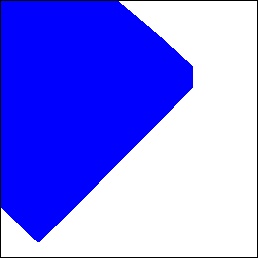}}
			\vspace{1pt}
			\centerline{\includegraphics[width=\textwidth]{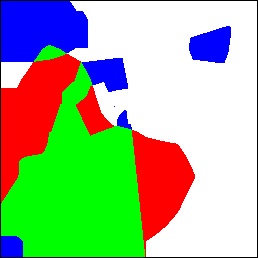}}
			\vspace{1pt}
                \centerline{\includegraphics[width=\textwidth]{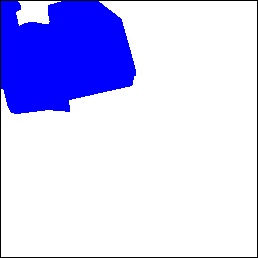}}
			\vspace{1pt}
                \centerline{\includegraphics[width=\textwidth]{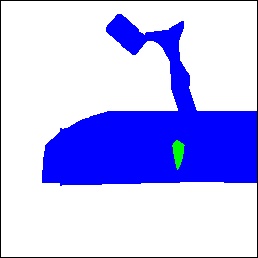}}
			\vspace{1pt}
                \centerline{\includegraphics[width=\textwidth]{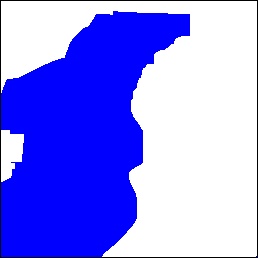}}
			\vspace{1pt}
			\centerline{(d)}
		\end{minipage}
		\hspace{-5pt}
		\begin{minipage}{0.08\linewidth}
			\vspace{1pt}
			\centerline{\includegraphics[width=\textwidth]{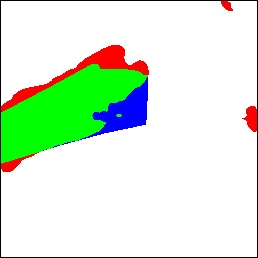}}
			\vspace{1pt}
			\centerline{\includegraphics[width=\textwidth]{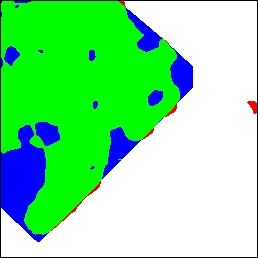}}
			\vspace{1pt}
			\centerline{\includegraphics[width=\textwidth]{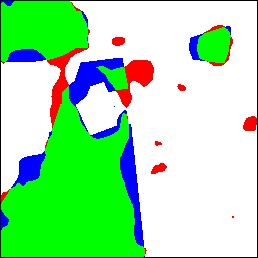}}
			\vspace{1pt}
                \centerline{\includegraphics[width=\textwidth]{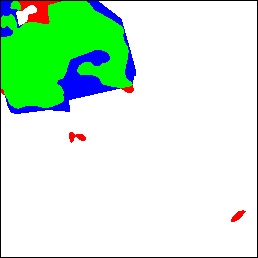}}
			\vspace{1pt}
                \centerline{\includegraphics[width=\textwidth]{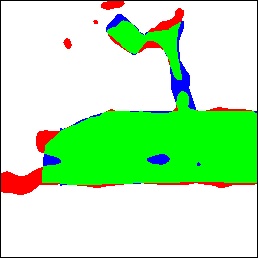}}
			\vspace{1pt}
                \centerline{\includegraphics[width=\textwidth]{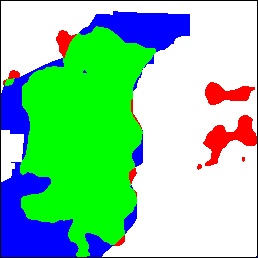}}
			\vspace{1pt}
			\centerline{(e)}
		\end{minipage}
		\hspace{-5pt}
		\begin{minipage}{0.08\linewidth}
			\vspace{1pt}
			\centerline{\includegraphics[width=\textwidth]{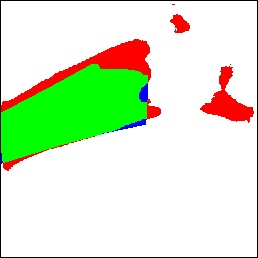}}
			\vspace{1pt}
			\centerline{\includegraphics[width=\textwidth]{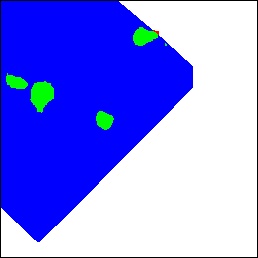}}
			\vspace{1pt}
			\centerline{\includegraphics[width=\textwidth]{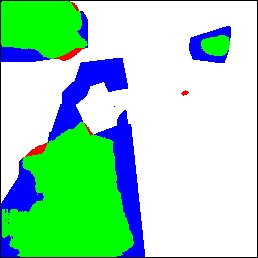}}
			\vspace{1pt}
                \centerline{\includegraphics[width=\textwidth]{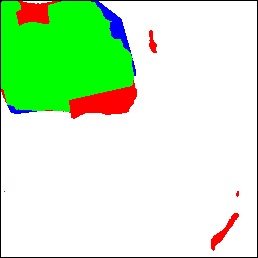}}
			\vspace{1pt}
                \centerline{\includegraphics[width=\textwidth]{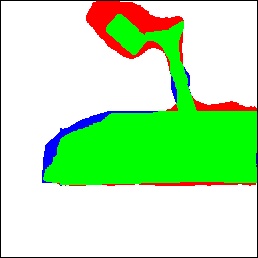}}
			\vspace{1pt}
                \centerline{\includegraphics[width=\textwidth]{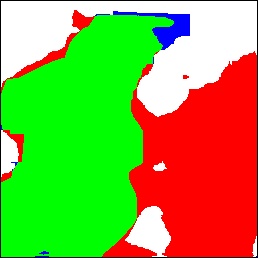}}
			\vspace{1pt}
			\centerline{(f)}
		\end{minipage}
  \hspace{-5pt}
		\begin{minipage}{0.08\linewidth}
			\vspace{1pt}
			\centerline{\includegraphics[width=\textwidth]{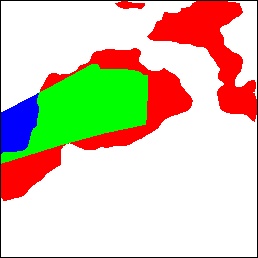}}
			\vspace{1pt}
			\centerline{\includegraphics[width=\textwidth]{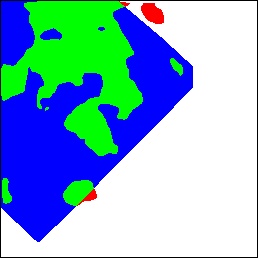}}
			\vspace{1pt}
			\centerline{\includegraphics[width=\textwidth]{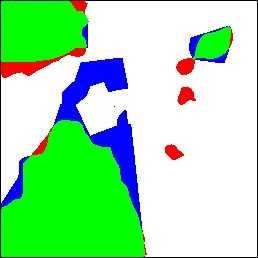}}
			\vspace{1pt}
                \centerline{\includegraphics[width=\textwidth]{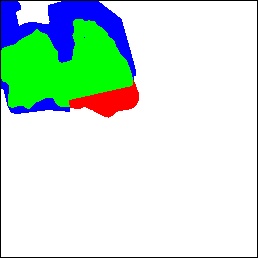}}
			\vspace{1pt}
                \centerline{\includegraphics[width=\textwidth]{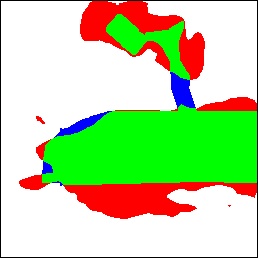}}
			\vspace{1pt}
                \centerline{\includegraphics[width=\textwidth]{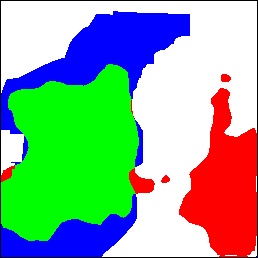}}
			\vspace{1pt}
			\centerline{(g)}
		\end{minipage}
    \hspace{-5pt}
	\begin{minipage}{0.08\linewidth}
		\vspace{1pt}
		\centerline{\includegraphics[width=\textwidth]{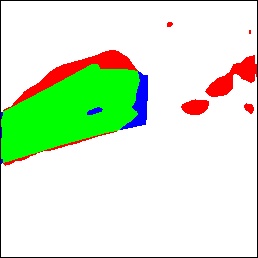}}
		\vspace{1pt}
		\centerline{\includegraphics[width=\textwidth]{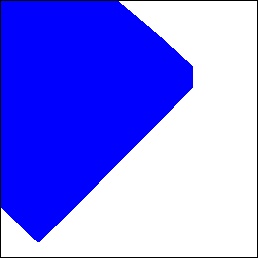}}
		\vspace{1pt}
		\centerline{\includegraphics[width=\textwidth]{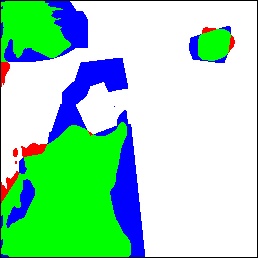}}
		\vspace{1pt}
			\centerline{\includegraphics[width=\textwidth]{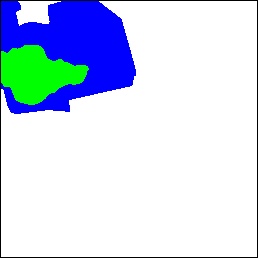}}
		\vspace{1pt}
			\centerline{\includegraphics[width=\textwidth]{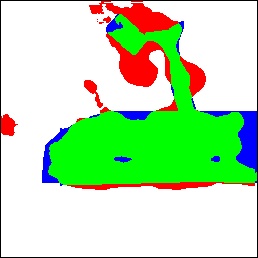}}
		\vspace{1pt}
			\centerline{\includegraphics[width=\textwidth]{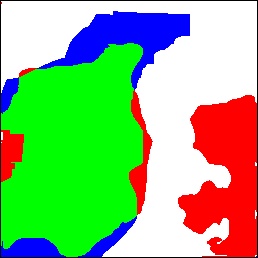}}
		\vspace{1pt}
		\centerline{(h)}
	\end{minipage}
  \hspace{-5pt}
  \begin{minipage}{0.08\linewidth}
	\vspace{1pt}
	\centerline{\includegraphics[width=\textwidth]{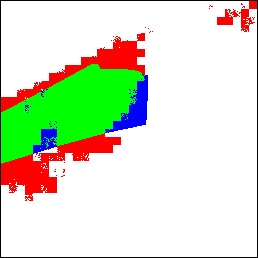}}
	\vspace{1pt}
	\centerline{\includegraphics[width=\textwidth]{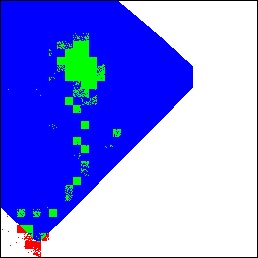}}
	\vspace{1pt}
	\centerline{\includegraphics[width=\textwidth]{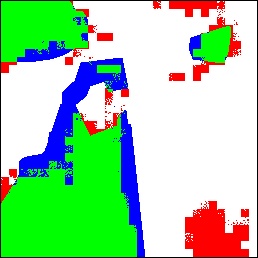}}
	\vspace{1pt}
		\centerline{\includegraphics[width=\textwidth]{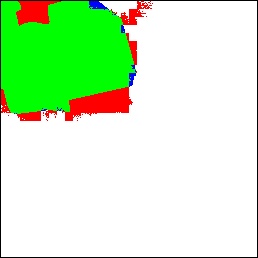}}
	\vspace{1pt}
		\centerline{\includegraphics[width=\textwidth]{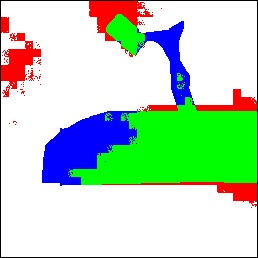}}
	\vspace{1pt}
		\centerline{\includegraphics[width=\textwidth]{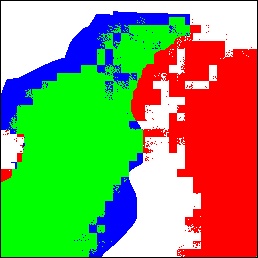}}
	\vspace{1pt}
	\centerline{(i)}
\end{minipage}
  \hspace{-5pt}
  \begin{minipage}{0.08\linewidth}
	\vspace{1pt}
	\centerline{\includegraphics[width=\textwidth]{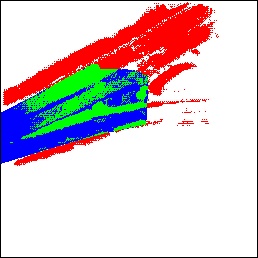}}
	\vspace{1pt}
	\centerline{\includegraphics[width=\textwidth]{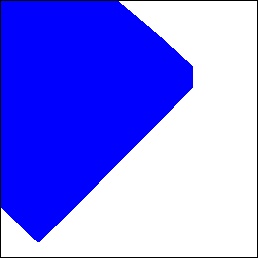}}
	\vspace{1pt}
	\centerline{\includegraphics[width=\textwidth]{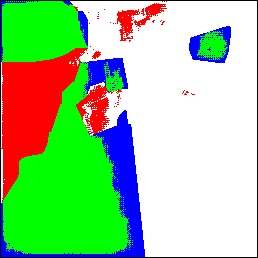}}
	\vspace{1pt}
		\centerline{\includegraphics[width=\textwidth]{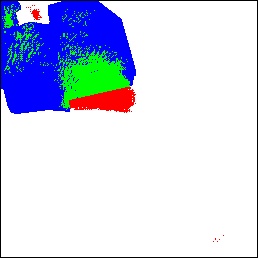}}
	\vspace{1pt}
		\centerline{\includegraphics[width=\textwidth]{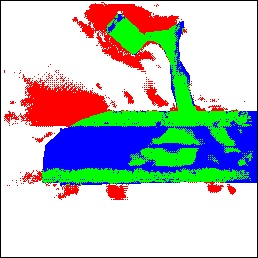}}
	\vspace{1pt}
		\centerline{\includegraphics[width=\textwidth]{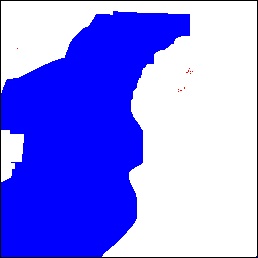}}
	\vspace{1pt}
	\centerline{(j)}
\end{minipage}
  \hspace{-5pt}
  \begin{minipage}{0.08\linewidth}
	\vspace{1pt}
	\centerline{\includegraphics[width=\textwidth]{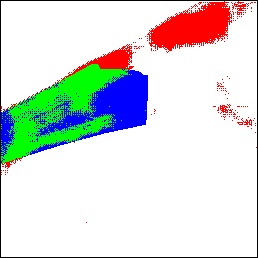}}
	\vspace{1pt}
	\centerline{\includegraphics[width=\textwidth]{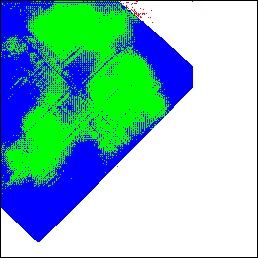}}
	\vspace{1pt}
	\centerline{\includegraphics[width=\textwidth]{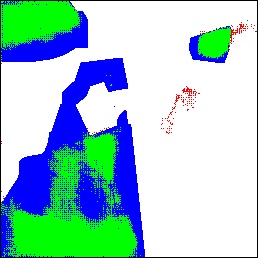}}
	\vspace{1pt}
		\centerline{\includegraphics[width=\textwidth]{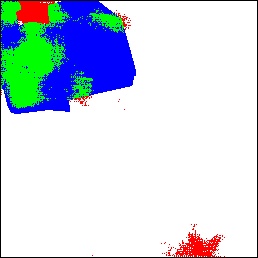}}
	\vspace{1pt}
		\centerline{\includegraphics[width=\textwidth]{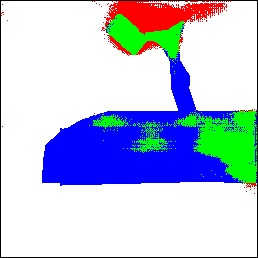}}
	\vspace{1pt}
		\centerline{\includegraphics[width=\textwidth]{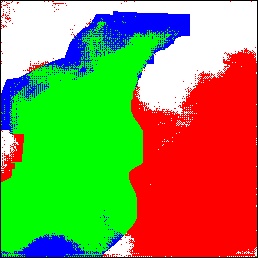}}
	\vspace{1pt}
	\centerline{(k)}
\end{minipage}
  \hspace{-5pt}
  \begin{minipage}{0.08\linewidth}
	\vspace{1pt}
	\centerline{\includegraphics[width=\textwidth]{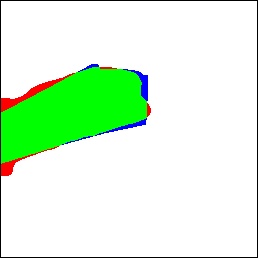}}
	\vspace{1pt}
	\centerline{\includegraphics[width=\textwidth]{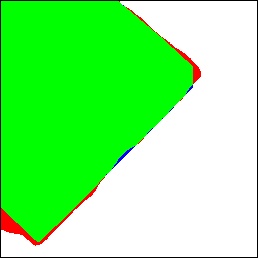}}
	\vspace{1pt}
	\centerline{\includegraphics[width=\textwidth]{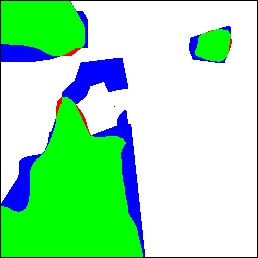}}
	\vspace{1pt}
		\centerline{\includegraphics[width=\textwidth]{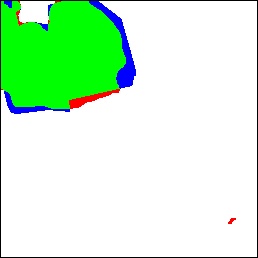}}
	\vspace{1pt}
		\centerline{\includegraphics[width=\textwidth]{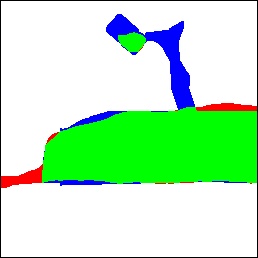}}
	\vspace{1pt}
		\centerline{\includegraphics[width=\textwidth]{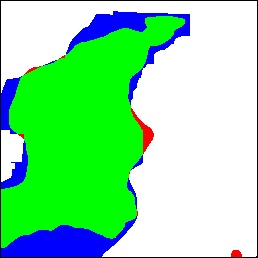}}
	\vspace{1pt}
	\centerline{(l)}
\end{minipage}
		\caption{Visual comparisons of the proposed method and the state-of-the-art approaches on the CLCD dataset. (a) T1 images. (b) T2 images. (c) Ground truth. (d) A2Net. (e) BIT. (f) ChangeFormer. (g) DMINet. (h) ICIFNet. (i) RDPnet. (j) SiamUnet-diff. (k) SNUNet. (l) MaskCD. The rendered colors represent true positives (green), true negatives (white), false positives (red), and false negatives (blue).} 
		\label{CLCDQuali}
	\end{figure*}

\begin{figure*}
	\centering
	\begin{minipage}{0.08\linewidth}
		\vspace{1pt}
		\centerline{\includegraphics[width=\textwidth]{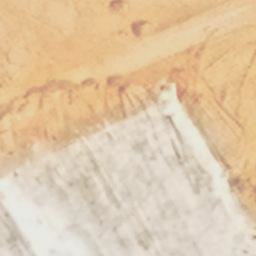}}
		\vspace{1pt}
		\centerline{\includegraphics[width=\textwidth]{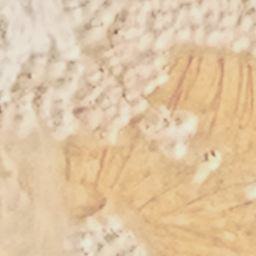}}
		\vspace{1pt}
		\centerline{\includegraphics[width=\textwidth]{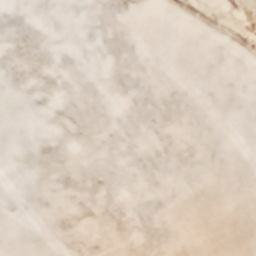}}
		\vspace{1pt}
		\centerline{\includegraphics[width=\textwidth]{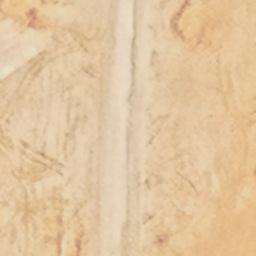}}
		\vspace{1pt}
		\centerline{\includegraphics[width=\textwidth]{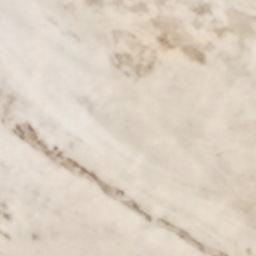}}
		\vspace{1pt}
		\centerline{\includegraphics[width=\textwidth]{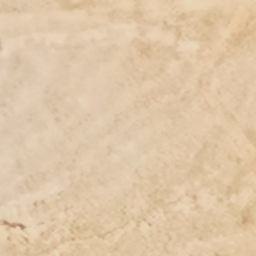}}
		\vspace{1pt}
		\centerline{(a)}
	\end{minipage}
	\hspace{-5pt}
	\begin{minipage}{0.08\linewidth}
		\vspace{1pt}
		\centerline{\includegraphics[width=\textwidth]{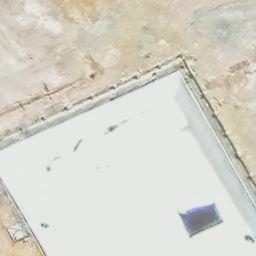}}
		\vspace{1pt}
		\centerline{\includegraphics[width=\textwidth]{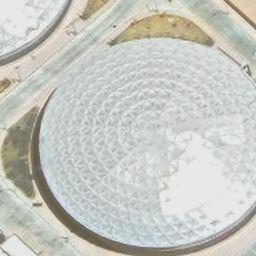}}
		\vspace{1pt}
		\centerline{\includegraphics[width=\textwidth]{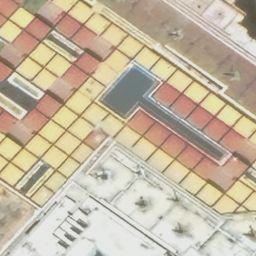}}
		\vspace{1pt}
		\centerline{\includegraphics[width=\textwidth]{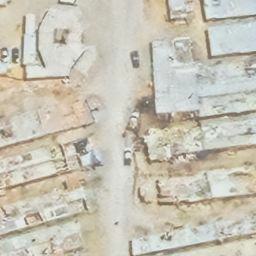}}
		\vspace{1pt}
		\centerline{\includegraphics[width=\textwidth]{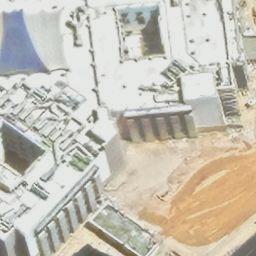}}
		\vspace{1pt}
		\centerline{\includegraphics[width=\textwidth]{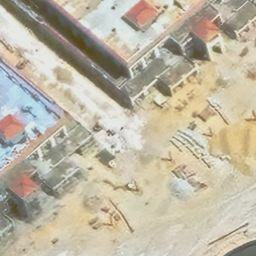}}
		\vspace{1pt}
		\centerline{(b)}
	\end{minipage}
	\hspace{-5pt}
	\begin{minipage}{0.08\linewidth}
		\vspace{1pt}
		\centerline{\includegraphics[width=\textwidth]{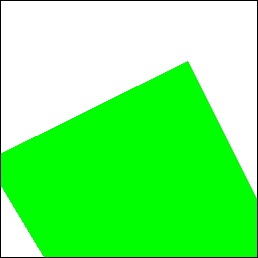}}
		\vspace{1pt}
		\centerline{\includegraphics[width=\textwidth]{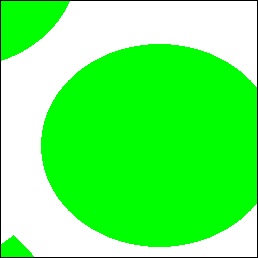}}
		\vspace{1pt}
		\centerline{\includegraphics[width=\textwidth]{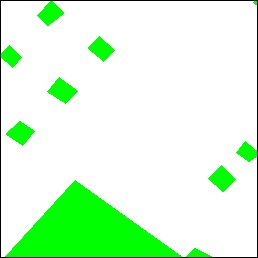}}
		\vspace{1pt}
		\centerline{\includegraphics[width=\textwidth]{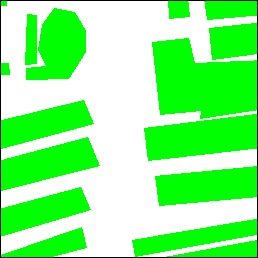}}
		\vspace{1pt}
		\centerline{\includegraphics[width=\textwidth]{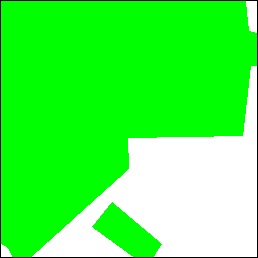}}
		\vspace{1pt}
		\centerline{\includegraphics[width=\textwidth]{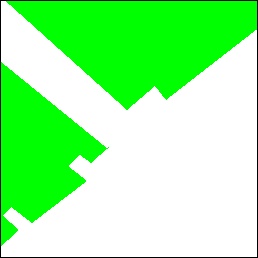}}
		\vspace{1pt}
		\centerline{(c)}
	\end{minipage}
	\hspace{-5pt}
	\begin{minipage}{0.08\linewidth}
		\vspace{1pt}
		\centerline{\includegraphics[width=\textwidth]{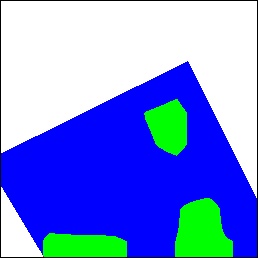}}
		\vspace{1pt}
		\centerline{\includegraphics[width=\textwidth]{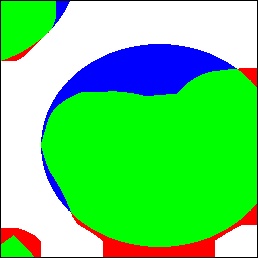}}
		\vspace{1pt}
		\centerline{\includegraphics[width=\textwidth]{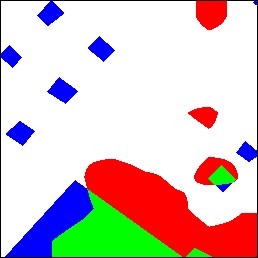}}
		\vspace{1pt}
		\centerline{\includegraphics[width=\textwidth]{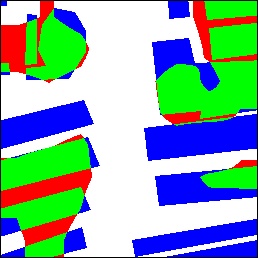}}
		\vspace{1pt}
		\centerline{\includegraphics[width=\textwidth]{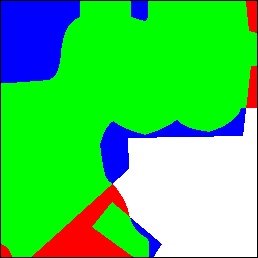}}
		\vspace{1pt}
		\centerline{\includegraphics[width=\textwidth]{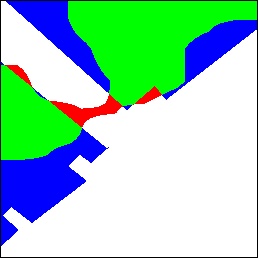}}
		\vspace{1pt}
		\centerline{(d)}
	\end{minipage}
	\hspace{-5pt}
	\begin{minipage}{0.08\linewidth}
		\vspace{1pt}
		\centerline{\includegraphics[width=\textwidth]{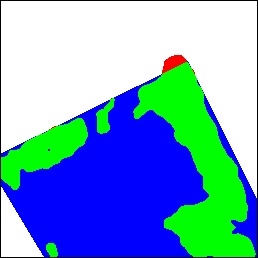}}
		\vspace{1pt}
		\centerline{\includegraphics[width=\textwidth]{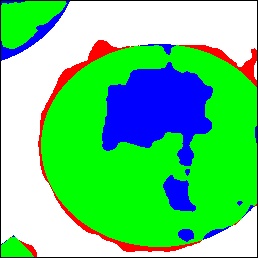}}
		\vspace{1pt}
		\centerline{\includegraphics[width=\textwidth]{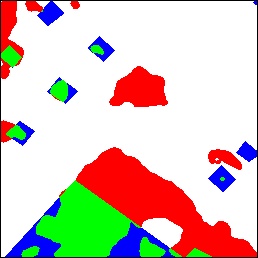}}
		\vspace{1pt}
		\centerline{\includegraphics[width=\textwidth]{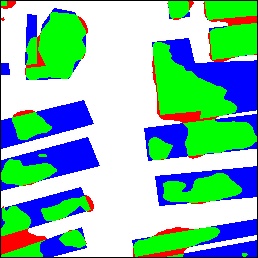}}
		\vspace{1pt}
		\centerline{\includegraphics[width=\textwidth]{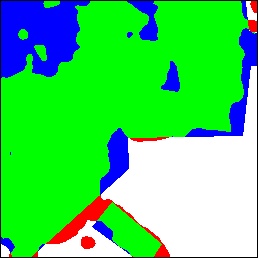}}
		\vspace{1pt}
		\centerline{\includegraphics[width=\textwidth]{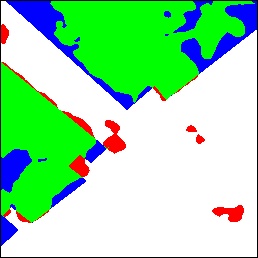}}
		\vspace{1pt}
		\centerline{(e)}
	\end{minipage}
	\hspace{-5pt}
	\begin{minipage}{0.08\linewidth}
		\vspace{1pt}
		\centerline{\includegraphics[width=\textwidth]{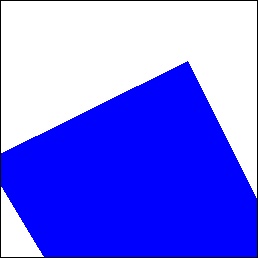}}
		\vspace{1pt}
		\centerline{\includegraphics[width=\textwidth]{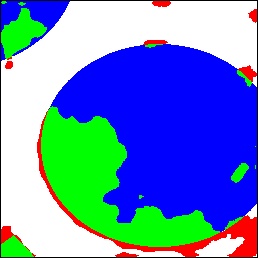}}
		\vspace{1pt}
		\centerline{\includegraphics[width=\textwidth]{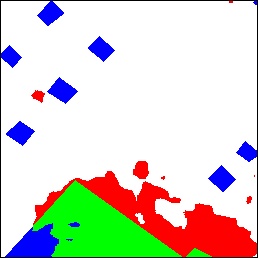}}
		\vspace{1pt}
		\centerline{\includegraphics[width=\textwidth]{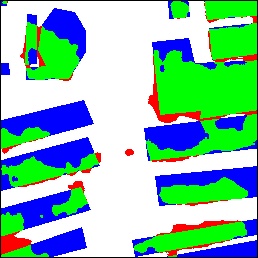}}
		\vspace{1pt}
		\centerline{\includegraphics[width=\textwidth]{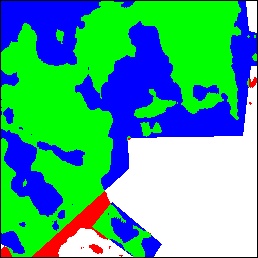}}
		\vspace{1pt}
		\centerline{\includegraphics[width=\textwidth]{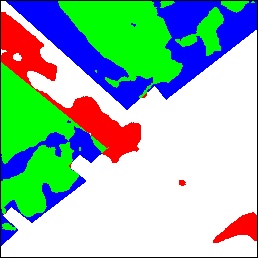}}
		\vspace{1pt}
		\centerline{(f)}
	\end{minipage}
	\hspace{-5pt}
	\begin{minipage}{0.08\linewidth}
		\vspace{1pt}
		\centerline{\includegraphics[width=\textwidth]{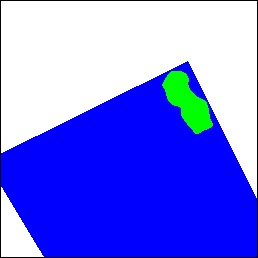}}
		\vspace{1pt}
		\centerline{\includegraphics[width=\textwidth]{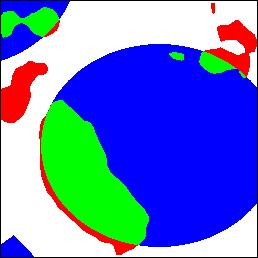}}
		\vspace{1pt}
		\centerline{\includegraphics[width=\textwidth]{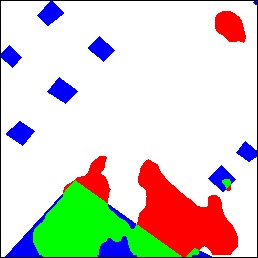}}
		\vspace{1pt}
		\centerline{\includegraphics[width=\textwidth]{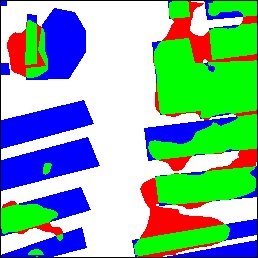}}
		\vspace{1pt}
		\centerline{\includegraphics[width=\textwidth]{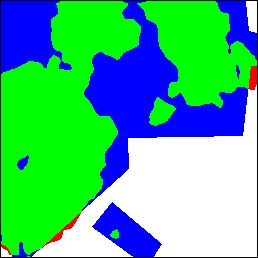}}
		\vspace{1pt}
		\centerline{\includegraphics[width=\textwidth]{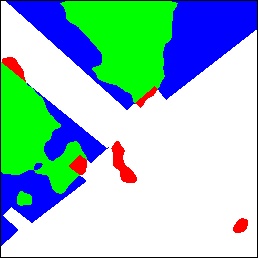}}
		\vspace{1pt}
		\centerline{(g)}
	\end{minipage}
	\hspace{-5pt}
	\begin{minipage}{0.08\linewidth}
		\vspace{1pt}
		\centerline{\includegraphics[width=\textwidth]{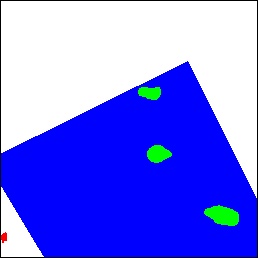}}
		\vspace{1pt}
		\centerline{\includegraphics[width=\textwidth]{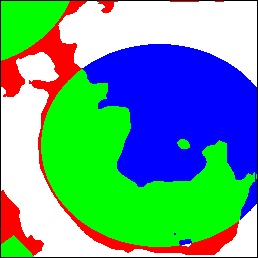}}
		\vspace{1pt}
		\centerline{\includegraphics[width=\textwidth]{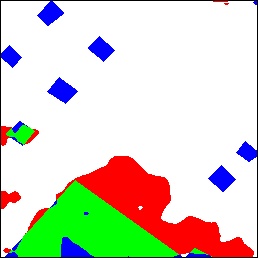}}
		\vspace{1pt}
		\centerline{\includegraphics[width=\textwidth]{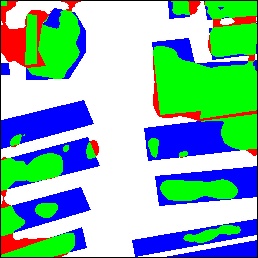}}
		\vspace{1pt}
		\centerline{\includegraphics[width=\textwidth]{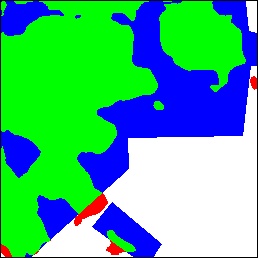}}
		\vspace{1pt}
		\centerline{\includegraphics[width=\textwidth]{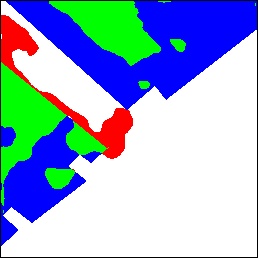}}
		\vspace{1pt}
		\centerline{(h)}
	\end{minipage}
	\hspace{-5pt}
	\begin{minipage}{0.08\linewidth}
		\vspace{1pt}
		\centerline{\includegraphics[width=\textwidth]{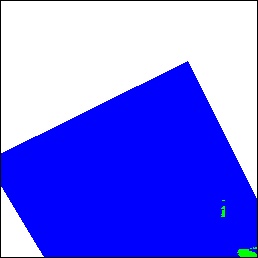}}
		\vspace{1pt}
		\centerline{\includegraphics[width=\textwidth]{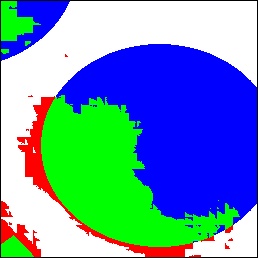}}
		\vspace{1pt}
		\centerline{\includegraphics[width=\textwidth]{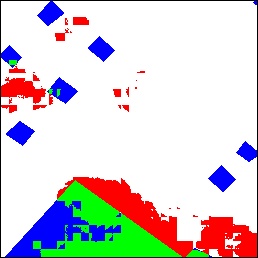}}
		\vspace{1pt}
		\centerline{\includegraphics[width=\textwidth]{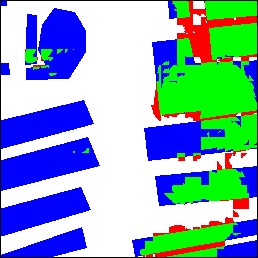}}
		\vspace{1pt}
		\centerline{\includegraphics[width=\textwidth]{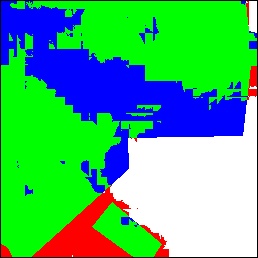}}
		\vspace{1pt}
		\centerline{\includegraphics[width=\textwidth]{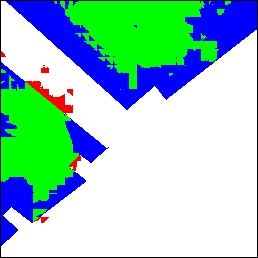}}
		\vspace{1pt}
		\centerline{(i)}
	\end{minipage}
	\hspace{-5pt}
	\begin{minipage}{0.08\linewidth}
		\vspace{1pt}
		\centerline{\includegraphics[width=\textwidth]{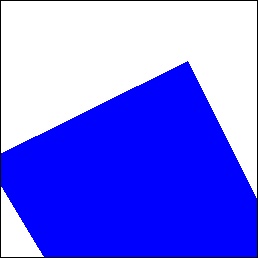}}
		\vspace{1pt}
		\centerline{\includegraphics[width=\textwidth]{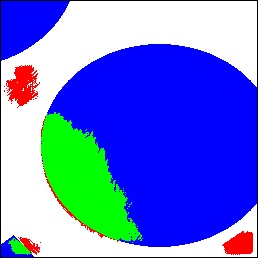}}
		\vspace{1pt}
		\centerline{\includegraphics[width=\textwidth]{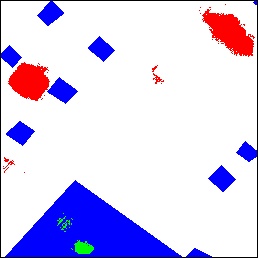}}
		\vspace{1pt}
		\centerline{\includegraphics[width=\textwidth]{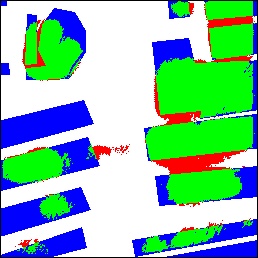}}
		\vspace{1pt}
		\centerline{\includegraphics[width=\textwidth]{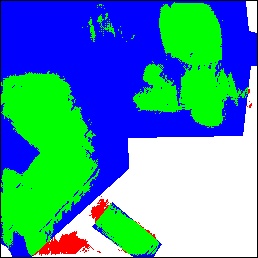}}
		\vspace{1pt}
		\centerline{\includegraphics[width=\textwidth]{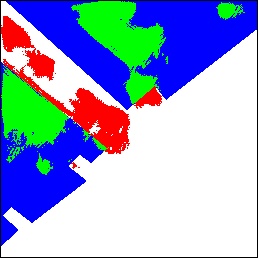}}
		\vspace{1pt}
		\centerline{(j)}
	\end{minipage}
	\hspace{-5pt}
	\begin{minipage}{0.08\linewidth}
		\vspace{1pt}
		\centerline{\includegraphics[width=\textwidth]{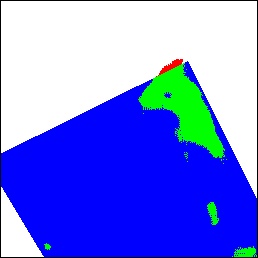}}
		\vspace{1pt}
		\centerline{\includegraphics[width=\textwidth]{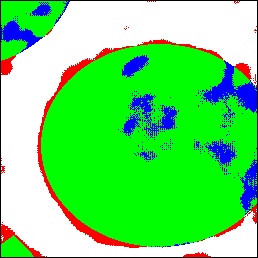}}
		\vspace{1pt}
		\centerline{\includegraphics[width=\textwidth]{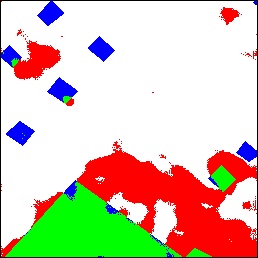}}
		\vspace{1pt}
		\centerline{\includegraphics[width=\textwidth]{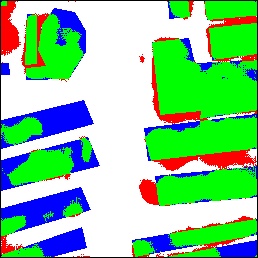}}
		\vspace{1pt}
		\centerline{\includegraphics[width=\textwidth]{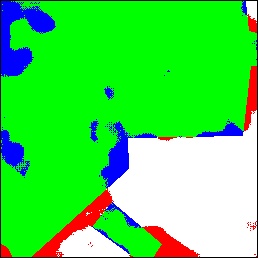}}
		\vspace{1pt}
		\centerline{\includegraphics[width=\textwidth]{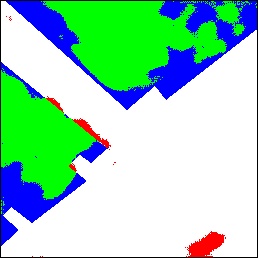}}
		\vspace{1pt}
		\centerline{(k)}
	\end{minipage}
	\hspace{-5pt}
	\begin{minipage}{0.08\linewidth}
		\vspace{1pt}
		\centerline{\includegraphics[width=\textwidth]{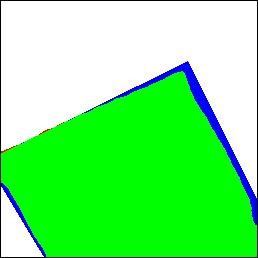}}
		\vspace{1pt}
		\centerline{\includegraphics[width=\textwidth]{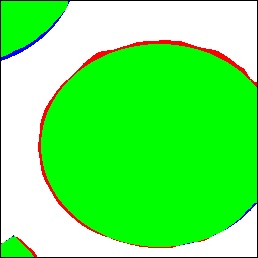}}
		\vspace{1pt}
		\centerline{\includegraphics[width=\textwidth]{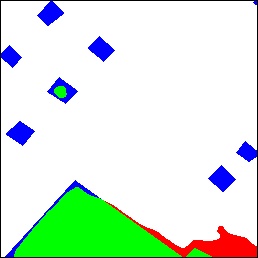}}
		\vspace{1pt}
		\centerline{\includegraphics[width=\textwidth]{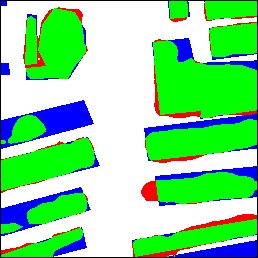}}
		\vspace{1pt}
		\centerline{\includegraphics[width=\textwidth]{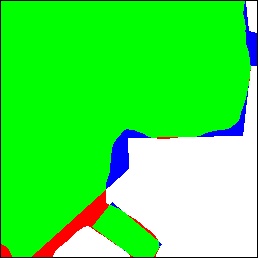}}
		\vspace{1pt}
		\centerline{\includegraphics[width=\textwidth]{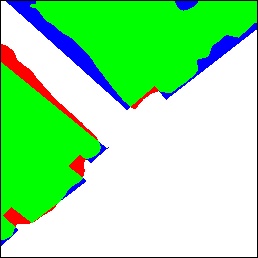}}
		\vspace{1pt}
		\centerline{(l)}
	\end{minipage}
	\caption{Visual comparisons of the proposed method and the state-of-the-art approaches on the EGY-BCD dataset. (a) T1 images. (b) T2 images. (c) Ground truth. (d) A2Net. (e) BIT. (f) ChangeFormer. (g) DMINet. (h) ICIFNet. (i) RDPnet. (j) SiamUnet-diff. (k) SNUNet. (l) MaskCD. The rendered colors represent true positives (green), true negatives (white), false positives (red), and false negatives (blue).} 
	\label{EGYBCDQuali}
\end{figure*}

\begin{figure*}
	\centering
	\begin{minipage}{0.08\linewidth}
		\vspace{1pt}
		\centerline{\includegraphics[width=\textwidth]{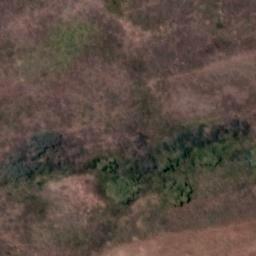}}
		\vspace{1pt}
		\centerline{\includegraphics[width=\textwidth]{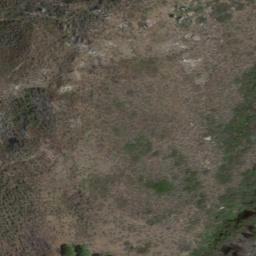}}
		\vspace{1pt}
		\centerline{\includegraphics[width=\textwidth]{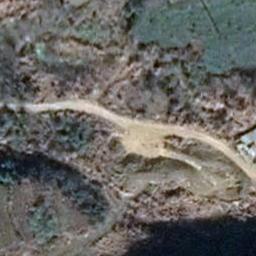}}
		\vspace{1pt}
		\centerline{\includegraphics[width=\textwidth]{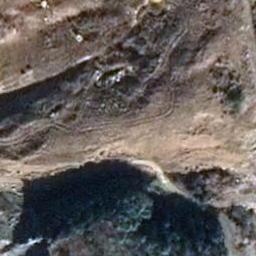}}
		\vspace{1pt}
		\centerline{\includegraphics[width=\textwidth]{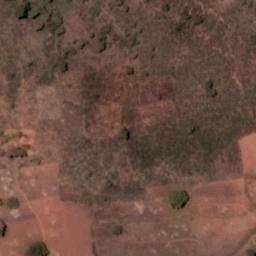}}
		\vspace{1pt}
		\centerline{\includegraphics[width=\textwidth]{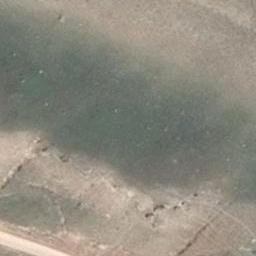}}
		\vspace{1pt}
		\centerline{(a)}
	\end{minipage}
	\hspace{-5pt}
	\begin{minipage}{0.08\linewidth}
		\vspace{1pt}
		\centerline{\includegraphics[width=\textwidth]{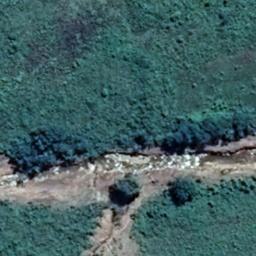}}
		\vspace{1pt}
		\centerline{\includegraphics[width=\textwidth]{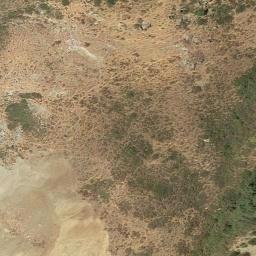}}
		\vspace{1pt}
		\centerline{\includegraphics[width=\textwidth]{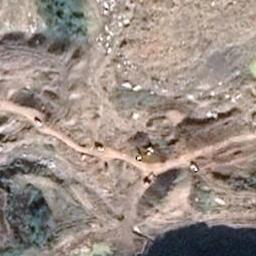}}
		\vspace{1pt}
		\centerline{\includegraphics[width=\textwidth]{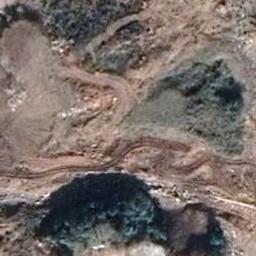}}
		\vspace{1pt}
		\centerline{\includegraphics[width=\textwidth]{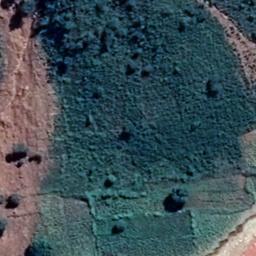}}
		\vspace{1pt}
		\centerline{\includegraphics[width=\textwidth]{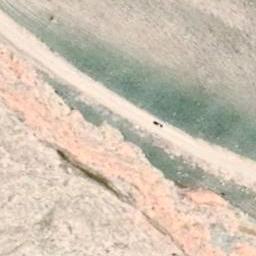}}
		\vspace{1pt}
		\centerline{(b)}
	\end{minipage}
	\hspace{-5pt}
	\begin{minipage}{0.08\linewidth}
		\vspace{1pt}
		\centerline{\includegraphics[width=\textwidth]{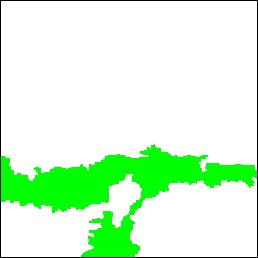}}
		\vspace{1pt}
		\centerline{\includegraphics[width=\textwidth]{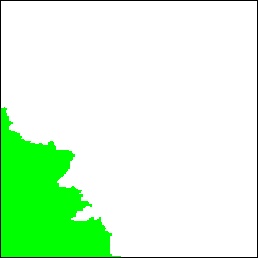}}
		\vspace{1pt}
		\centerline{\includegraphics[width=\textwidth]{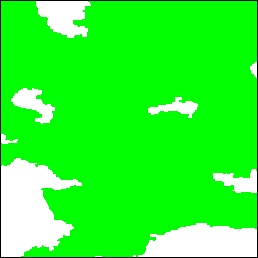}}
		\vspace{1pt}
		\centerline{\includegraphics[width=\textwidth]{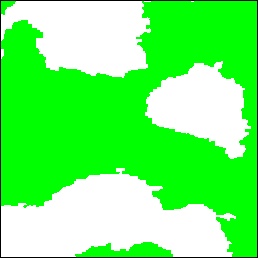}}
		\vspace{1pt}
		\centerline{\includegraphics[width=\textwidth]{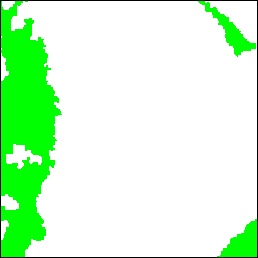}}
		\vspace{1pt}
		\centerline{\includegraphics[width=\textwidth]{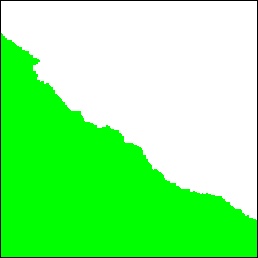}}
		\vspace{1pt}
		\centerline{(c)}
	\end{minipage}
	\hspace{-5pt}
	\begin{minipage}{0.08\linewidth}
		\vspace{1pt}
		\centerline{\includegraphics[width=\textwidth]{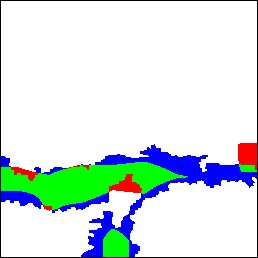}}
		\vspace{1pt}
		\centerline{\includegraphics[width=\textwidth]{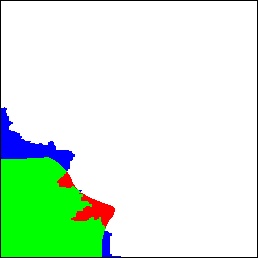}}
		\vspace{1pt}
		\centerline{\includegraphics[width=\textwidth]{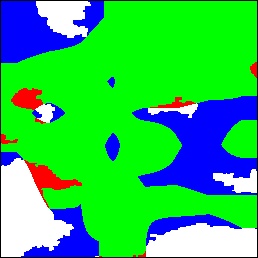}}
		\vspace{1pt}
		\centerline{\includegraphics[width=\textwidth]{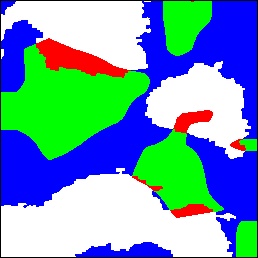}}
		\vspace{1pt}
		\centerline{\includegraphics[width=\textwidth]{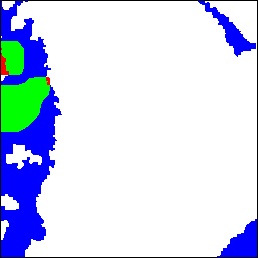}}
		\vspace{1pt}
		\centerline{\includegraphics[width=\textwidth]{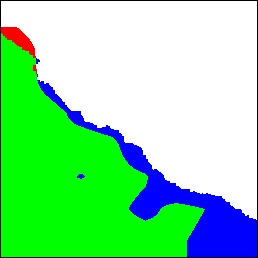}}
		\vspace{1pt}
		\centerline{(d)}
	\end{minipage}
	\hspace{-5pt}
	\begin{minipage}{0.08\linewidth}
		\vspace{1pt}
		\centerline{\includegraphics[width=\textwidth]{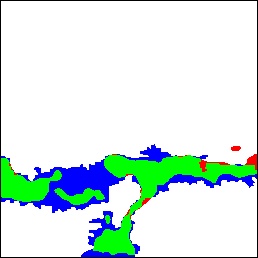}}
		\vspace{1pt}
		\centerline{\includegraphics[width=\textwidth]{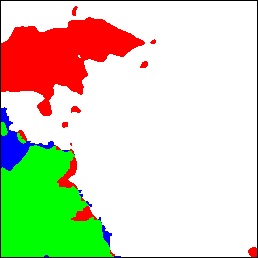}}
		\vspace{1pt}
		\centerline{\includegraphics[width=\textwidth]{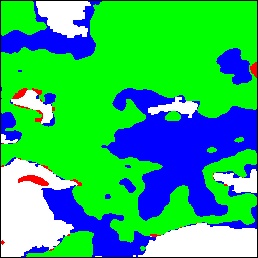}}
		\vspace{1pt}
		\centerline{\includegraphics[width=\textwidth]{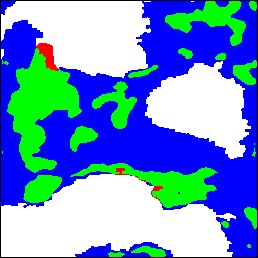}}
		\vspace{1pt}
		\centerline{\includegraphics[width=\textwidth]{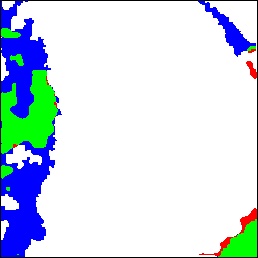}}
		\vspace{1pt}
		\centerline{\includegraphics[width=\textwidth]{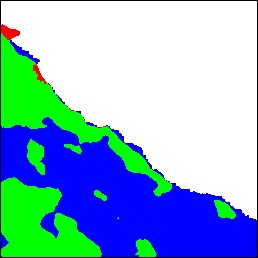}}
		\vspace{1pt}
		\centerline{(e)}
	\end{minipage}
	\hspace{-5pt}
	\begin{minipage}{0.08\linewidth}
		\vspace{1pt}
		\centerline{\includegraphics[width=\textwidth]{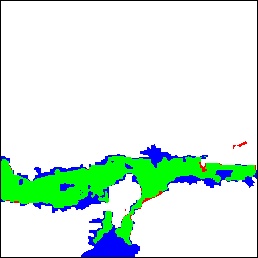}}
		\vspace{1pt}
		\centerline{\includegraphics[width=\textwidth]{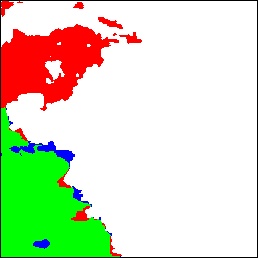}}
		\vspace{1pt}
		\centerline{\includegraphics[width=\textwidth]{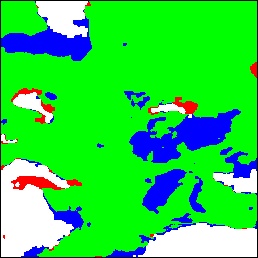}}
		\vspace{1pt}
		\centerline{\includegraphics[width=\textwidth]{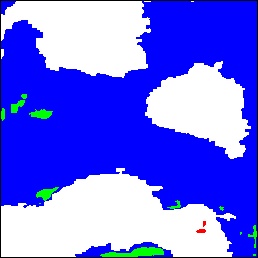}}
		\vspace{1pt}
		\centerline{\includegraphics[width=\textwidth]{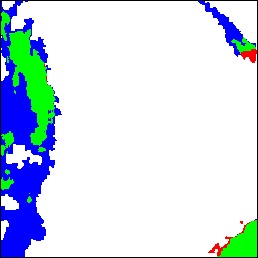}}
		\vspace{1pt}
		\centerline{\includegraphics[width=\textwidth]{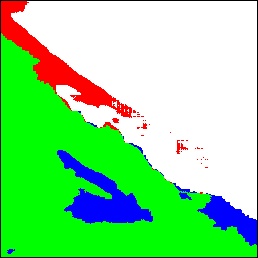}}
		\vspace{1pt}
		\centerline{(f)}
	\end{minipage}
	\hspace{-5pt}
	\begin{minipage}{0.08\linewidth}
		\vspace{1pt}
		\centerline{\includegraphics[width=\textwidth]{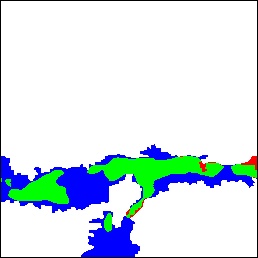}}
		\vspace{1pt}
		\centerline{\includegraphics[width=\textwidth]{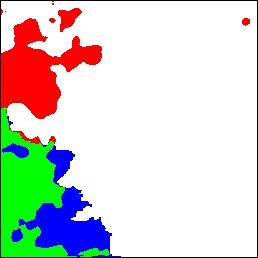}}
		\vspace{1pt}
		\centerline{\includegraphics[width=\textwidth]{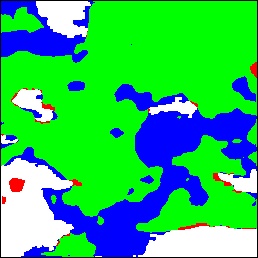}}
		\vspace{1pt}
		\centerline{\includegraphics[width=\textwidth]{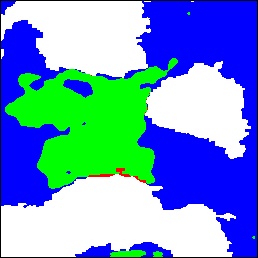}}
		\vspace{1pt}
		\centerline{\includegraphics[width=\textwidth]{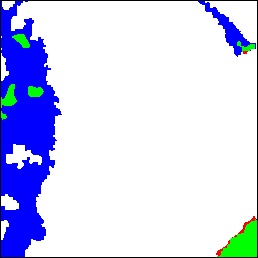}}
		\vspace{1pt}
		\centerline{\includegraphics[width=\textwidth]{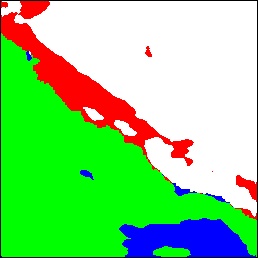}}
		\vspace{1pt}
		\centerline{(g)}
	\end{minipage}
	\hspace{-5pt}
	\begin{minipage}{0.08\linewidth}
		\vspace{1pt}
		\centerline{\includegraphics[width=\textwidth]{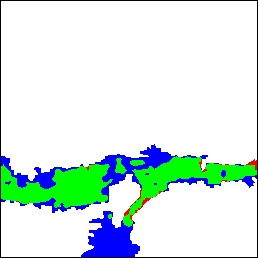}}
		\vspace{1pt}
		\centerline{\includegraphics[width=\textwidth]{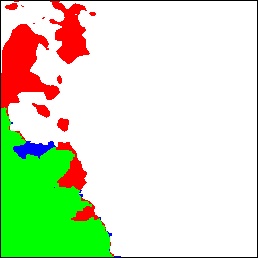}}
		\vspace{1pt}
		\centerline{\includegraphics[width=\textwidth]{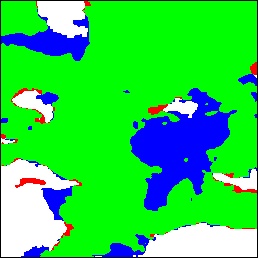}}
		\vspace{1pt}
		\centerline{\includegraphics[width=\textwidth]{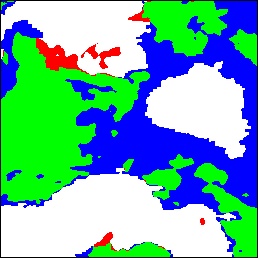}}
		\vspace{1pt}
		\centerline{\includegraphics[width=\textwidth]{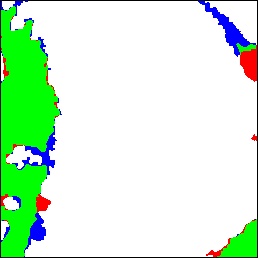}}
		\vspace{1pt}
		\centerline{\includegraphics[width=\textwidth]{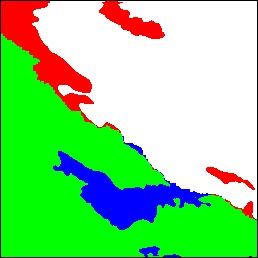}}
		\vspace{1pt}
		\centerline{(h)}
	\end{minipage}
	\hspace{-5pt}
	\begin{minipage}{0.08\linewidth}
		\vspace{1pt}
		\centerline{\includegraphics[width=\textwidth]{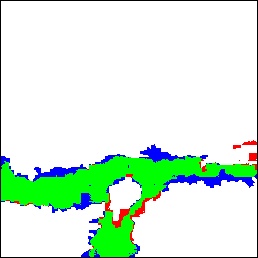}}
		\vspace{1pt}
		\centerline{\includegraphics[width=\textwidth]{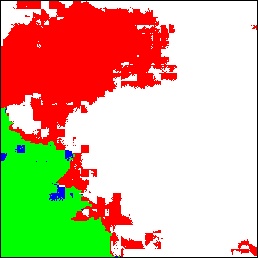}}
		\vspace{1pt}
		\centerline{\includegraphics[width=\textwidth]{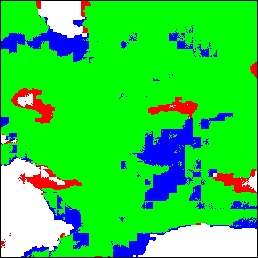}}
		\vspace{1pt}
		\centerline{\includegraphics[width=\textwidth]{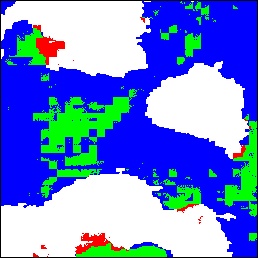}}
		\vspace{1pt}
		\centerline{\includegraphics[width=\textwidth]{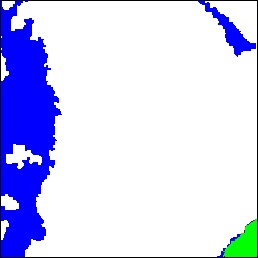}}
		\vspace{1pt}
		\centerline{\includegraphics[width=\textwidth]{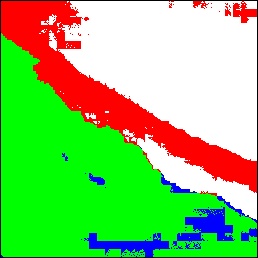}}
		\vspace{1pt}
		\centerline{(i)}
	\end{minipage}
	\hspace{-5pt}
	\begin{minipage}{0.08\linewidth}
		\vspace{1pt}
		\centerline{\includegraphics[width=\textwidth]{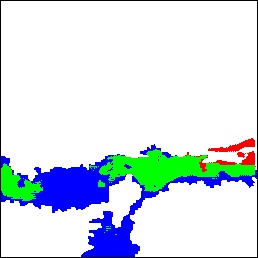}}
		\vspace{1pt}
		\centerline{\includegraphics[width=\textwidth]{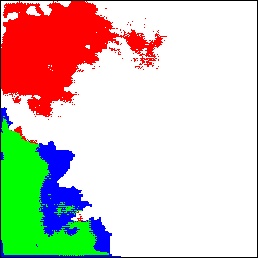}}
		\vspace{1pt}
		\centerline{\includegraphics[width=\textwidth]{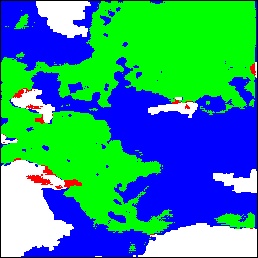}}
		\vspace{1pt}
		\centerline{\includegraphics[width=\textwidth]{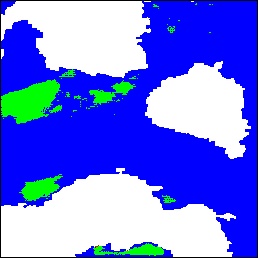}}
		\vspace{1pt}
		\centerline{\includegraphics[width=\textwidth]{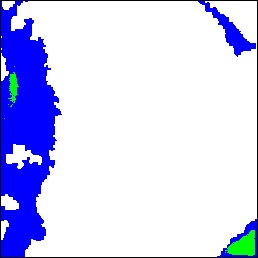}}
		\vspace{1pt}
		\centerline{\includegraphics[width=\textwidth]{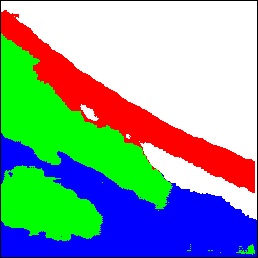}}
		\vspace{1pt}
		\centerline{(j)}
	\end{minipage}
	\hspace{-5pt}
	\begin{minipage}{0.08\linewidth}
		\vspace{1pt}
		\centerline{\includegraphics[width=\textwidth]{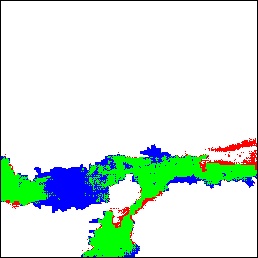}}
		\vspace{1pt}
		\centerline{\includegraphics[width=\textwidth]{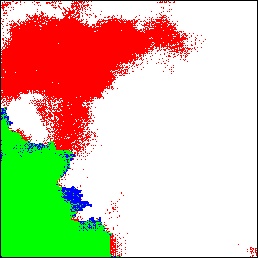}}
		\vspace{1pt}
		\centerline{\includegraphics[width=\textwidth]{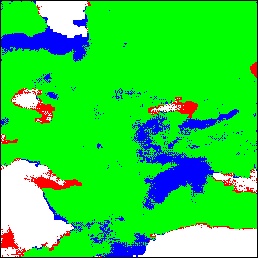}}
		\vspace{1pt}
		\centerline{\includegraphics[width=\textwidth]{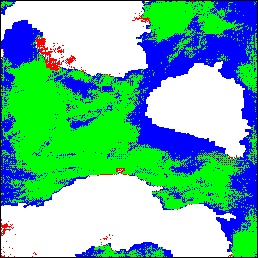}}
		\vspace{1pt}
		\centerline{\includegraphics[width=\textwidth]{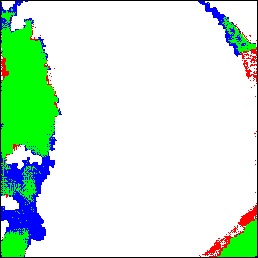}}
		\vspace{1pt}
		\centerline{\includegraphics[width=\textwidth]{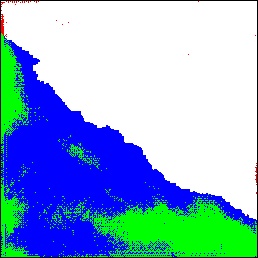}}
		\vspace{1pt}
		\centerline{(k)}
	\end{minipage}
	\hspace{-5pt}
	\begin{minipage}{0.08\linewidth}
		\vspace{1pt}
		\centerline{\includegraphics[width=\textwidth]{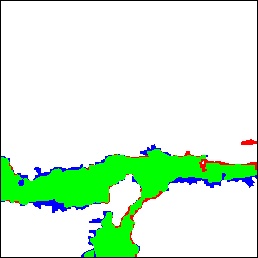}}
		\vspace{1pt}
		\centerline{\includegraphics[width=\textwidth]{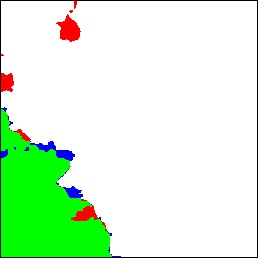}}
		\vspace{1pt}
		\centerline{\includegraphics[width=\textwidth]{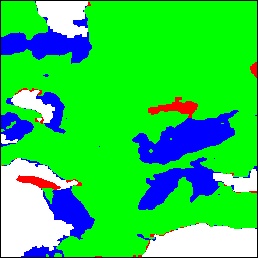}}
		\vspace{1pt}
		\centerline{\includegraphics[width=\textwidth]{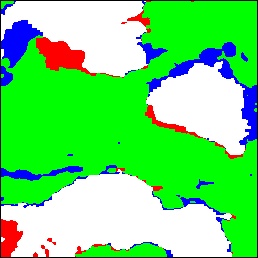}}
		\vspace{1pt}
		\centerline{\includegraphics[width=\textwidth]{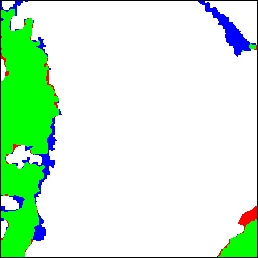}}
		\vspace{1pt}
		\centerline{\includegraphics[width=\textwidth]{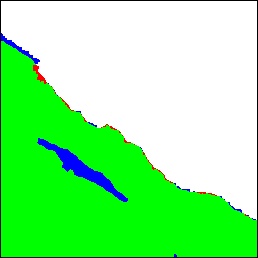}}
		\vspace{1pt}
		\centerline{(l)}
	\end{minipage}
	\caption{Visual comparisons of the proposed method and the state-of-the-art approaches on the GVLM dataset. (a) T1 images. (b) T2 images. (c) Ground truth. (d) A2Net. (e) BIT. (f) ChangeFormer. (g) DMINet. (h) ICIFNet. (i) RDPnet. (j) SiamUnet-diff. (k) SNUNet. (l) MaskCD. The rendered colors represent true positives (green), true negatives (white), false positives (red), and false negatives (blue).} 
	\label{GVLMQuali}
\end{figure*}

\begin{figure*}
	\centering
	\begin{minipage}{0.08\linewidth}
		\vspace{1pt}
		\centerline{\includegraphics[width=\textwidth]{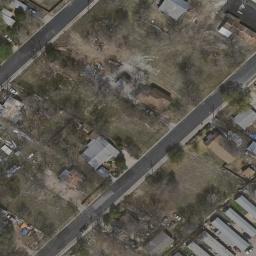}}
		\vspace{1pt}
		\centerline{\includegraphics[width=\textwidth]{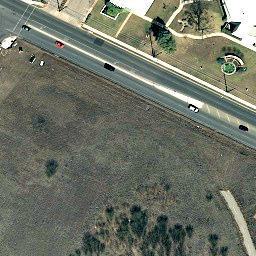}}
		\vspace{1pt}
		\centerline{\includegraphics[width=\textwidth]{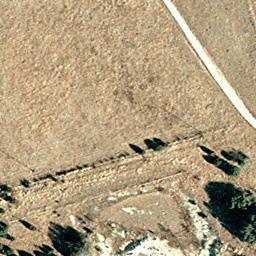}}
		\vspace{1pt}
		\centerline{\includegraphics[width=\textwidth]{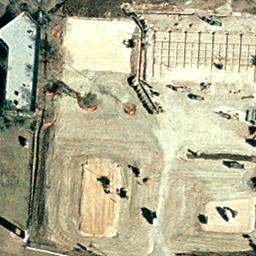}}
		\vspace{1pt}
		\centerline{\includegraphics[width=\textwidth]{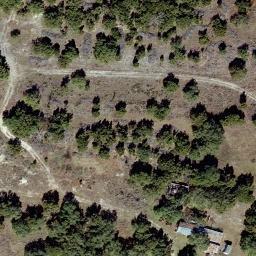}}
		\vspace{1pt}
		\centerline{\includegraphics[width=\textwidth]{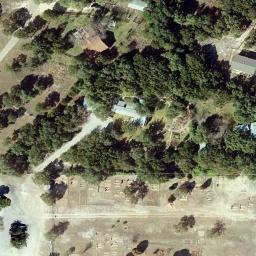}}
		\vspace{1pt}
		\centerline{(a)}
	\end{minipage}
	\hspace{-5pt}
	\begin{minipage}{0.08\linewidth}
		\vspace{1pt}
		\centerline{\includegraphics[width=\textwidth]{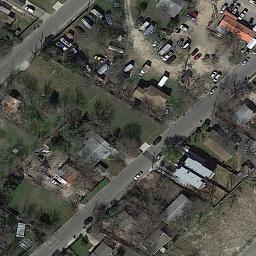}}
		\vspace{1pt}
		\centerline{\includegraphics[width=\textwidth]{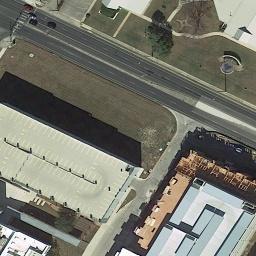}}
		\vspace{1pt}
		\centerline{\includegraphics[width=\textwidth]{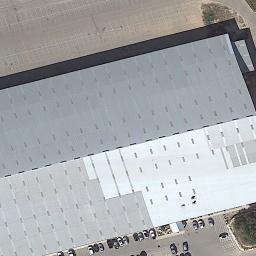}}
		\vspace{1pt}
		\centerline{\includegraphics[width=\textwidth]{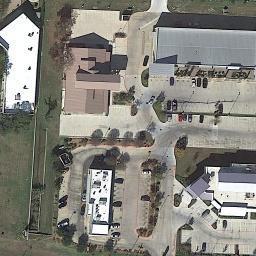}}
		\vspace{1pt}
		\centerline{\includegraphics[width=\textwidth]{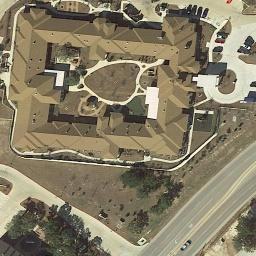}}
		\vspace{1pt}
		\centerline{\includegraphics[width=\textwidth]{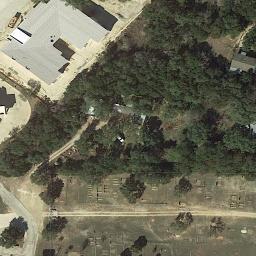}}
		\vspace{1pt}
		\centerline{(b)}
	\end{minipage}
	\hspace{-5pt}
	\begin{minipage}{0.08\linewidth}
		\vspace{1pt}
		\centerline{\includegraphics[width=\textwidth]{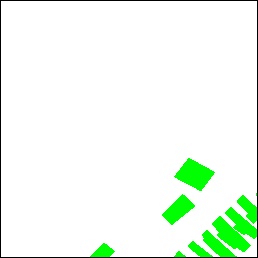}}
		\vspace{1pt}
		\centerline{\includegraphics[width=\textwidth]{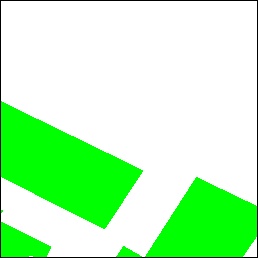}}
		\vspace{1pt}
		\centerline{\includegraphics[width=\textwidth]{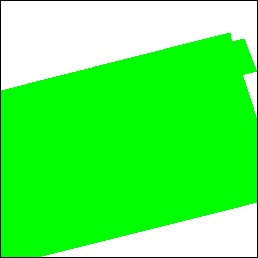}}
		\vspace{1pt}
		\centerline{\includegraphics[width=\textwidth]{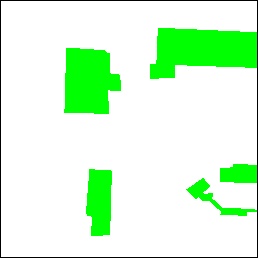}}
		\vspace{1pt}
		\centerline{\includegraphics[width=\textwidth]{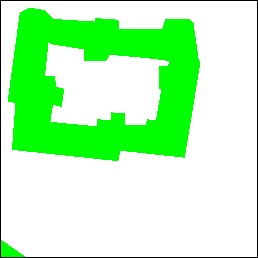}}
		\vspace{1pt}
		\centerline{\includegraphics[width=\textwidth]{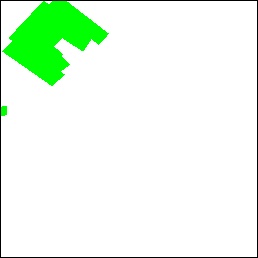}}
		\vspace{1pt}
		\centerline{(c)}
	\end{minipage}
	\hspace{-5pt}
	\begin{minipage}{0.08\linewidth}
		\vspace{1pt}
		\centerline{\includegraphics[width=\textwidth]{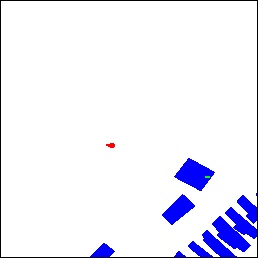}}
		\vspace{1pt}
		\centerline{\includegraphics[width=\textwidth]{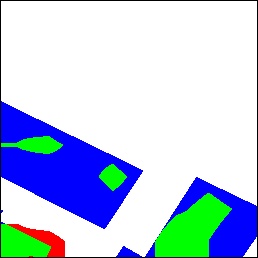}}
		\vspace{1pt}
		\centerline{\includegraphics[width=\textwidth]{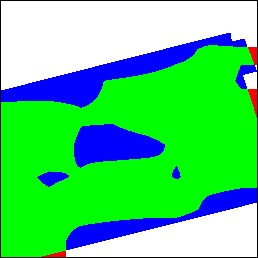}}
		\vspace{1pt}
		\centerline{\includegraphics[width=\textwidth]{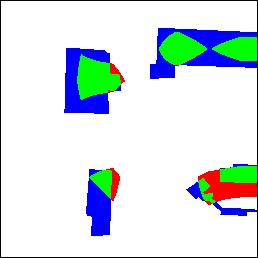}}
		\vspace{1pt}
		\centerline{\includegraphics[width=\textwidth]{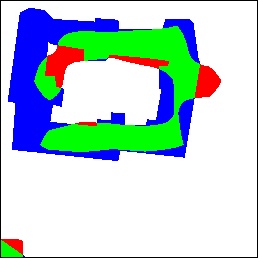}}
		\vspace{1pt}
		\centerline{\includegraphics[width=\textwidth]{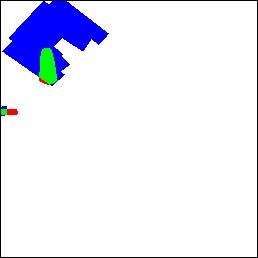}}
		\vspace{1pt}
		\centerline{(d)}
	\end{minipage}
	\hspace{-5pt}
	\begin{minipage}{0.08\linewidth}
		\vspace{1pt}
		\centerline{\includegraphics[width=\textwidth]{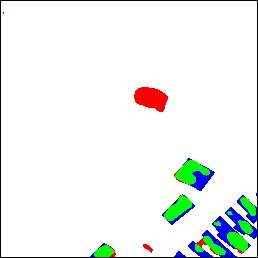}}
		\vspace{1pt}
		\centerline{\includegraphics[width=\textwidth]{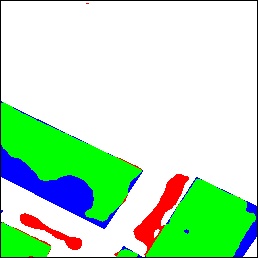}}
		\vspace{1pt}
		\centerline{\includegraphics[width=\textwidth]{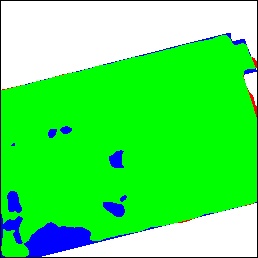}}
		\vspace{1pt}
		\centerline{\includegraphics[width=\textwidth]{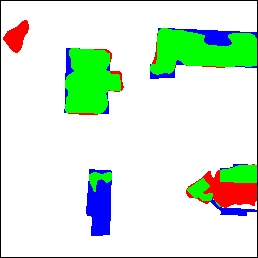}}
		\vspace{1pt}
		\centerline{\includegraphics[width=\textwidth]{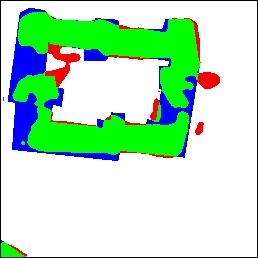}}
		\vspace{1pt}
		\centerline{\includegraphics[width=\textwidth]{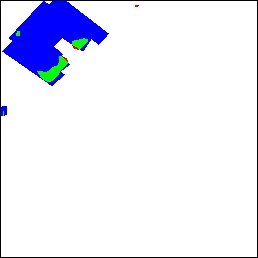}}
		\vspace{1pt}
		\centerline{(e)}
	\end{minipage}
	\hspace{-5pt}
	\begin{minipage}{0.08\linewidth}
		\vspace{1pt}
		\centerline{\includegraphics[width=\textwidth]{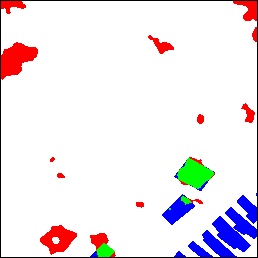}}
		\vspace{1pt}
		\centerline{\includegraphics[width=\textwidth]{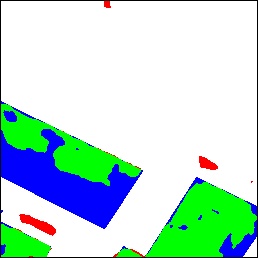}}
		\vspace{1pt}
		\centerline{\includegraphics[width=\textwidth]{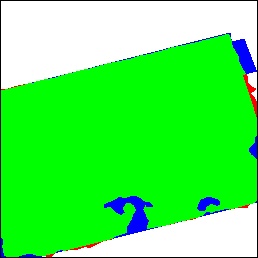}}
		\vspace{1pt}
		\centerline{\includegraphics[width=\textwidth]{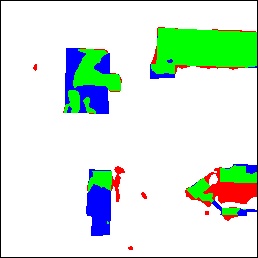}}
		\vspace{1pt}
		\centerline{\includegraphics[width=\textwidth]{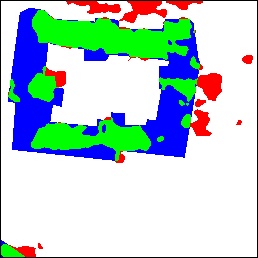}}
		\vspace{1pt}
		\centerline{\includegraphics[width=\textwidth]{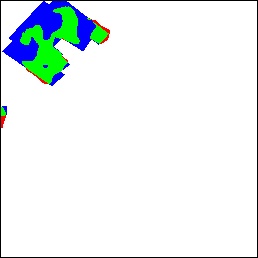}}
		\vspace{1pt}
		\centerline{(f)}
	\end{minipage}
	\hspace{-5pt}
	\begin{minipage}{0.08\linewidth}
		\vspace{1pt}
		\centerline{\includegraphics[width=\textwidth]{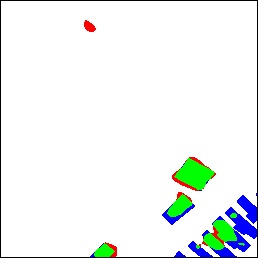}}
		\vspace{1pt}
		\centerline{\includegraphics[width=\textwidth]{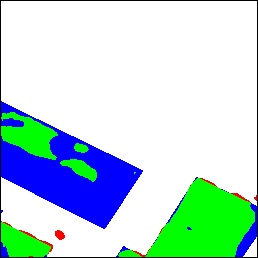}}
		\vspace{1pt}
		\centerline{\includegraphics[width=\textwidth]{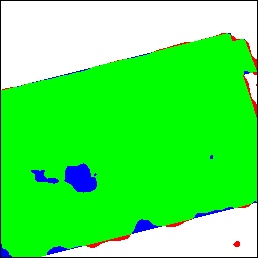}}
		\vspace{1pt}
		\centerline{\includegraphics[width=\textwidth]{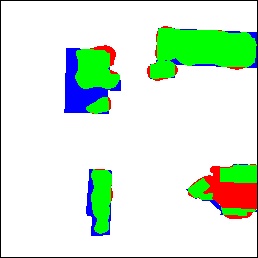}}
		\vspace{1pt}
		\centerline{\includegraphics[width=\textwidth]{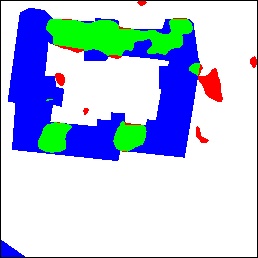}}
		\vspace{1pt}
		\centerline{\includegraphics[width=\textwidth]{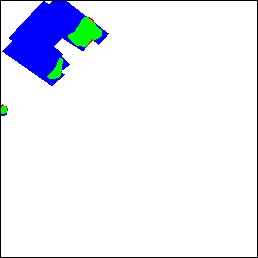}}
		\vspace{1pt}
		\centerline{(g)}
	\end{minipage}
	\hspace{-5pt}
	\begin{minipage}{0.08\linewidth}
		\vspace{1pt}
		\centerline{\includegraphics[width=\textwidth]{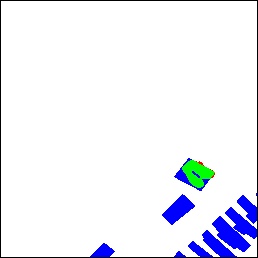}}
		\vspace{1pt}
		\centerline{\includegraphics[width=\textwidth]{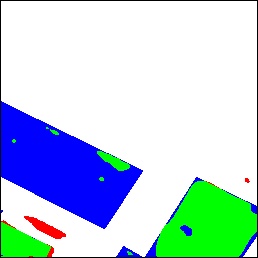}}
		\vspace{1pt}
		\centerline{\includegraphics[width=\textwidth]{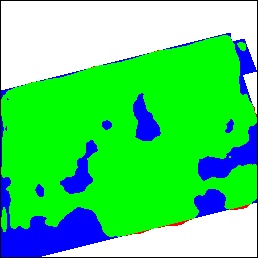}}
		\vspace{1pt}
		\centerline{\includegraphics[width=\textwidth]{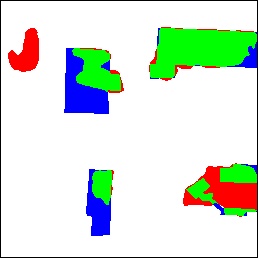}}
		\vspace{1pt}
		\centerline{\includegraphics[width=\textwidth]{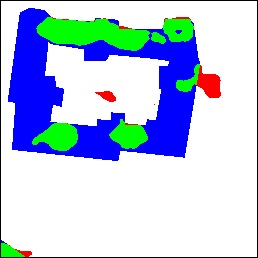}}
		\vspace{1pt}
		\centerline{\includegraphics[width=\textwidth]{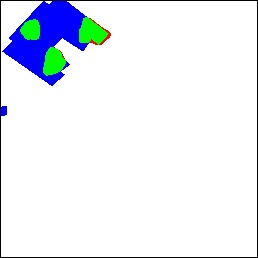}}
		\vspace{1pt}
		\centerline{(h)}
	\end{minipage}
	\hspace{-5pt}
	\begin{minipage}{0.08\linewidth}
		\vspace{1pt}
		\centerline{\includegraphics[width=\textwidth]{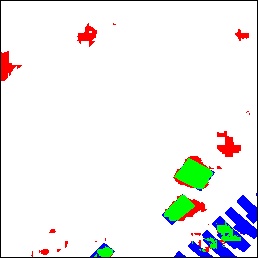}}
		\vspace{1pt}
		\centerline{\includegraphics[width=\textwidth]{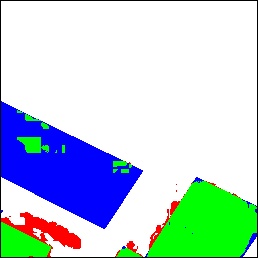}}
		\vspace{1pt}
		\centerline{\includegraphics[width=\textwidth]{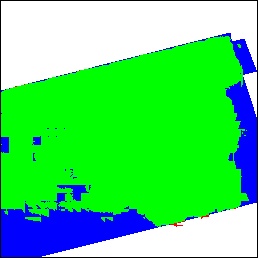}}
		\vspace{1pt}
		\centerline{\includegraphics[width=\textwidth]{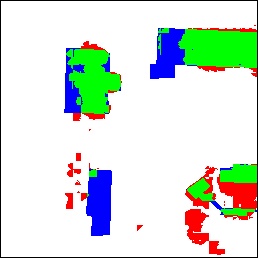}}
		\vspace{1pt}
		\centerline{\includegraphics[width=\textwidth]{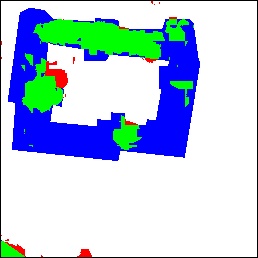}}
		\vspace{1pt}
		\centerline{\includegraphics[width=\textwidth]{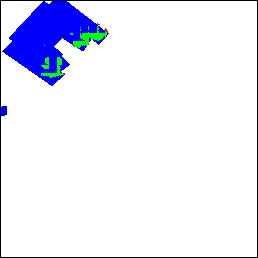}}
		\vspace{1pt}
		\centerline{(i)}
	\end{minipage}
	\hspace{-5pt}
	\begin{minipage}{0.08\linewidth}
		\vspace{1pt}
		\centerline{\includegraphics[width=\textwidth]{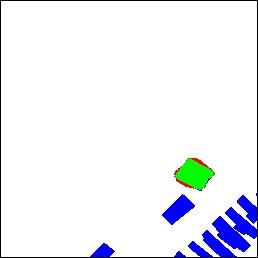}}
		\vspace{1pt}
		\centerline{\includegraphics[width=\textwidth]{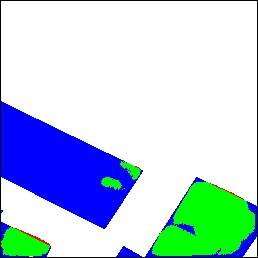}}
		\vspace{1pt}
		\centerline{\includegraphics[width=\textwidth]{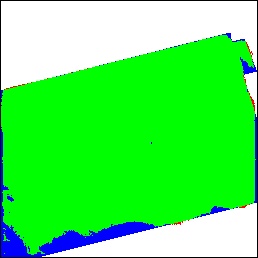}}
		\vspace{1pt}
		\centerline{\includegraphics[width=\textwidth]{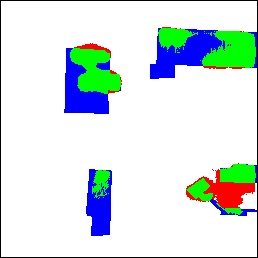}}
		\vspace{1pt}
		\centerline{\includegraphics[width=\textwidth]{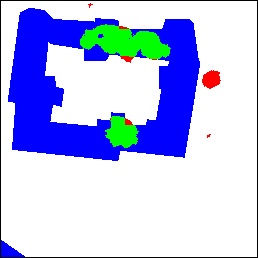}}
		\vspace{1pt}
		\centerline{\includegraphics[width=\textwidth]{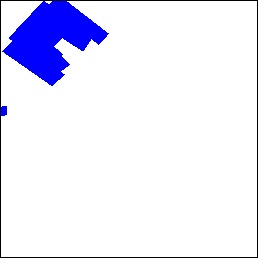}}
		\vspace{1pt}
		\centerline{(j)}
	\end{minipage}
	\hspace{-5pt}
	\begin{minipage}{0.08\linewidth}
		\vspace{1pt}
		\centerline{\includegraphics[width=\textwidth]{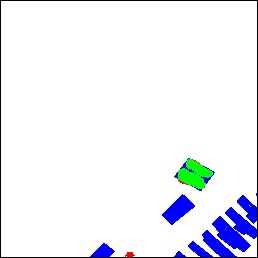}}
		\vspace{1pt}
		\centerline{\includegraphics[width=\textwidth]{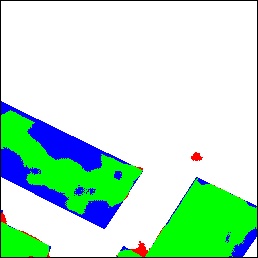}}
		\vspace{1pt}
		\centerline{\includegraphics[width=\textwidth]{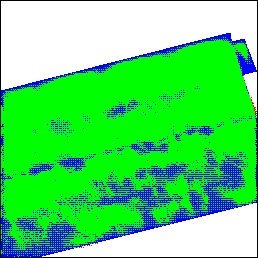}}
		\vspace{1pt}
		\centerline{\includegraphics[width=\textwidth]{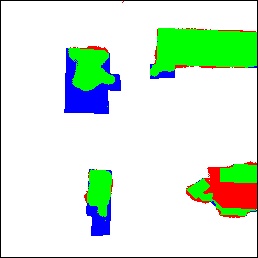}}
		\vspace{1pt}
		\centerline{\includegraphics[width=\textwidth]{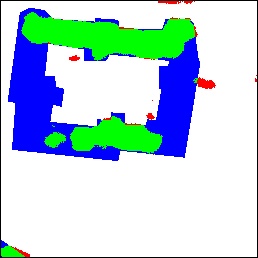}}
		\vspace{1pt}
		\centerline{\includegraphics[width=\textwidth]{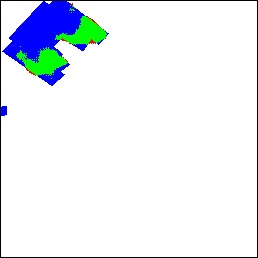}}
		\vspace{1pt}
		\centerline{(k)}
	\end{minipage}
	\hspace{-5pt}
	\begin{minipage}{0.08\linewidth}
		\vspace{1pt}
		\centerline{\includegraphics[width=\textwidth]{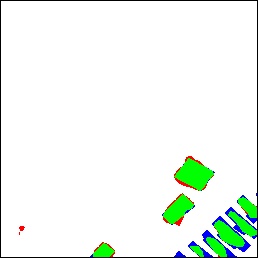}}
		\vspace{1pt}
		\centerline{\includegraphics[width=\textwidth]{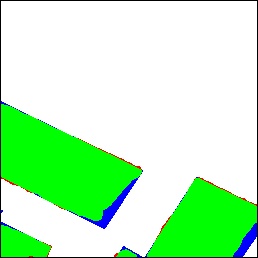}}
		\vspace{1pt}
		\centerline{\includegraphics[width=\textwidth]{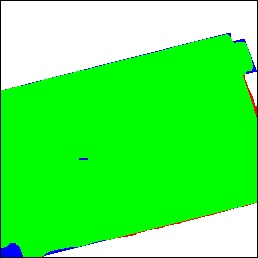}}
		\vspace{1pt}
		\centerline{\includegraphics[width=\textwidth]{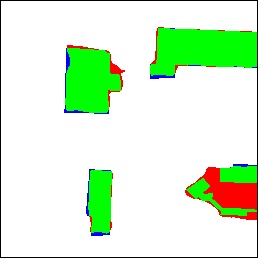}}
		\vspace{1pt}
		\centerline{\includegraphics[width=\textwidth]{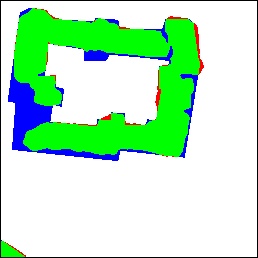}}
		\vspace{1pt}
		\centerline{\includegraphics[width=\textwidth]{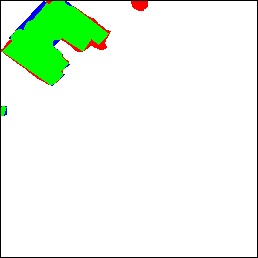}}
		\vspace{1pt}
		\centerline{(l)}
	\end{minipage}
	\caption{Visual comparisons of the proposed method and the state-of-the-art approaches on the LEVIRCD dataset. (a) T1 images. (b) T2 images. (c) Ground truth. (d) A2Net. (e) BIT. (f) ChangeFormer. (g) DMINet. (h) ICIFNet. (i) RDPnet. (j) SiamUnet-diff. (k) SNUNet. (l) MaskCD. The rendered colors represent true positives (green), true negatives (white), false positives (red), and false negatives (blue).} 
	\label{LEVIRCDQuali}
\end{figure*}

\begin{figure*}
	\centering
	\begin{minipage}{0.08\linewidth}
		\vspace{1pt}
		\centerline{\includegraphics[width=\textwidth]{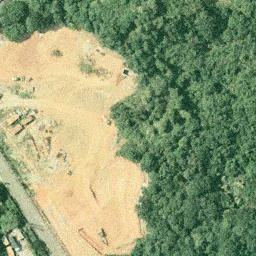}}
		\vspace{1pt}
		\centerline{\includegraphics[width=\textwidth]{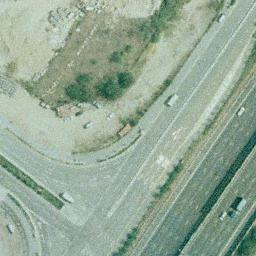}}
		\vspace{1pt}
		\centerline{\includegraphics[width=\textwidth]{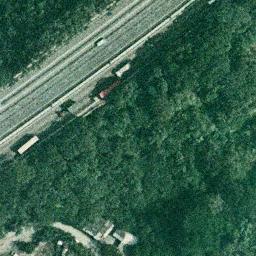}}
		\vspace{1pt}
		\centerline{\includegraphics[width=\textwidth]{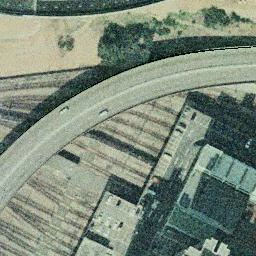}}
		\vspace{1pt}
		\centerline{\includegraphics[width=\textwidth]{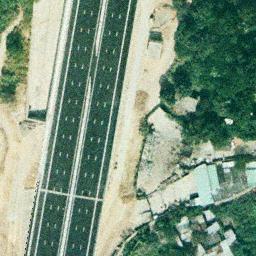}}
		\vspace{1pt}
		\centerline{\includegraphics[width=\textwidth]{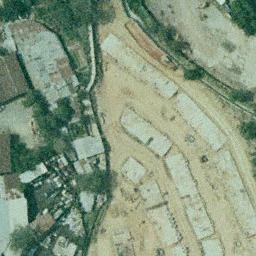}}
		\vspace{1pt}
		\centerline{(a)}
	\end{minipage}
	\hspace{-5pt}
	\begin{minipage}{0.08\linewidth}
		\vspace{1pt}
		\centerline{\includegraphics[width=\textwidth]{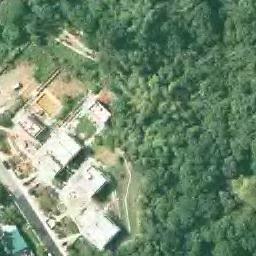}}
		\vspace{1pt}
		\centerline{\includegraphics[width=\textwidth]{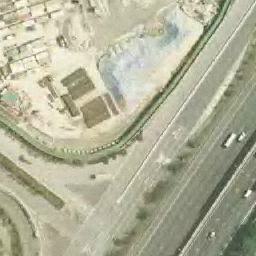}}
		\vspace{1pt}
		\centerline{\includegraphics[width=\textwidth]{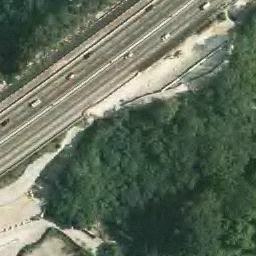}}
		\vspace{1pt}
		\centerline{\includegraphics[width=\textwidth]{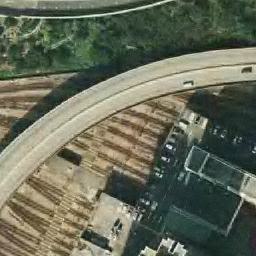}}
		\vspace{1pt}
		\centerline{\includegraphics[width=\textwidth]{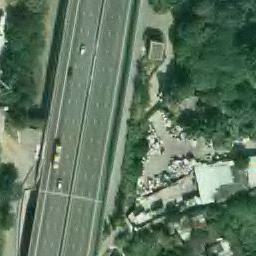}}
		\vspace{1pt}
		\centerline{\includegraphics[width=\textwidth]{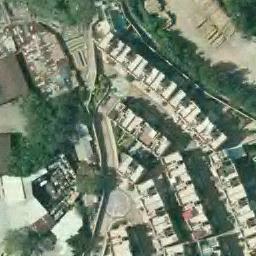}}
		\vspace{1pt}
		\centerline{(b)}
	\end{minipage}
	\hspace{-5pt}
	\begin{minipage}{0.08\linewidth}
		\vspace{1pt}
		\centerline{\includegraphics[width=\textwidth]{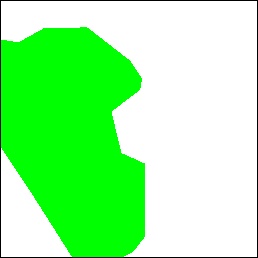}}
		\vspace{1pt}
		\centerline{\includegraphics[width=\textwidth]{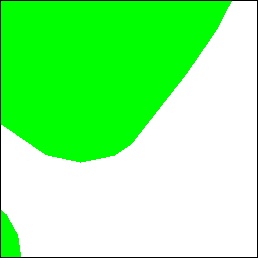}}
		\vspace{1pt}
		\centerline{\includegraphics[width=\textwidth]{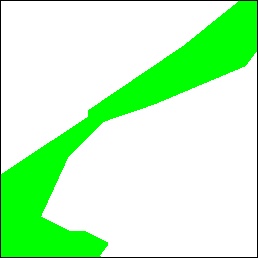}}
		\vspace{1pt}
		\centerline{\includegraphics[width=\textwidth]{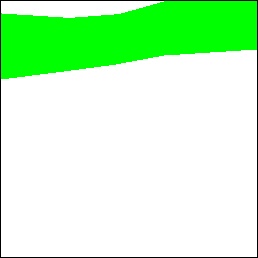}}
		\vspace{1pt}
		\centerline{\includegraphics[width=\textwidth]{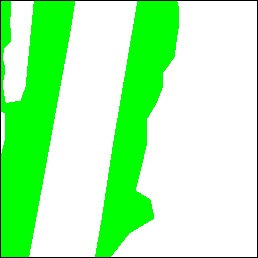}}
		\vspace{1pt}
		\centerline{\includegraphics[width=\textwidth]{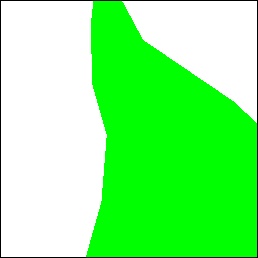}}
		\vspace{1pt}
		\centerline{(c)}
	\end{minipage}
	\hspace{-5pt}
	\begin{minipage}{0.08\linewidth}
		\vspace{1pt}
		\centerline{\includegraphics[width=\textwidth]{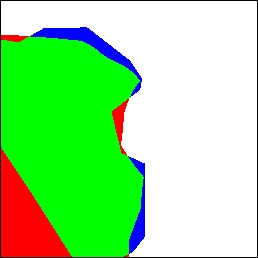}}
		\vspace{1pt}
		\centerline{\includegraphics[width=\textwidth]{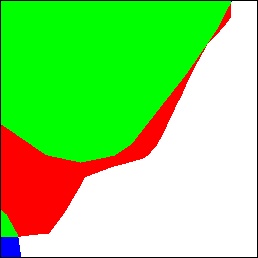}}
		\vspace{1pt}
		\centerline{\includegraphics[width=\textwidth]{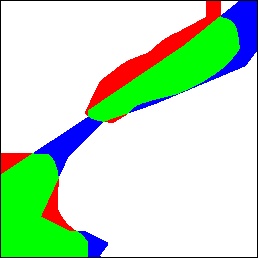}}
		\vspace{1pt}
		\centerline{\includegraphics[width=\textwidth]{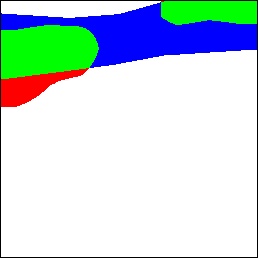}}
		\vspace{1pt}
		\centerline{\includegraphics[width=\textwidth]{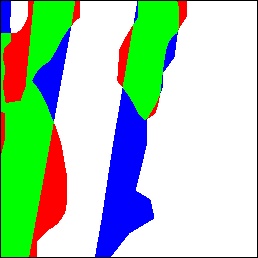}}
		\vspace{1pt}
		\centerline{\includegraphics[width=\textwidth]{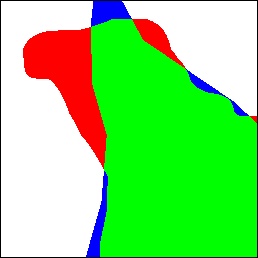}}
		\vspace{1pt}
		\centerline{(d)}
	\end{minipage}
	\hspace{-5pt}
	\begin{minipage}{0.08\linewidth}
		\vspace{1pt}
		\centerline{\includegraphics[width=\textwidth]{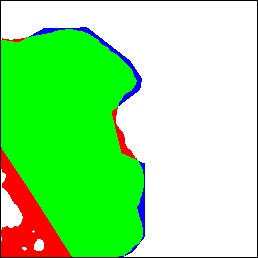}}
		\vspace{1pt}
		\centerline{\includegraphics[width=\textwidth]{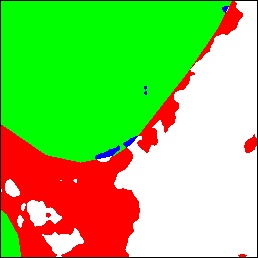}}
		\vspace{1pt}
		\centerline{\includegraphics[width=\textwidth]{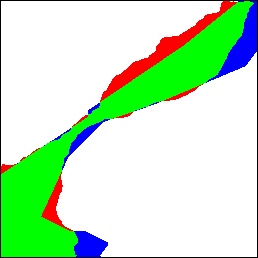}}
		\vspace{1pt}
		\centerline{\includegraphics[width=\textwidth]{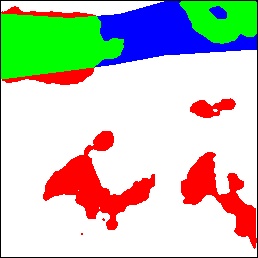}}
		\vspace{1pt}
		\centerline{\includegraphics[width=\textwidth]{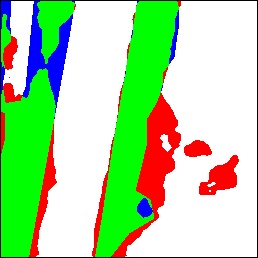}}
		\vspace{1pt}
		\centerline{\includegraphics[width=\textwidth]{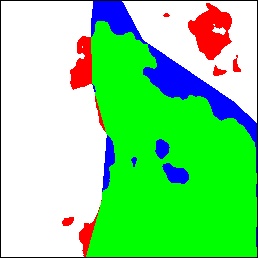}}
		\vspace{1pt}
		\centerline{(e)}
	\end{minipage}
	\hspace{-5pt}
	\begin{minipage}{0.08\linewidth}
		\vspace{1pt}
		\centerline{\includegraphics[width=\textwidth]{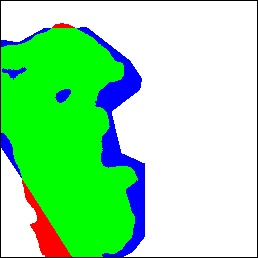}}
		\vspace{1pt}
		\centerline{\includegraphics[width=\textwidth]{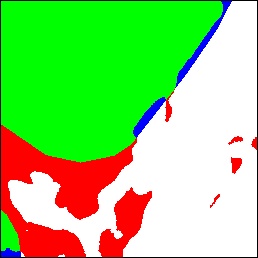}}
		\vspace{1pt}
		\centerline{\includegraphics[width=\textwidth]{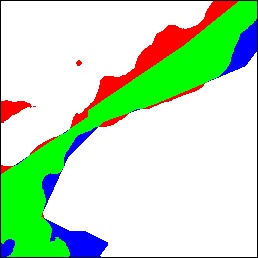}}
		\vspace{1pt}
		\centerline{\includegraphics[width=\textwidth]{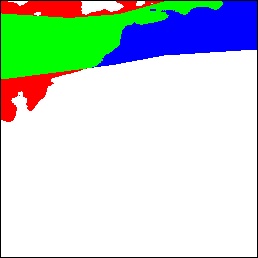}}
		\vspace{1pt}
		\centerline{\includegraphics[width=\textwidth]{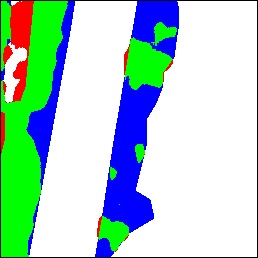}}
		\vspace{1pt}
		\centerline{\includegraphics[width=\textwidth]{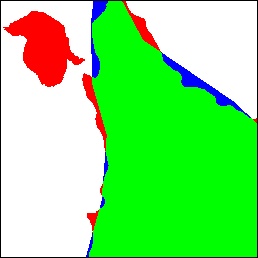}}
		\vspace{1pt}
		\centerline{(f)}
	\end{minipage}
	\hspace{-5pt}
	\begin{minipage}{0.08\linewidth}
		\vspace{1pt}
		\centerline{\includegraphics[width=\textwidth]{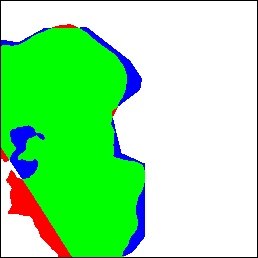}}
		\vspace{1pt}
		\centerline{\includegraphics[width=\textwidth]{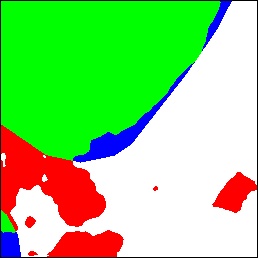}}
		\vspace{1pt}
		\centerline{\includegraphics[width=\textwidth]{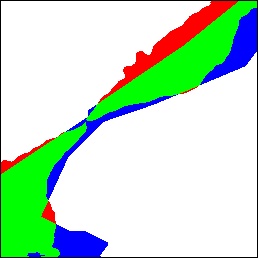}}
		\vspace{1pt}
		\centerline{\includegraphics[width=\textwidth]{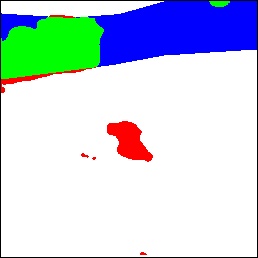}}
		\vspace{1pt}
		\centerline{\includegraphics[width=\textwidth]{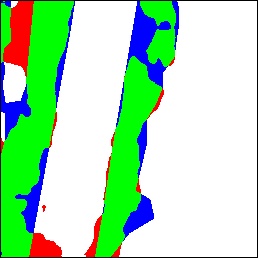}}
		\vspace{1pt}
		\centerline{\includegraphics[width=\textwidth]{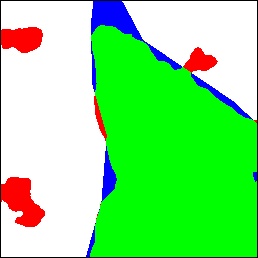}}
		\vspace{1pt}
		\centerline{(g)}
	\end{minipage}
	\hspace{-5pt}
	\begin{minipage}{0.08\linewidth}
		\vspace{1pt}
		\centerline{\includegraphics[width=\textwidth]{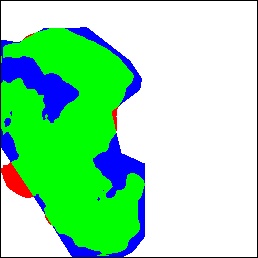}}
		\vspace{1pt}
		\centerline{\includegraphics[width=\textwidth]{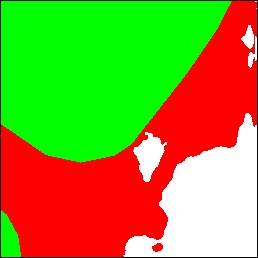}}
		\vspace{1pt}
		\centerline{\includegraphics[width=\textwidth]{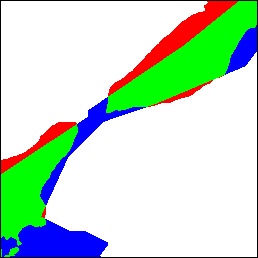}}
		\vspace{1pt}
		\centerline{\includegraphics[width=\textwidth]{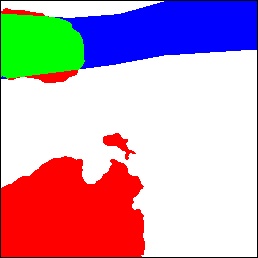}}
		\vspace{1pt}
		\centerline{\includegraphics[width=\textwidth]{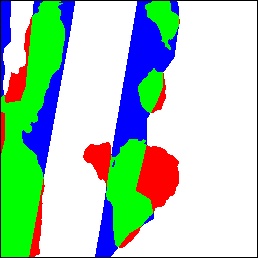}}
		\vspace{1pt}
		\centerline{\includegraphics[width=\textwidth]{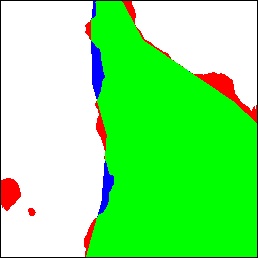}}
		\vspace{1pt}
		\centerline{(h)}
	\end{minipage}
	\hspace{-5pt}
	\begin{minipage}{0.08\linewidth}
		\vspace{1pt}
		\centerline{\includegraphics[width=\textwidth]{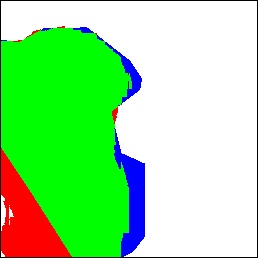}}
		\vspace{1pt}
		\centerline{\includegraphics[width=\textwidth]{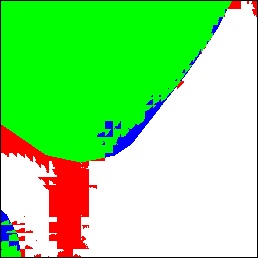}}
		\vspace{1pt}
		\centerline{\includegraphics[width=\textwidth]{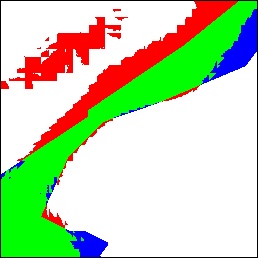}}
		\vspace{1pt}
		\centerline{\includegraphics[width=\textwidth]{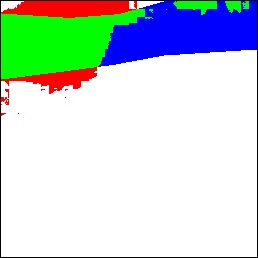}}
		\vspace{1pt}
		\centerline{\includegraphics[width=\textwidth]{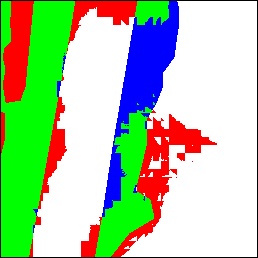}}
		\vspace{1pt}
		\centerline{\includegraphics[width=\textwidth]{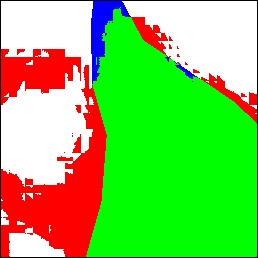}}
		\vspace{1pt}
		\centerline{(i)}
	\end{minipage}
	\hspace{-5pt}
	\begin{minipage}{0.08\linewidth}
		\vspace{1pt}
		\centerline{\includegraphics[width=\textwidth]{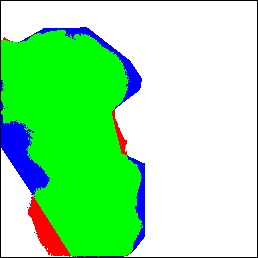}}
		\vspace{1pt}
		\centerline{\includegraphics[width=\textwidth]{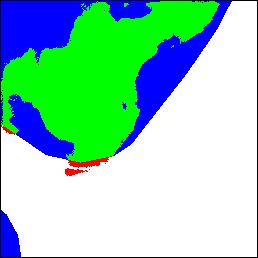}}
		\vspace{1pt}
		\centerline{\includegraphics[width=\textwidth]{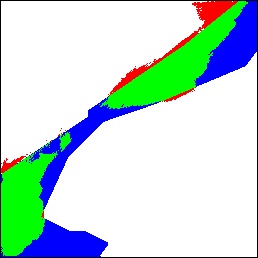}}
		\vspace{1pt}
		\centerline{\includegraphics[width=\textwidth]{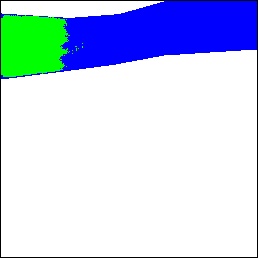}}
		\vspace{1pt}
		\centerline{\includegraphics[width=\textwidth]{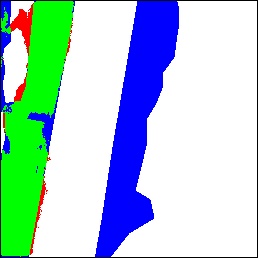}}
		\vspace{1pt}
		\centerline{\includegraphics[width=\textwidth]{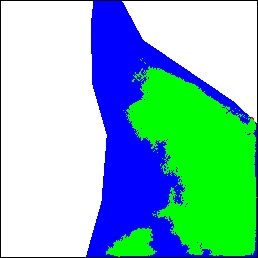}}
		\vspace{1pt}
		\centerline{(j)}
	\end{minipage}
	\hspace{-5pt}
	\begin{minipage}{0.08\linewidth}
		\vspace{1pt}
		\centerline{\includegraphics[width=\textwidth]{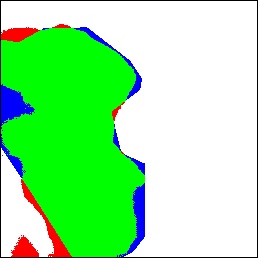}}
		\vspace{1pt}
		\centerline{\includegraphics[width=\textwidth]{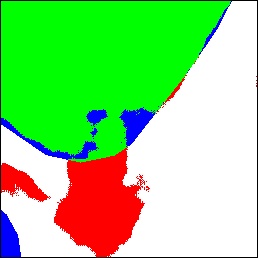}}
		\vspace{1pt}
		\centerline{\includegraphics[width=\textwidth]{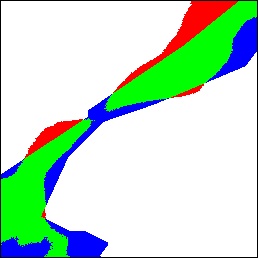}}
		\vspace{1pt}
		\centerline{\includegraphics[width=\textwidth]{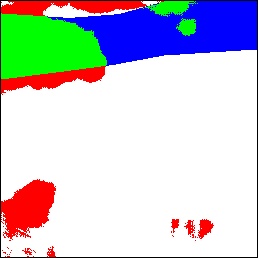}}
		\vspace{1pt}
		\centerline{\includegraphics[width=\textwidth]{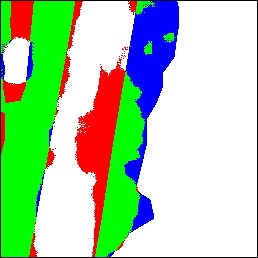}}
		\vspace{1pt}
		\centerline{\includegraphics[width=\textwidth]{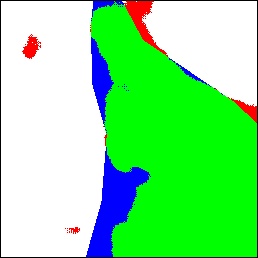}}
		\vspace{1pt}
		\centerline{(k)}
	\end{minipage}
	\hspace{-5pt}
	\begin{minipage}{0.08\linewidth}
		\vspace{1pt}
		\centerline{\includegraphics[width=\textwidth]{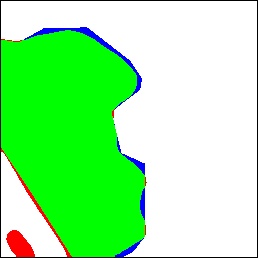}}
		\vspace{1pt}
		\centerline{\includegraphics[width=\textwidth]{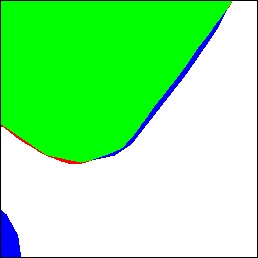}}
		\vspace{1pt}
		\centerline{\includegraphics[width=\textwidth]{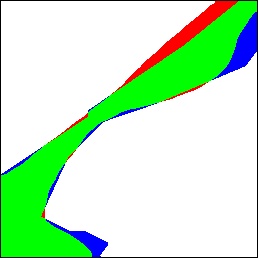}}
		\vspace{1pt}
		\centerline{\includegraphics[width=\textwidth]{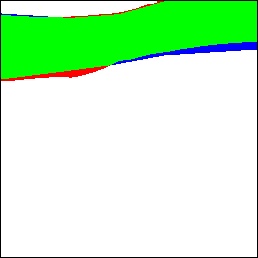}}
		\vspace{1pt}
		\centerline{\includegraphics[width=\textwidth]{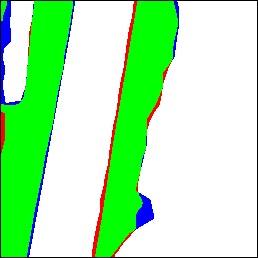}}
		\vspace{1pt}
		\centerline{\includegraphics[width=\textwidth]{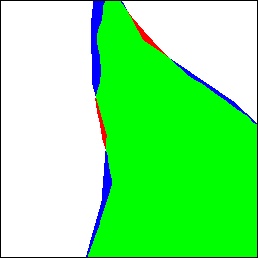}}
		\vspace{1pt}
		\centerline{(l)}
	\end{minipage}
	\caption{Visual comparisons of the proposed method and the state-of-the-art approaches on the SYSUCD dataset. (a) T1 images. (b) T2 images. (c) Ground truth. (d) A2Net. (e) BIT. (f) ChangeFormer. (g) DMINet. (h) ICIFNet. (i) RDPnet. (j) SiamUnet-diff. (k) SNUNet. (l) MaskCD. The rendered colors represent true positives (green), true negatives (white), false positives (red), and false negatives (blue).} 
	\label{SYSUQuali}
\end{figure*}

The visual comparisons of different methods on five RS-CD datasets are shown in Figs. \ref{CLCDQuali}-\ref{SYSUQuali}.  According to these representative results, we can observe that the proposed MaskCD method exhibits dominance in the following aspects:
\begin{itemize}
    \item Advantages in object integrity:
    In the second row of Fig. \ref{CLCDQuali}, the second row of Fig. \ref{EGYBCDQuali}, and the fourth row of Fig. \ref{GVLMQuali}, the main body of the changed objects cannot be correctly detected by the comparison methods, while the proposed MaskCD managed to detect the change objects with the best integrity. It lies in the fact that the proposed MaskCD focuses on mask generation and classification in parallel based on object-level representation learning rather than per-pixel classification.  Therefore, the integrity of changed objects can be preserved in the CD masks, leveraging the feature reconstruction capability of the MA-DETR module. 
    \item Advantages in reducing pseudo-changes:
    Compared with other methods, the proposed MaskCD can significantly reduce the pseudo-changes that are irrelevant to the specific types of changes in the datasets. In the second and sixth rows of Fig. \ref{GVLMQuali}, the comparison methods detected many pseudo-changes caused by spectral similarity with landslides denoted as the red patches in the change maps, while the proposed MaskCD can reduce false alarms. This is mainly due to the usage of the masked attention mechanism in the MA-DETR module, which can accurately compress the background noises and focus only on the desired change objects, resulting in a successful avoidance of pseudo-changes appearing in background areas. 
    \item Advantages in classifying small targets:
    The changed objects usually vary in scale and spectral characteristics, which increases the difficulty of correctly detecting and classifying different scenes, especially for small targets. As shown in the first, fourth, and sixth rows of Fig. \ref{LEVIRCDQuali}, the buildings of smaller scales fall in a poor detection from the results of comparison methods, while they are well retained and classified by the proposed MaskCD. These advantages are benefited by the multiscale self-attention mechanism employed in the MA-DETR modules, which can understand the change objects more precisely by capturing the long-range dependencies of the objects. In addition, by leveraging the strategy of mask classification, the changed objects can be more accurately detected. 
    
\end{itemize}
% \usepackage{multirow}

\subsection{Ablation Studies}
To verify the effectiveness of the components and configurations of the proposed MaskCD model, we conduct comprehensive ablation studies on EGY-BCD datasets. \ref{allinonetb}. The quantitative results of the ablation studies are displayed in Table. \ref{table:ablation_studies}.

\begin{table*}[htb]
\centering
\caption{Quantitative comparisons of the proposed method with diverse settings on EGY-BCD dataset, in terms of OA, precision, recall, F1, and mIoU.}
\label{table:ablation_studies}
\begin{tabular}{m{1cm}<{\centering}m{3.3cm}<{\centering}cm{1.2cm}<{\centering}m{1.2cm}<{\centering}m{1.2cm}<{\centering}m{1.2cm}<{\centering}m{1.2cm}<{\centering}}
\toprule
\multicolumn{1}{c|}{\multirow{2}{*}{No.}} & \multicolumn{1}{c|}{\multirow{2}{*}{Variants}} & \multicolumn{5}{c}{Metrics}                \\ \cmidrule{3-7}
\multicolumn{1}{c|}{}                     & \multicolumn{1}{c|}{}                          & OA     & Pre    & Rec    & F1     & mIoU   \\ \midrule
\multicolumn{1}{c|}{\#01}                 & \multicolumn{1}{c|}{MaskCD}                    & 0.9824 & 0.8844 & 0.8370 & 0.8598 & 0.8678 \\ \midrule
\multicolumn{7}{l}{DeformMHSA (DeformMHSA) and Masked Attention (MA) Modules}                                                                \\ \midrule
\multicolumn{1}{c|}{\#02}                 & \multicolumn{1}{c|}{w/0 DeformMHSA}                 & 0.9809 & 0.8572 & 0.8420 & 0.8493 & 0.8592 \\
\multicolumn{1}{c|}{\#03}                 & \multicolumn{1}{c|}{w/0 MA}                    & 0.9760 & 0.8226 & 0.7989 & 0.8103 & 0.8280 \\
\multicolumn{1}{c|}{\#04}                 & \multicolumn{1}{c|}{w/0 DeformMHSA and MA}          & 0.9738 & 0.8768 & 0.6856 & 0.7695 & 0.7990 \\ \midrule
\multicolumn{7}{l}{Different Numbers of Query Embeddings (QE)}                                                                          \\ \midrule
\multicolumn{1}{c|}{\#05}                 & \multicolumn{1}{c|}{\#QE=25}                   & 0.9812 & 0.8667 & 0.8367 & 0.8513 & 0.8606 \\
\multicolumn{1}{c|}{\#06}                 & \multicolumn{1}{c|}{\#QE=50}                   & 0.9814 & 0.8741 & 0.8318 & 0.8522 & 0.8615 \\
\multicolumn{1}{c|}{\#07}                 & \multicolumn{1}{c|}{\#QE=100}                   & 0.9817 & 0.8719 & 0.8372 & 0.8542 & 0.8631 \\ 
\multicolumn{1}{c|}{\#08}                 & \multicolumn{1}{c|}{\#QE=125}                   & 0.9815 &	0.8686&	0.8400&	0.8539&	0.8628 \\ \midrule
\multicolumn{7}{l}{Classification Methods}                                                                                              \\ \midrule
\multicolumn{1}{c|}{\#09}                 & \multicolumn{1}{c|}{Per-pixel Classification}  & 0.9757 & 0.8331 & 0.7760 & 0.8036 & 0.8230 \\ \bottomrule
\end{tabular}
\end{table*}

\subsubsection{Effectiveness of DeformMHSA module}
The DeformMHSA is designed to elevate multi-scale bi-temporal deep features into change-aware representations. To assess the efficacy of DeformMHSA, we directly forward multi-scale deep representations into the transformer decoder without DeformMHSA. As illustrated in Table. \ref{table:ablation_studies}, the exclusion of the DeformMHSA module results in a substantial performance decline, underscoring the pivotal role of aggregating multi-level deformable attention in the MaskCD model.
\subsubsection{Effectiveness of masked attention module}
The masked attention mechanism is geared towards compressing background noise, directing focus specifically to foreground-changed and unchanged objects. To evaluate the effectiveness of masked attention, we set $\mathcal{M}=0$, ensuring all spatial features contribute to cross-attention computation. As illustrated in Table. \ref{table:ablation_studies}, the degradation of the masked attention to cross-attention leads to decreased CD performance. Furthermore, when we eliminate both the DeformMHSA and masked attention modules from the MaskCD model, more significant performance degradation can be observed, indicating the collaborative effectiveness of these modules.
\subsubsection{Choices of the number of query embeddings}
The query embeddings can represent the class information and interact with per-pixel embeddings for the generation of masks. To this end, the quantity of query embeddings primarily corresponds to the number of masks predicted during mask classification. To explore the best configuration of this number, we set it to different values ranging from $25$ to $125$ with a step of $25$. It can be found that query embeddings with a length $75$ achieve the best performance with balanced computational cost. According to empirical studies in previous mask classification research, the number of query embeddings is usually determined based on the complexity of the scenes. Since RS images usually contain an extensive amount of objects that can be changed or unchanged in time series, the choice of $75$ query embeddings is reasonable and cost-effective in RS-CD.
\subsubsection{Effectiveness of mask classification}
As a core idea of the MaskCD, the mask classification proposed in this paper prompts a new paradigm for CD. To validate the effectiveness of the mask classification, we replace the transformer model and mask classification module with a $1\times 1$ convolutional layer, which accepts the final layer output of the FPN of the pixel-level module. As illustrated in Table. \ref{table:ablation_studies}, the quantitative results exhibit a significant improvement in mask classification compared with per-pixel classification. Besides, we display the results of this ablation study for qualitative comparison in Fig. \ref{figure:pixel_ablation}. Compared with the results of mask classification, it can be observed that the change maps predicted by per-pixel classification lost some detailed boundaries, especially in the (c)(d) rows of Fig. \ref{figure:pixel_ablation}. In addition, the integrity of the changed objects in per-pixel classification is not well-preserved, and the detected objects are linked together incorrectly, as seen in the (b)(c)(e) rows of Fig. \ref{figure:pixel_ablation}. Finally, it can be found there are more speckle noises in the predictions of per-pixel classification, while the prediction of mask classification seems to be consistent with the actual ground objects. 
\subsubsection{Formations of Predicted Masks}
\begin{figure}
\setlength\tabcolsep{1.5pt}
\settowidth\rotheadsize{Unchanged}
\setkeys{Gin}{width=\hsize}
\begin{tabularx}{\linewidth}{l XXXXXX }% <-- here is determined table width
\rothead{\centering Image}        &   \includegraphics[valign=m]{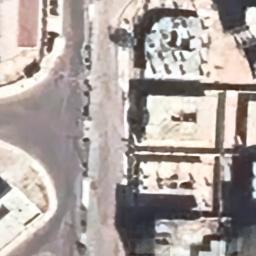}
                        &   \includegraphics[valign=m]{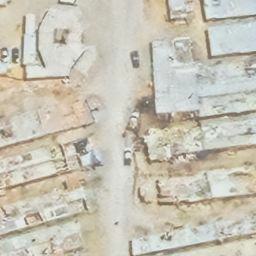}
                        &   \includegraphics[valign=m]{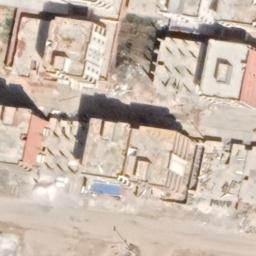}
                        &   \includegraphics[valign=m]{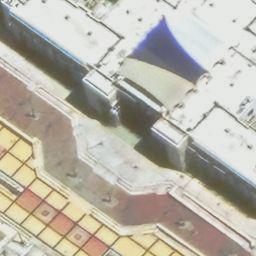}
                        &   \includegraphics[valign=m]{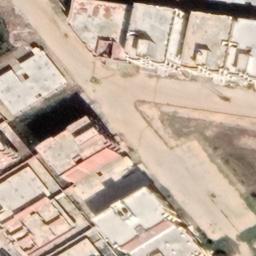}     \\
\rothead{\centering label}        &   \includegraphics[valign=m]{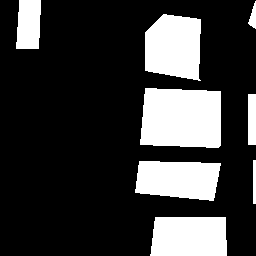}
                        &   \includegraphics[valign=m]{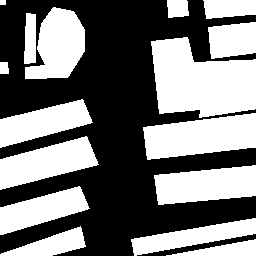}
                        &   \includegraphics[valign=m]{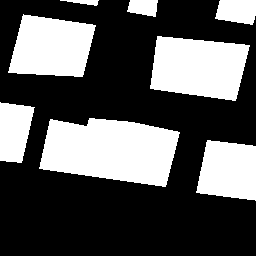}
                        &   \includegraphics[valign=m]{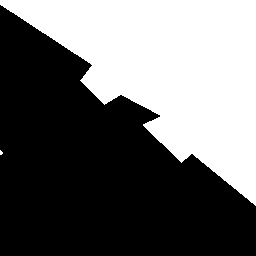}
                        &   \includegraphics[valign=m]{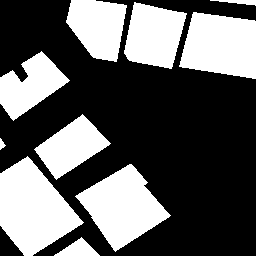}     \\  
\rothead{\centering Changed} &   \includegraphics[valign=m]{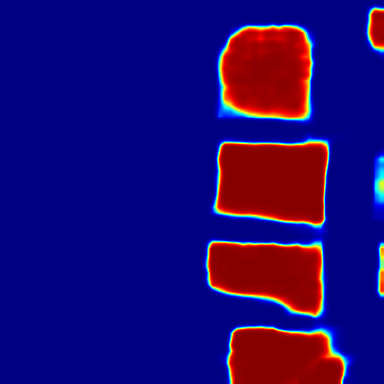}
                        &   \includegraphics[valign=m]{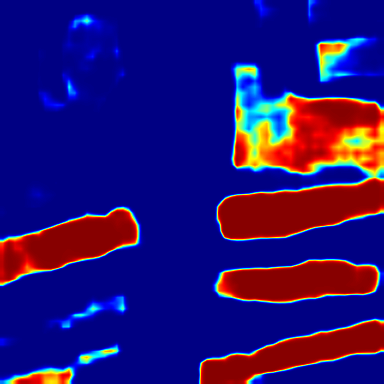}
                        &   \includegraphics[valign=m]{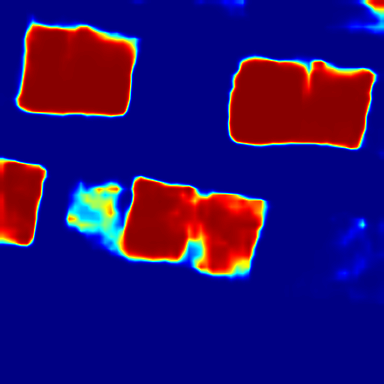}
                        &   \includegraphics[valign=m]{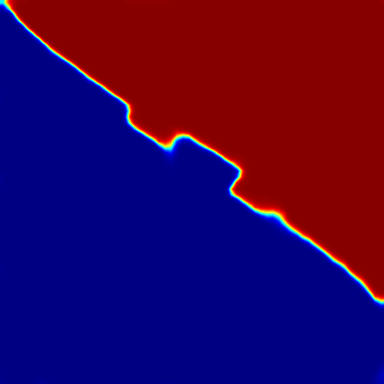}
                        &   \includegraphics[valign=m]{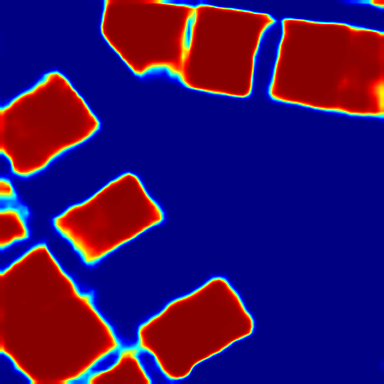}     \\
\rothead{\centering Unchanged} &   \includegraphics[valign=m]{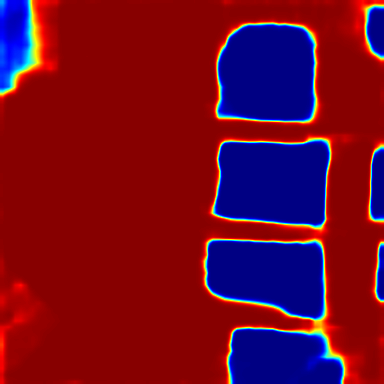} 
                        &   \includegraphics[valign=m]{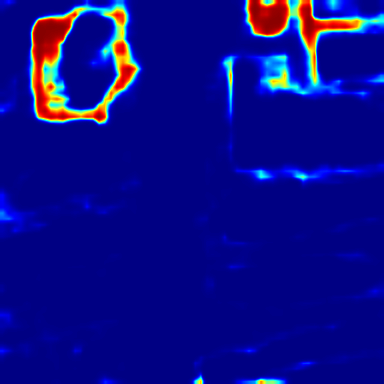} 
                        &   \includegraphics[valign=m]{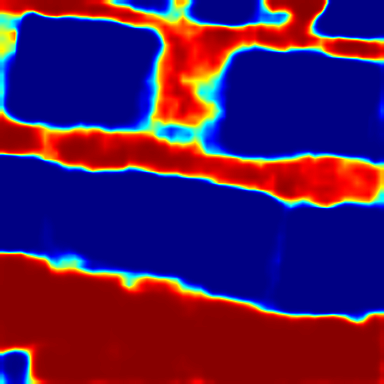} 
                        &   \includegraphics[valign=m]{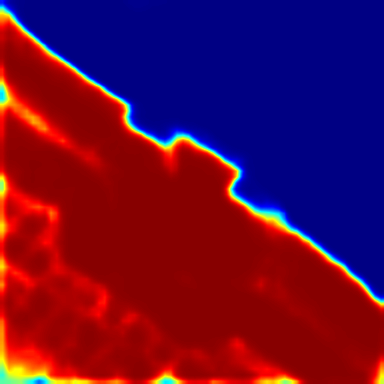} 
                        &   \includegraphics[valign=m]{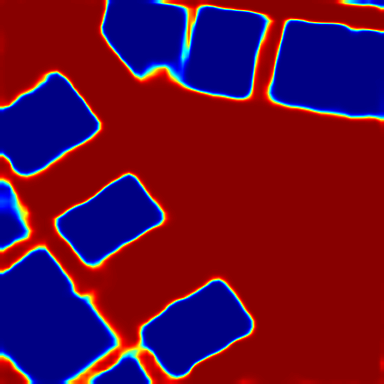} \\
\rothead{\centering Mask} &   \includegraphics[valign=m]{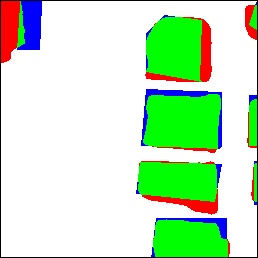}
                        &   \includegraphics[valign=m]{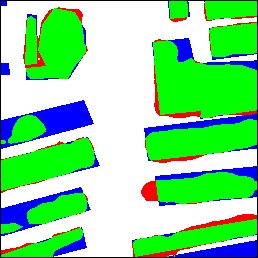}
                        &   \includegraphics[valign=m]{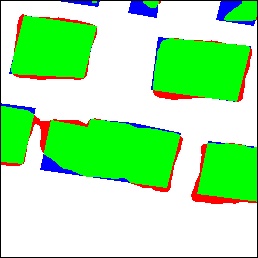}
                        &   \includegraphics[valign=m]{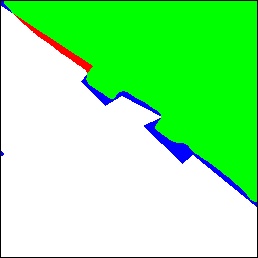}
                        &   \includegraphics[valign=m]{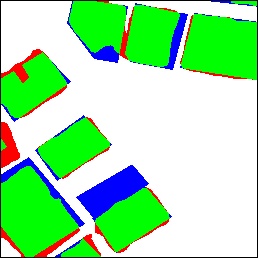}     \\
\rothead{\centering Per-pixel} &   \includegraphics[valign=m]{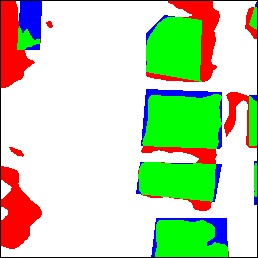}
                        &   \includegraphics[valign=m]{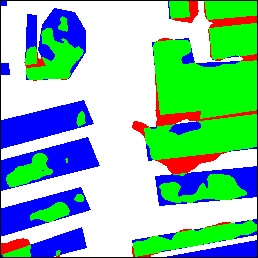}
                        &   \includegraphics[valign=m]{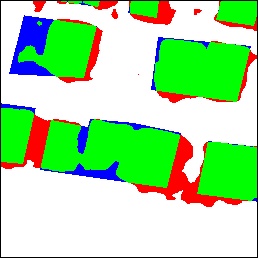}
                        &   \includegraphics[valign=m]{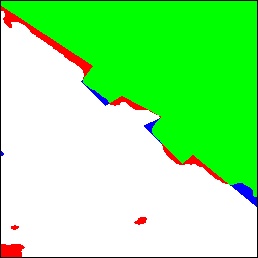}
                        &   \includegraphics[valign=m]{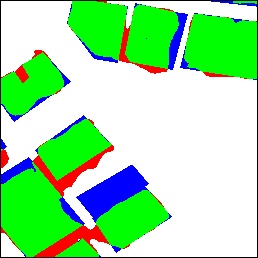}\\
                        &   \centering (a) 
                        &   \centering (b)
                        &   \centering (c)
                        &   \centering (d)
                        &   \centering (e)
\end{tabularx}
    \caption{Qualitative comparisons for mask classifications and per-pixel classification. (a-e) display five selected samples for illustration. The first and second rows are the post-change image and the label. The third and fourth rows display one predicted mask sample predicted to changed class and one predicted to unchanged class, respectively. The fourth and fifth rows present the results from the mask classification and per-pixel classification, respectively.}
\label{figure:pixel_ablation}
    \end{figure}

As shown in Fig. \ref{figure:pixel_ablation}, multiple predicted masks are presented to visually depict the mask classification process. For each sample, one mask predicted to be the unchanged class and one mask predicted to be the changed class are showcased. It can be observed that the masks predicted for the changed and unchanged classes focus on heterogeneous objects that have various spectral features, resulting in better CD performance at the object-based level.

\section{Conclusion}\label{s5}
In this article, we have proposed the MaskCD model for CD from VHR bi-temporal RS imagery, with a new paradigm introduced by taking CD as a mask classification task.
The core idea is to identify the change objects from bi-temporal images at the mask level through adaptively predicting and classifying object masks. Specifically, a hierarchical transformer-based siamese encoder is utilized to generate bi-temporal deep features, from which a cross-level change representation perceiver is built to extract change-aware representations as per-pixel embeddings that preserve multiscale contextual information of objects on the image. The MA-DETR decoder is adopted to obtain the per-segment embeddings that are the foundations for predicting masks and the corresponding class labels in a mask classification module. As a result, the MaskCD can directly obtain a change map at the mask label that can detect the objects with integrity and reduce background noise. 

Experiments on CD tasks on various scenes and land cover types, including building, cropland, urban, and landslides, show that MaskCD exhibits obvious advantages in CD from bitemporal VHR datasets. Compared with state-of-the-art CD methods, MaskCD can achieve better performance with fine-grained change maps.
More specifically, it can repress background noises while preserving detailed changes and handling multiple types of changed objects with superior integrity due to the exploitation of the novel mask classification paradigm. 
% In addition, MaskCD achieves dramatically faster convergence performance and high computational efficiency due to the deformable sampling strategy and masked attention mechanism, making MaskCD attractive for practical applications.

\clearpage
\bibliographystyle{IEEEtranN}
\bibliography{Ref}

\end{document}